%% file: main.tex
\documentclass[sigconf]{acmart}

\AtBeginDocument{%
  \providecommand\BibTeX{{%
    \normalfont B\kern-0.5em{\scshape i\kern-0.25em b}\kern-0.8em\TeX}}}

\copyrightyear{2022}
\acmYear{2022}
\setcopyright{rightsretained}
\acmConference[FAccT '22]{2022 ACM Conference on Fairness,
Accountability, and Transparency}{June 21--24, 2022}{Seoul, Republic of
Korea}
\acmBooktitle{2022 ACM Conference on Fairness, Accountability, and
Transparency (FAccT '22), June 21--24, 2022, Seoul, Republic of
Korea}
\acmDOI{10.1145/3531146.3533153}
\acmISBN{978-1-4503-9352-2/22/06}

\input{ules_latex_defaults}

\usepackage{caption}
\usepackage{subcaption}
\usepackage{wrapfig} %
\usepackage{tikz} %

\begin{document} %

\title[Post-Hoc Expl. Fail to Achieve their Purpose in Adv. Contexts]{Post-Hoc Explanations Fail to Achieve their Purpose\\ in Adversarial Contexts
}

\author{Sebastian Bordt}
\email{sebastian.bordt@uni-tuebingen.de}
\affiliation{
\institution{University of T{\"u}bingen, Germany}\country{}
}

\author{Mich{\`e}le Finck}
\email{michele.finck@uni-tuebingen.de}
\affiliation{
\institution{University of T{\"u}bingen, Germany}\country{}
}

\author{Eric Raidl}
\email{eric.raidl@uni-tuebingen.de}
\affiliation{
\institution{University of T{\"u}bingen, Germany}\country{}
}

\author{Ulrike von Luxburg}
\email{ulrike.luxburg@uni-tuebingen.de}
\affiliation{
\institution{University of T{\"u}bingen, Germany}\country{}
}

\begin{abstract}
Existing and planned legislation stipulates various obligations to provide information about machine learning algorithms and their functioning, often interpreted as obligations to ``explain''. Many researchers suggest using post-hoc explanation algorithms for this purpose. In this paper, we combine legal, philosophical and technical arguments to show that post-hoc explanation algorithms are unsuitable to achieve the law's objectives. Indeed, most situations where explanations are requested are adversarial, meaning that  the explanation provider and receiver have opposing interests and incentives, so that the provider might manipulate the explanation for her own ends. We show that this fundamental conflict cannot be resolved because of the high degree of ambiguity of post-hoc explanations in realistic application scenarios. As a consequence, post-hoc explanation algorithms are unsuitable to achieve the transparency objectives inherent to the legal norms. Instead, there is a need to more explicitly discuss the objectives underlying ``explainability'' obligations as these can often be better achieved through other mechanisms. There is an urgent need for a more open and honest discussion regarding the potential and limitations of post-hoc explanations in adversarial contexts, in particular in light of the current negotiations of the European Union's draft Artificial Intelligence Act.
\end{abstract}


\maketitle

\section{Introduction}
\label{sec-intro}
Explainability is one of the concepts dominating debates about the ethics and regulation of machine learning algorithms. 
Intuitively, requests for explainability are reactions to the prevalent unease about machine learning algorithms, including 
concerns regarding discrimination, biases, manipulation, and data protection. The fact that machine learning systems are often ``black boxes'' is considered a major hurdle towards their implementation, supervision and control, and explainability is often praised 
as a remedy against such risks. 
Existing legislation such as the EU General Data Protection Regulation ('GDPR') has sometimes been interpreted as containing a ``right for explanation''. The draft Artificial Intelligence Act, a piece of proposed EU legislation, alludes to explainability but does, in its current form, not make clear whether and when exactly explainability is legally required. 
On the technical side, explainability has evolved into its own field of research \cite{Molnar20}. The current machine learning literature knows two different approaches towards explainability. One approach is to build machine learning models that are constrained to be  ``inherently interpretable'' \cite{Rudin19}. The other approach is to use any machine learning model, even a ``back-box'', and then employ any of an increasing number of approaches in order to ``explain'' the behavior of the black-box after the decision has been made (``post-hoc''). Because there exists no general way to summarize the entire behavior of a black-box model, these explanations are usually local, meaning that they only describe the behavior of the function for a single prediction or decision.
The natural advantage of local post-hoc explanation methods, such as feature highlighting methods \cite{RibeiroEtal16, LundbergEtal2017} and counterfactual explanations \cite{WachterEtal17}, is that they place no constraints on model complexity and do not require model disclosure \cite{Barocas2020}. This has led a number of researchers to suggest that these methods might be able to comply with existing legal requirements \cite{WachterEtal17,Barocas2020}. 

$\,$\\
In this paper, we put forward an important distinction that has not yet been extensively discussed in the literature on explainable AI: whether the explanation's context is adversarial or cooperative. By ``cooperative contexts'' we broadly summarize situations where all involved parties have aligned interests. This includes model development and debugging, scientific discovery, and, to a degree, areas such as medical diagnosis. In a cooperative context, the explanation provider and the explanation receiver share the same interests: to identify the most suitable and insightful explanation algorithm for the given problem. 
In ``adversarial contexts'', in contrast, parties have opposing interests. This is the case for example when a bank denies a customer a loan and the customer wants to contest the decision because it was discriminatory. Since the explanation provider anticipates that one might use the provided explanations to challenge the functioning of the system, the explanation provider does not have any incentive to provide ``true'' insights into the functioning of the system; but rather to render the internal functioning of the machine learning system incontestable. Indeed, it has been pointed out repeatedly that post-hoc explanation algorithms can be manipulated or cheated upon %
\cite{SlackEtal20,%
AndersEtal20,%
SlackEtal21}.  %
Many machine learning papers on explanation algorithms implicitly consider  collaborative contexts where explanations are used to improve machine learning algorithms and can help developers to understand the biases of complex  systems, 
or where they are used in an explorative spirit towards new scientific discoveries 
\cite{ZedBoe22}. %
In contrast, the legal discussion focuses predominantly on adversarial scenarios. Here explainability is portrayed as a mechanism to add more transparency, fairness and accountability to AI, and post-hoc explanations are often seen as a technical tool to achieve these goals.\\

Combining insights from computer science, philosophy and law, we offer a critical multidisciplinary perspective on the usage of post-hoc explanations to achieve transparency and accountability obligations in adversarial contexts. We highlight the blurry legal landscape around explainability as well as the philosophical and technical limitations of post-hoc explanations.  
In Section~\ref{sec-adversarial} we introduce different scenarios -- cooperative and adversarial -- under which an external examiner might audit a black-box and its generated explanations. We focus on adversarial scenarios -- where the explanation provider has opposing interests to the explanation receiver -- and local post-hoc explanations -- where the explanation explains a single decision for one particular person. 
In Section~\ref{sec-legal} we argue that existing and planned legislation,specifically the GDPR and the EU Artificial Intelligence Act, can either be read as portraying explainability as one possible mechanism to achieve more transparency or as presenting it as a free-standing objective. We also highlight the current lack of legal certainty as to how existing legal norms around explainability ought to be interpreted and implemented. These issues have been the source of confusion and uncertainty. This is why we propose to capture the role of explainability by a discussion of its motivations: Explanations are thought to build trust, and also enable actions, such as debugging, contesting, recourse.  
In Section~\ref{sec-the-problems-with-ph-explanations} we show from a philosophical and technical perspective that the goals associated with explainability are unlikely to be achieved by post-hoc explanations. The reason is that the truth assumptions under which explanations are expected to fulfill their legal goal are lacking in the adversarial context. To the contrary, due to the inherent geometric ambiguity of local post-hoc explanations, the explanation provider has a multitude of options to influence explanations in a subtle, undetectable way and to pick those that suit her goals. 
 In Section~\ref{sec-testing} we show that testing explanations is also problematic. While at best we can test for internal consistency of the explanation with the decision, in more typical cases the explanations become redundant and we would better rely on testing decisions and predictions directly. 
In Section~\ref{sec-discussion} we conclude and argue that there needs to be a deeper and more honest debate about what the underlying objectives of explainability obligations are.  
We also argue that one needs to be honest about the fact that using a black-box entails considerable discretion: Neither post-hoc explanation methods, nor regulation can completely compel the deployer of a black-box to align his interests with the public good. As such, if one is absolutely unwilling to award any discretion to the deployer of the black-box, the only solution is to forbid its deployment and favor inherently interpretable or otherwise constrained machine learning methods. The question under which circumstances the deployment of a black-box might still be admissible depends on our ability to examine and audit the black-box. How exactly this might be done is still an area for future research. We hope that our paper contributes to an open discussion regarding the (lack of) potential of post-hoc explanations in the context of the on-going negotiation of the Artificial Intelligence Act.

\section{Explanations in cooperative and adversarial contexts}
\label{sec-adversarial}

In this work we broadly distinguish between ``cooperative'' and ``adversarial'' explanation contexts. In a {\bf cooperative context}, all parties involved in the process of building the system, providing explanations and using the system share the same goal: to create a system as good and supportive as possible. Prototypical examples are model debugging and scientific research. But also a company building a medical decision support system, say for skin cancer detection, will closely collaborate with the doctors who use it \cite{tschandl2020human}. The company's goal would be to provide explanations that are as helpful as possible. The situation is very different in {\bf adversarial contexts}, where parties do not share the same goal, such as in the oft-repeated example of a denial of a loan application. Here, the applicant and bank have opposing interests and incentives. Accordingly, should the bank be mandated to provide the applicant with an explanation, this explanation will be shaped by the bank's incentives and existing power asymmetries. For reasons that we outline below, the distinction between cooperative and adversarial contexts is crucial. In particular, we argue that local post-hoc explanations, which have a variety of use-cases in the cooperative scenario, are  pointless or even harmful in adversarial contexts.

\subsection{Parties involved in the adversarial explanation process} 
\label{subsec-parties}

We consider adversarial explanation contexts where an AI decision system is used to make decisions about individuals. Prominent examples are university admissions, job and loan applications, or  bail and sentencing decisions. Under existing and planned legislation, such as the EU Artificial Intelligence Act, the {\em creator of the system} ought to provide information about how the system comes to its decisions (see Section \ref{sec-legal} below for a detailed discussion of the legal background). 
The creator of the system is the entity that has built the machine learning system and uses it to support decision making.\footnote{The creator is mainly the developer. But since the developer develops the system for a user, their interests typically align. Hence we do not distinguish developer and user, and use the term ``creator'' instead. } The creator could be a private company (such as a bank) or a public entity (such as a university). The {\em decision subject} is the person about whom the automated system makes a decision: the person who applies for a loan, or the person who applies to for university admission. %
After the decision has been communicated, the {\em explanation recipient} asks for an explanation, which is communicated by the {\em explanation provider}. The explanation recipient could be the decision subject herself, or an external {\em examiner} who is supposed to investigate the decisions or explanations on behalf of the decision subject or to defend her interests. The explanation provider is typically the creator of the system.\footnote{Similar distinctions were introduced by \cite{Tomsett18}.}

\subsection{Machine learning problem: Supervised learning, tabular data, point-wise post-hoc explanations}
In our technical discussion, we assume that the inputs $x \in \R^d$ of a decision algorithm are given in {\bf tabular} form. Each dimension of the input encodes a different property of a person, for example age, income, etc.  Typically, the number of dimensions $d$ is large: persons are described by dozens or hundreds of features. A machine learning algorithm is used to learn a {\bf decision function} $f: \R^d \to \R$. The resulting decision $y=f(x)$ for input $x$ could be a binary decision (``receives the loan'' or ``does not receive the loan'') or a numeric risk score on which such a decision is based, as in the often discussed COMPAS algorithm to predict recidivism risk. 
We focus on {\em supervised machine learning}, where $f$ is learned based on {\bf training data} consisting of pairs $(x_1, y_1), ..., (x_n, y_n)$ with $x_i$ the training points and $y_i$ the training labels. An explanation algorithm $E$ is an algorithm operating on a decision function with the purpose of explaining it. We focus on {\bf local post-hoc explanation algorithms}: The explanation algorithm $E$ gets queried with  a data point $x$ and the corresponding decision $y$, and produces an explanation $E(x,y)$.  Internally, the algorithm has access to the decision function $f$, and in some cases also to the training data. The explanation $E(x, y)$ is supposed to explain why the decision function $f$ came to decide $y$ for $x$. The explanation can be in linguistic form. For example, 
``The low income of Mr. Smith was relevant for the refusal of the loan'' or ``Mr. Smith would have received the loan had his income been 10.000 Euros higher''.

\subsection{Explanation algorithms that fall into this framework}

In this paper we consider local post-hoc explanation algorithms such as LIME, SHAP, and DiCE \citep{RibeiroEtal16,LundbergEtal2017,mothilal2020explaining}. The explanations generated by these algorithms do not provide a global or holistic view of the decision function $f$ but merely try to explain individual decisions $y=f(x)$. The often-cited advantage of these algorithms is that they work, at least in principle, for {\em any} decision function \citep{RibeiroEtal16,Barocas2020}. Different algorithms take different approaches as to what constitutes an explanation: LIME and SHAP provide {\em feature attributions} that aim to quantify the influence of the different input-features for the particular decision. Feature attributions correspond to the linguistic form ``The low income of Mr. Smith was relevant for the refusal of the loan''. Another approach is to provide {\em counterfactual explanations} \citep{WachterEtal17}.
These explanations are based on searching for a sufficiently close or the closest alternative point $x'$ to the actual input point $x$ that yields a decision $y'=f(x')$ that differs from the original decision $y=f(x)$.
Comparing the two we can arrive at factors that are relevant to the decision  \citep{kommiya2021towards}.
The resulting counterfactual explanations have 
the linguistic form ``Mr. Smith would have received the loan had his income been 10.000 Euros higher''.

\section{Legal framework: Explainability in EU Law}
\label{sec-legal}

This paper argues that post-hoc explanation algorithms are unsuitable in adversarial contexts. Before we elaborate this %
from a philosophical and technical perspective (Section \ref{sec-the-problems-with-ph-explanations}), it is important to understand the related legal framework. We focus on European Union law as the EU has often been a first-mover regarding the regulation of data and its analysis, and over time its legislation will likely inspire other jurisdictions (for a broader view, see \cite{KamUrb21}). Our analysis focuses on the draft Artificial Intelligence Act (AIA), a piece of proposed legislation that would be the first to specifically target AI. This pioneering approach would be a global blueprint for the regulation of AI. In its current form it creates different legal obligations for different AI applications on the basis of the perceived risks. The AIA would apply to general AI systems (Section~\ref{sec:AIA}). We also consider the General Data Protection Regulation (GDPR), which applies to the processing of personal data (Section \ref{sec:GDPR}). It will be seen that whereas EU law contains various obligations to provide information about a machine learning algorithm and its functioning, it remains unclear how these legal norms should be implemented from a technical perspective and whether explainability should be understood as a free-standing legal obligation or whether it should rather be seen as one of various mechanisms to achieve algorithmic transparency (Section \ref{sec:subcomponent}). To better understand the latter we also review their underlying rationales and objectives from a philosophical and legal perspective (Section \ref{sec-rationales-behind-norms}).

\subsection{The draft Artificial Intelligence Act (AIA)} 
\label{sec:AIA}
The current draft of the AIA defines AI systems as ``software (...) that can, for a given set of human-defined objectives, generate outputs such as content, predictions, recommendations, or decisions influencing the environments they interact with''. Generally, the AIA regulates AI on the basis of its perceived risk by introducing four different categories of AI. Most relevant to our discussion are the two categories of systems that are high-risk, as opposed to systems that are not high-risk (the remaining two categories are practices that are subject to qualified prohibitions, and a residual category of AI systems that includes law enforcement software, emotion recognition system, biometric categorisation systems and deep fakes) \cite{Veale2021}. The stronger the risk, the heavier regulatory obligations apply, also regarding transparency and interpretability.\\

There are two categories of {\bf high-risk AI systems}. First, AI systems that relate to products that are already subject to supranational harmonisation, namely AI systems intended to be used as a safety component of a product, which are themselves products covered by Union harmonising legislation or which are required to undergo third-party conformity assessments. Second, a list of systems that are currently considered to carry a high-risk such as, for instance, biometric identification systems, systems for the management and operation of critical infrastructure, those used in education and employment, some law enforcement systems as well as others (see further Art 3(1) of the draft AIA).
Article 13 governs explainability for high-risk AI systems, which have to be ``designed and developed in such a way to ensure that their operation is sufficiently transparent to enable users to \textit{interpret} the system’s output and use it appropriately''. Furthermore, users (the entity deploying the AI) need to have access to instructions for use in an appropriate digital format that contains information about the characteristics, capabilities and limitations of performance, including information about the level of accuracy, robustness and cybersecurity, risks to health, safety or fundamental rights, specifications for the input data, expected lifetime of the AI system and necessary maintenance measures. Finally, human oversight must be ensured. These measures are designed to minimize risks to health, safety or fundamental rights. Human oversight shall either be (i) identified and built into the system by the provider before it is placed on the market or put into service, or (ii) identified by the provider before the system is placed on the market or put into service but only implemented by the user.\\[20pt]

{\em In its current version, the AIA would thus require that high-risk AI systems are sufficiently transparent to enable the interpretation of the system's output.} Is this an explainability obligation? Recital 47 sheds some light on how to interpret these notions. It specifies that high-risk AI systems should be transparent to a ``certain degree'' to ``address the opacity that may make certain AI systems incomprehensible to or too complex for natural persons''. To this end, users ``should be able to interpret the system output and use it appropriately'' through the provision of ``relevant documentation and instructions of use''. {\em This does not read like an obligation to make systems explainable in the sense that the way in which data has been processed must be entirely traceable.} Rather, the AIA would require that an ``interpretation'' of the output must be facilitated through sufficient transparency. Importantly, this does not necessarily seem to imply that an absolute truth must be identified post-hoc (see Sections \ref{sec-algorithm-incomplete} and \ref{sec-explanations-algorithms-own-world} below) but rather the overall functioning of the system and how it comes to an output. 
{\em The draft AIA leaves open the question of what transparency and interpretability imply from a technical perspective.}  This certainly includes the elements listed in its Article 16 such as technical documentation, keeping logs or quality management systems. Article 13 leaves open whether there are additional requirements and what, exactly, interpretability requires from a technical perspective. If input data ought to be entirely traceable, ``black-box'' systems cannot be used in high-risk applications. This highlights that it is important to think about the objectives of transparency and explainability. If these can be achieved through alternative means, excluding black-box systems such as deep neural networks from high-risk scenarios (such as healthcare as devices falling under the Medical Devices Regulation qualify as high-risk) might unduly hinder innovation in important domains.\\

Article 52 AIA creates some general transparency obligations for {\bf AI systems that are not high-risk}. These are general disclosure obligations such as to (i) inform users that they are interacting with an AI system unless this is obvious from context, (ii) users of an emotion recognition system or biometric categorization system shall inform natural persons exposed thereto, (iii) deep fakes must be disclosed as such. Some exceptions apply where the AI is used in the context of law enforcement. These are thus obligations of transparency that require disclosure that AI is used, as opposed to how it is used. \\

To summarize, {\em the draft AIA would thus not, in its current form, create a general explainability obligation for machine learning systems.}  Such an obligation clearly is not foreseen in relation to AI systems that are not qualified as high-risk. Arguably, there is also no explainability obligation in relation to high-risk AI systems. Rather, what is required is transparency of the system's functioning and output generation. This transparency must make these elements interpretable but not necessarily amount to the provision of an explanation as it is commonly understood in computer science.

\subsection{The General Data Protection Regulation (GDPR) }
\label{sec:GDPR}

The GDPR creates some general transparency requirements that form part of the data controller’s (the entity that determines the purposes and means of processing) general informational obligations vis-à-vis the data subject (the natural person that personal data relates to). In addition, it also contains a specific regime for ``solely automated data processing''. In contrast to the draft AIA, which creates vague obligations resting on the user, the GDPR creates specific rights for the individual subjected to such decisions. \\

Article 13 requires that data controllers provide specific information to data subjects where personal data is collected from them at the time of collection such as whether ``automated decision-making'' is used, and, if so, provide {\em ``meaningful information about the logic involved$\,$\footnote{The exact interpretation of ``logic'' in the GDPR is not settled but likely does not refer to understandings of this term in philosophy or computer science.}, as well as the significance and the envisaged consequences of such processing''.}  Article 14(1)(h) creates the same obligation in cases where data is not directly collected from the data subject. Pursuant to Recital 62 this information does not have to be provided where it is redundant, or where compliance proves impossible or involves a disproportionate effort. The same wording can also be found in Article 15, which deals with the data subject’s right to access data. Whereas Articles 13 and 14 relate to the pre-processing stage, data subjects can exercise their rights under Article 15 at any time, including after processing has taken place. This raises the question of whether – despite the identical wording of these provisions – Article 15 may substantively require something different when referring to the ``logic'' of the automated decision-making process. \\

{\em There is no general right to an explanation under the GDPR.} Some explainability requirements may, however, arise in respect of machine learning algorithms that produce legal effect or similarly significantly affect a data subject. Article 22 creates a qualified prohibition of ``solely automated data processing'', including profiling. %
This implies that such techniques can only be used in some circumstances, namely (i) where necessary to enter into or perform a contract between the data subject and controller, (ii) where it is authorized by law or where the data subject has provided explicit consent. In these circumstances automated processing can take place, but the data subject has the right to human intervention and to express her point of view and to contest the decision. Recital 71  mentions an additional element, namely that the data subject has the right ``to obtain an \textit{explanation}'' after human review of the decision ``and to challenge this decision''.\footnote{Children should not be subject to automated decision-making.} Recitals, however, do not have the same legally binding force as the text of the GDPR itself. \\

Over the past years there has been a vivid academic debate around whether the reference to ``an explanation'' in Recital 71 amounts to a ``right to an explanation'' that data subjects can exercise via-à-vis controllers 
\cite{WacMitFlo17} 
\cite{MalCom17}
\cite{SelPow18}
\cite{Edwards17}. 
The Article 29 Working Party's guidance suggests that Article 22, read in conjunction with Recital 71, should be understood to require that controllers (i) tell data subjects that they are engaging in automated decision making, (ii) deliver meaningful information about the logic, and (iii) explain the processing’s significance and envisaged consequences. The information provided should include details about the categories of data; why data is seen as pertinent; how profiles are built; why the profile is relevant for the decision-making process and how it is used to reach a decision about the data subject. The last three criteria appear to apply to profiling only \cite{wp16}. 
Information with respect to the ``logic'' means ``simple ways to tell the data subject about the rationale behind, or the criteria relied on in reaching the decision''. What is required is ``not necessarily a complex explanation of the algorithms used or disclosure of the full algorithm''. Nonetheless, the information transmitted to the data subject should be sufficiently comprehensive to ``understand the reasons for the decision''. Thus, an explanation of algorithms or disclosure of the full algorithm isn’t ``necessarily'' required and that the controller ought to find ``simple ways to tell the data subject about the rationale behind, or the criteria relied on in reaching the decision''. {\em Unfortunately, this guidance leaves a lot of room for doubt regarding what exactly is required of controllers.} In any event the GDPR does not create a general right to an explanation but applies only to automated decision-making that legally affect the data subject or have similarly significant effects on them. 

\subsection{Explainability as a sub-component of transparency} 
\label{sec:subcomponent}

While there is a persistent myth that EU law requires that all decisions based on AI are ``explainable'' our analysis has painted a more nuanced picture. First, there is no overarching explainability norm that would apply to any usage of AI. To what degree secondary law requires explanations has not been authoritatively settled. Ultimately, the Court of Justice of the European Union will need to settle this question in respect of the GDPR. Concerning the draft AIA, however, legislators should clarify in the final text whether explainability is a free-standing legal obligation in respect of high-risk AI systems or whether it should rather be understood as a sub-component of transparency.
As shown above, it is indeed possible to read references to explainability as elements of the broader transparency obligation. Article 13 AIA is explicitly about transparency,  but the reference that this transparency must allow users to ``interpret the system’s output'' has been understood as an explainability obligation by some. Further iterations should clarify the link between transparency and explainability to enhance legal certainty. An analysis of the history behind the AIA confirms the lack of precision of the AIA itself. The EU High Level Expert Group on AI's report on the one hand portrayed explainability as a component of transparency. On the other hand, it repeatedly referred to another concept, ``explicability'', which was introduced as an ethical principle and as the ``procedural dimension'' of fairness. In contrast, the AIA White Paper made no reference to explainability other than to mention that symbolic reasoning could help make deep neural networks more explainable. This part of the AIA legislative history underlines the lack of consensus about what \textit{exactly} explainability is.
Similarly, the GDPR could also be read as referring to explainability as a sub-component of transparency. Articles 12-15 derive from the core data protection principle of transparency in Article 5(1)(a) and likewise, one reading of Article 22 in conjunction with relevant recitals could also be understood as a more general transparency rather than explainability obligation. \\

This, of course, raises the question of what transparency means and what it should enable. There is broad consensus that the GDPR requires that decisions reached through automated decision making be justifiable. Indeed, Hildebrandt has highlighted that data protection requires ``the justification of such decision-making rather than an explanation in the sense of its heuristics'' (p. 113 in \cite{Hildebrandt19}). 
Kamimski and Urban deem that justification should enable ``understanding, revealing and making challengeable the normative grounds of a decision''
(p. 1980 in \cite{KamUrb21}).
Wachter, Mittelstadt and Russell have argued that explainability is ultimately designed to help the data subject understand, contest and alter decisions and that this could also be achieved by counterfactual explanations \cite{WachterEtal17}. 
If explainability is merely one means of achieving transparency, there needs to be a more thorough discussion as to what other, alternative, means of achieving transparency there are, particularly in situations where explainability strictu sensu proves impossible. Considering the lack of consensus as to how the legislative texts of the AIA and the GDPR ought to be interpreted and applied in practice, it is helpful to consider their underlying objectives.

\subsection{Rationale and objectives of explainability norms in an adversarial setting}
\label{sec-rationales-behind-norms}

The vague formulation of explainability rights, coupled with uncertainty regarding their function makes it legitimate to ask whether explanations serve any meaningful purpose.  Indeed, as Edwards and Veale \cite{Edwards17} have argued, ``the search for a legally enforceable right to an explanation may be at best distracting and at worst nurture a new kind of transparency fallacy''.  This is essentially a warning that if explainability obligations just become a box-ticking exercise,
they might give a misleading appearance of compliance rather than to be of any real value to the decision subject. In addition, explainability rights in the GDPR inevitably also suffer from the general shortcomings of the low enforcement of the GDPR.\\

In order to better understand the above-examined norms we propose to consider their underlying objectives.
Before discussing legislative history let us recapitulate what philosophers have identified as main objectives for algorithmic explanations.\footnote{
Questions regarding ``Explanations'' have been discussed since the beginning of philosophy, with a strong revival in the philosophy of science of the last century, treating scientific explanations
\cite{Braithwaite1953,Popper1959,Hempel1965,Achinstein1983,Salmon1971}, causal explanations \cite{Lewis1973,Spohn1980,Spirtes1993,Pearl2000,Woodward2003}, and non-causal explanations \cite{Reutlinger2018}.
We refer the interested reader to
\cite{Woodward2021,Salmon1989}
and restrict our discussion to the context of machine learning.}
One major motivation for explainability of AI systems is the hope that this may foster  \textbf{trust} in these systems  \cite{Chazette21,Kaestner21jul,Eschenbach21,EthicsEU2019}. This has been called the ``Explainability-Trust'' hypothesis \cite{Kaestner21}.\footnote{For further references, see \S 2 therein.}
The hypothesis is controversial, and it is not exactly clear how explanations would induce trust. The underlying rationale seems to rest on an analogy with human interactions. 
Consider decisions made by human experts. When the decision doesn't satisfy us, we are drawn to ask for an additional explanation. Given such an explanation, we may check whether it conforms to our expectations about good decision making. If so, this may be a ground for further trusting the decision maker. This is not a one-shot process, but an ever evolving interaction on a long term time-scale. We tend to trust a person that proved repeatedly to predict correctly, make good decisions, or provide well informed explanations. The trust raising potential of an explanation however requires that we can submit explanations and decisions to tests, possibly by delegating it to other experts. The trust raising potential of a single explanation thus presupposes that the explanation provider stays in the information-exchange on the long run: only then does she have an incentive to provide a correct explanation, since an incorrect one would lead to a loss of trust in the long run but not in a one-shot exchange.  
If an algorithm rather than a human expert makes a decision, we might have similar expectations. We would like to engage in a similar information exchange with an algorithm as we engage with humans. The demand for an explanation is then a demand for a piece of communicative interaction.
The hope that this builds trust stems from the intuition that the interaction with the algorithm is similar to the interaction  among humans, as depicted above. This assumption may however fail either because the algorithmic explanations cannot be submitted to sensible tests or because the exchange is one-shot and not long run. In the first case, explanations loose their trust raising potential. In the second, the explanation provider may not have the incentive to tell the truth. 
A second implicit motivation for explainability stems from the idea that information provided by explanations can be {\bf used to perform actions}, and may in fact be needed for such actions. In the adversarial setting, a data subject might want to use an explanation to \emph{contest} a decision \cite{WachterEtal17,Barocas2020}, by claiming, or arguing that the decision is not right, not good, or not fair. The data subject might also want to use the explanation for \emph{recourse}, in order to do better next time \cite{WachterEtal17,Venkatasubramanian2020,karimi2021survey} (see also \cite{Adadi2018,Liao2021}).\footnote{Other actions belong more properly to the collaborative setting, such as debugging, improving, correcting, learning, understanding and testing.} 
But such explanations are only of value when true or correct. A false explanation will not help in doing better next time, and may even be devised such as to render a decision incontestable.  %
\\

The two motivations from philosophy --- building trust and enabling recipients to act --- can also be found as objectives in the legal texts. 
The  EU High Level Expert Group on AI described explainability as one {\bf tool to achieve trust} in AI systems  
\cite{EthicsEU2019}.\footnote{With the consequence that explainability would also play a role in stimulating the adoption of AI and the competitiveness of the internal market.} The AIA provides that explainability norms are designed to allow users to fully understand the capacities and limitations of high-risk systems, leading again to trust. 
Partly related to trust, one can understand explainability as a tool for {\bf risk management}, in line with the AIA's overall risk-based approach. Indeed, for high-risk AI systems, transparency must be ensured by monitoring the system's operation, detect signs of anomalies, dysfunctions and unexpected performance in order to counteract automation bias or %
to potentially intervene in the system (the idea of a ``stop button''). The European Commission White Paper also emphasized the risk-based approach and stressed that due to the potential scale of AI  systems 
\cite{european2020white}: 
a hidden bias or an incorrect assumption of an AI system, say deciding on tens of thousands of university admission decisions, will have a large systemic effect. This differentiates large-scale AI systems from human decision-making systems.
In philosophy explanations are considered as a tool towards future actions. Similarly, the legal discussion also portrays explainability as an {\bf enabling right}. The High-Level Expert Group on AI has drawn attention to the fact that to be able to contest decisions, they must be traceable. 
Also outside the AIA and the GDPR, explainability serves a related purpose. In consumer protection law, explainability is linked to the {\bf unequal power dynamics} between the business and the consumer. In the public administration, it has been argued that being subjected to an intransparent black-box decision would undermine {\bf human dignity} and is also to be avoided, unlike in the private sector, individuals cannot vote with their feet and go elsewhere. \\

{\em Overall, the motivations for explanations seem to presuppose that such explanations are true or correct. } Only then does a single explanation raise trust, and only then can an explanation be used to perform the intended actions, such as contesting or recourse. We will, however, see in the next section that this truth-presupposition for explanations fails in adversarial scenarios of algorithmic post-hoc explanations.

\section{The problems with post-hoc explanations in adversarial contexts}
\label{sec-the-problems-with-ph-explanations}

\begin{figure*}[t]
     \centering
     \begin{subfigure}[b]{0.24\textwidth}
         \centering
         \includegraphics[width=\textwidth]{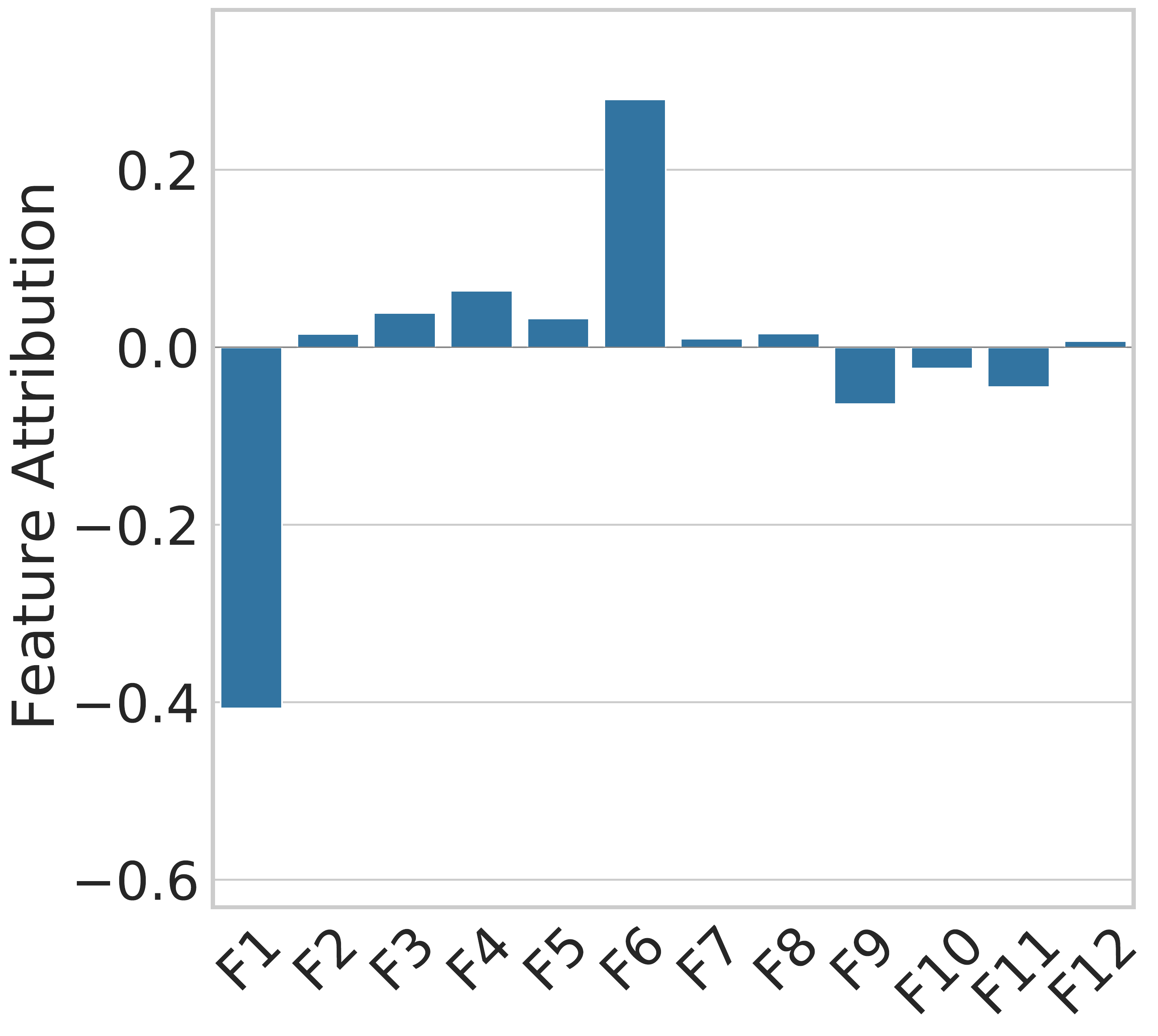}
         \caption{SHAP}
         \label{fig:adults_shap}
     \end{subfigure}
     \hfill
     \begin{subfigure}[b]{0.24\textwidth}
         \centering
         \includegraphics[width=\textwidth]{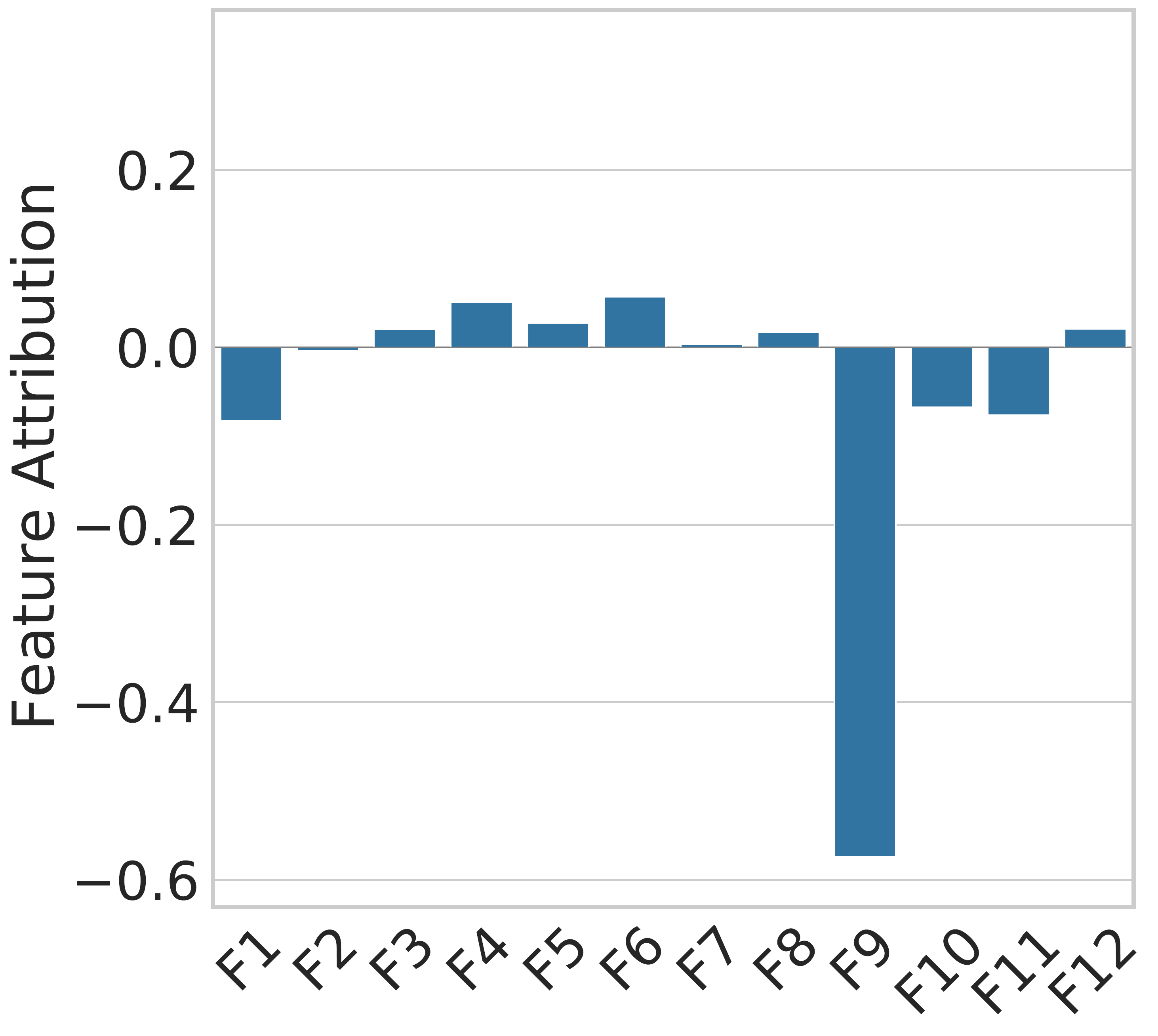}
         \caption{LIME}
         \label{fig:adults_lime}
     \end{subfigure}
     \hfill
     \begin{subfigure}[b]{0.24\textwidth}
         \centering
         \includegraphics[width=\textwidth]{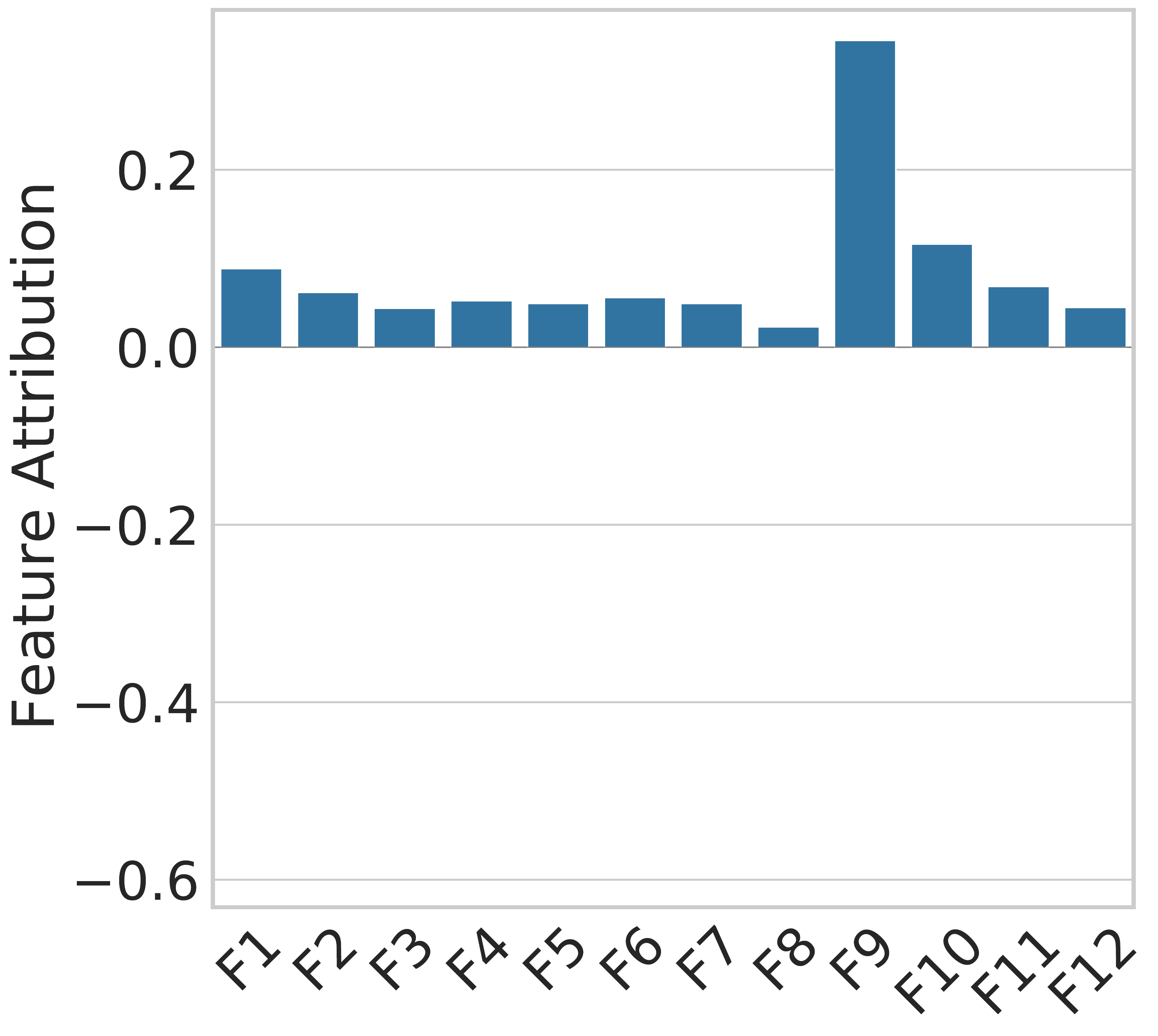}
         \caption{DiCE}
         \label{fig:adults_dice}
     \end{subfigure}
     \hfill
     \begin{subfigure}[b]{0.24\textwidth}
         \centering
         \includegraphics[width=\textwidth]{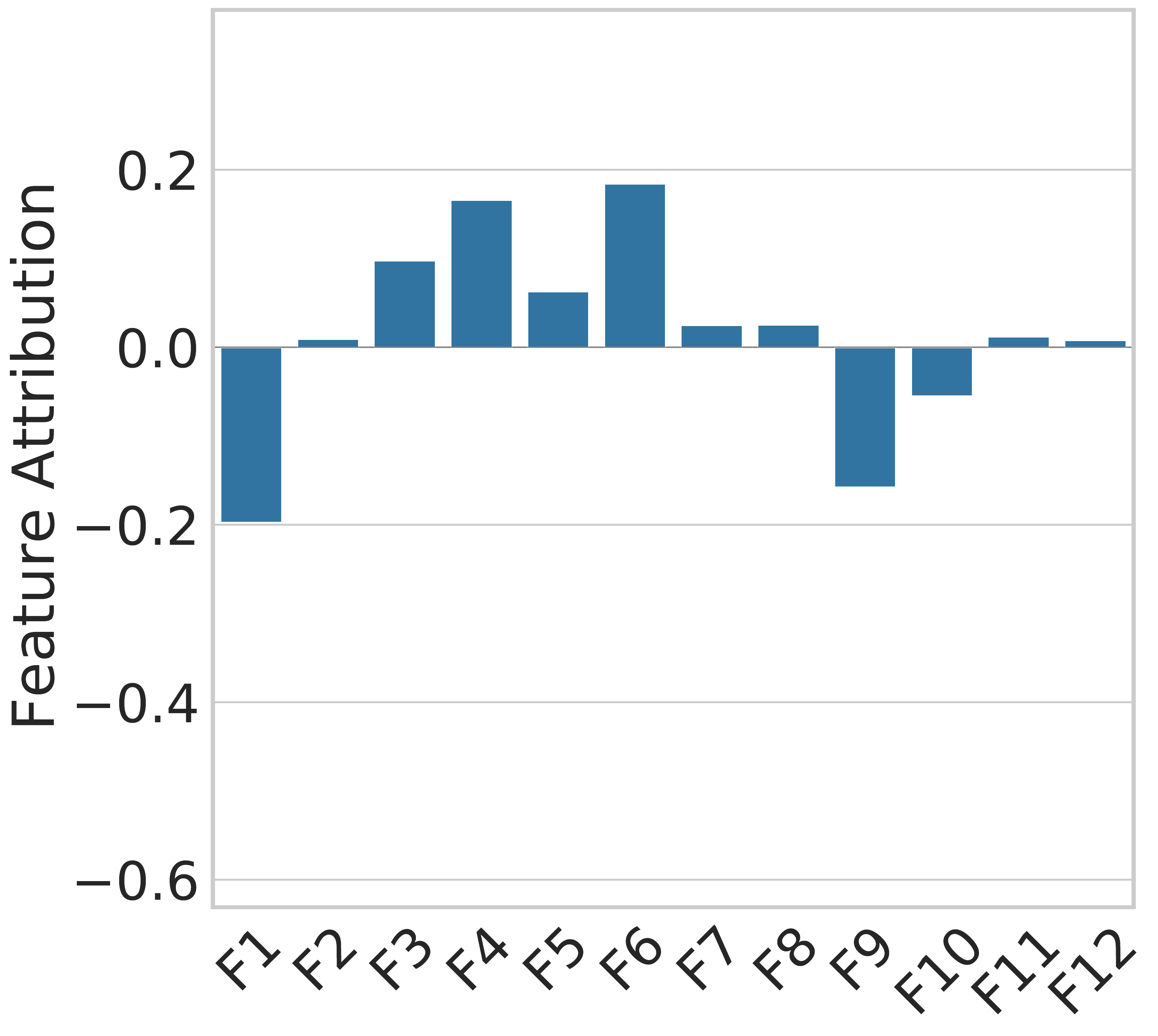}
         \caption{Interventional SHAP}
         \label{fig:adults_interventional}
     \end{subfigure}
        \caption{Different explanation algorithms lead to different explanations. Depicted are the feature attribution explanations of four different explanation algorithms: Exact SHAP for trees \cite{lundberg2020local}, LIME \cite{RibeiroEtal16}, DiCE \cite{kommiya2021towards}, and Interventional SHAP \citep{janzing2020feature}. All four explanation algorithms attempt to explain the prediction for the {\it same} individual with the {\it same} decision function (a gradient boosted tree) on the {\it same} dataset (Adult-Income). The idea of feature attribution explanations is to determine how much each dimension of the input contributed towards the decision. The figures depict these attributions by drawing a bar for each of the 12 input dimensions. The larger the bar, the higher is the influence of the corresponding feature. Some methods distinguish between positive and negative attributions.
        In the depicted example, the first bar in Panel (a) is relatively large, which indicates that the SHAP algorithm determined that the value of the first feature contributed strongly to the prediction. The DiCE algorithm in Panel (c), in contrast, determined that the value of feature 9 contributed most strongly to the prediction. More figures showing results for other data points can be found in the supplement. }
        \label{fig:different_explanations}
\end{figure*}

We now discuss the problems with post-hoc explanations in adversarial scenarios. What can we expect from an algorithmic explanation in these contexts? We roughly know what to expect from human explanations. For example, witnesses giving evidence in court are expected to tell the truth. Can we expect something similar of an algorithmic explanation? If the algorithm decided, for example, to reject a loan application, can we expect to discover the true reason why it decided to do so? The answer is that we cannot, for two reasons. First, the algorithm's view of the world is coarse-grained and incomplete, and this significantly restricts the vocabulary available for potential explanations (Section \ref{sec-algorithm-incomplete}). Second, even within the limited picture of the world that the algorithm has access to (the ``algorithm's own world'') uniquely preferred or ``ground truth'' explanations do not exist (Section \ref{sec-explanations-algorithms-own-world}). This directly ties with the computer science perspective of why post-hoc explanations should not be used in adversarial contexts: the task of providing post-hoc explanations is underdetermined. The objective of the adversary explanation provider is to deploy a classifier that has high accuracy and generate post-hoc explanations that cannot be contested by the data subject or an examiner. We argue that due to the high degree of ambiguity inherent to algorithmic explanations, the adversary has sufficient degrees of freedom to devise incontestable explanations  -- even without explicitly optimizing against a particular explanation method \citep{SlackEtal20,SlackEtal21a}. We identify four key quantities that allow the adversary to influence the resulting explanations: the choice of an explanation algorithm and its particular parameters (Sections \ref{sec-cs-different-algorithms-different-explanations} and \ref{sec-cs-parameters}); the exact shape of the high-dimensional decision boundary (Section \ref{sec-cs-decision-boundary}); and, when applicable, the choice of the reference dataset (Section \ref{sec-cs-reference-dataset}). This section contains a number of figures and simulation results. Additional figures can be found in the supplement. The code to replicate the results in this paper is available at \url{https://github.com/tml-tuebingen/facct-post-hoc}.

\subsection{The algorithm's view of the world is coarse-grained and incomplete - this limits potential explanations}
\label{sec-algorithm-incomplete}

Learning and explanation algorithms only have access to a coarse-grained description of the real world. Their vocabulary is restricted to certain features, and possible relations between them.
The ``experience'' of such algorithms given by the finite training data is formulated in the restricted vocabulary and provides only a small window to the world. Overall, the algorithm's representation of the real world is coarse-grained and incomplete.\footnote{Similar issues were discussed in \cite{Barocas2020,JacobsWallach21}.} The learning algorithm just sees features and training labels. The explanation algorithm, additionally, sees the learning algorithm's association between input and output. This is what we call ``the world of the explanation algorithm'', and this is all what it can exploit. As a consequence, all the explanation algorithm could talk about are geometric properties in the world of the algorithm: 
distances of points to the decision surface, proximity between points, their true or predicted labels, the gradient of the decision function at a point, the necessary change of a feature to change the decision, etc. Although a true explanation for a decision might exist in the real world, it might not be represented in the data or other aspects of the algorithm's world, which could thus not provide any such explanation. This is even the case in a cooperative setting. Consider the example of a medical diagnosis of a disease for which a true (say, causal) explanation exists in the real world. If the learning algorithm was trained on feature-based data such as age, blood pressure, etc, the explanation algorithm could suggest that age was the cause. However, in reality the cause for the disease may not be age, but rather a smoking habit that was not represented in the data. So even if a true explanation exists (say, a cause) this may neither be identifiable nor expressible by the explanation algorithm.

\subsection{Even within the algorithm's own world, a unique preferred reason does not exist} \label{sec-explanations-algorithms-own-world}

Even within the limited world that the explanation algorithm has access to, a ``true  internal reason'' why the learned decision function comes to a certain decision does generally not exist. This is particularly the case for complicated black-box functions. Even machine learning experts digging into the learning algorithm or properties of the function could not reveal a unique true reason. 
All we can do is to provide vague approximations of how the algorithm arrives at its decision, by summarizing which features contributed how much to the decision (the approach of LIME and SHAP), or whether a change in some features would alter the decision (the approach of counterfactual explanations). For example, in the case of a loan rejection, we might want to know whether it was rather our low income or our postal code which determined the decision, and whether we could change something about the decision, if in the future we had a higher income or moved to another area. However, these explanation attempts are all subject to choices. A mathematically unique way to determine how much each feature of a complicated black-box function contributed to the decision does not exist. 
Consequently, all feature attribution methods rely on particular assumptions and mechanisms in order to construct explanations: LIME, for example, looks at the gradient of the decision function at the point to be explained \citep{RibeiroEtal16,GarLux20}. SHAP compares the point with other datapoints from a reference population \citep{LundbergEtal2017,GhalebikesabiEtal21}. Yet another approach would be to re-train the classifier on subsets of features or to use counterfactual feature importance, where one looks at the distance to the decision surface in various directions. All these mechanisms and choices seem plausible 
but, as we will see in the next section, they all deliver different explanations.

\subsection{Different explanation algorithms lead to different explanations}
\label{sec-cs-different-algorithms-different-explanations}

Different explanation algorithms lead to different explanations \citep{krishna2022disagreement}. This is true even if the algorithms have access to exactly the same information (the geometry of the data, the learned decision function, etc). In an adversarial context, this is problematic because it means that the creator of the system can modify the explanations by choosing a particular explanation algorithm. In practice, different explanation algorithms lead to different explanations even on the most simple machine learning problems. In high dimensions, that is in real-world problems, the difference between the explanations obtained from two different explanation algorithms can be so significant that the explanations are entirely different. This is illustrated in Figure \ref{fig:different_explanations}. The figure depicts the feature attribution explanations that four different explanation algorithms determined for the {\it same} individual. From the difference between the four panels in Figure \ref{fig:different_explanations} it is quite clear that different explanation algorithms can lead to markedly different explanations, even if they all attempt to explain the same decision for the same individual.\footnote{The reader who is acquainted with the internal mechanics of the depicted explanation method might feel that a direct comparison between the different methods is unwarranted, because different methods measure different aspects of the underlying decision function \citep{camburu2019can}. Note, however, that this is exactly the point that we want to make by explicitly contrasting the different attributions.} Details on the machine learning problem, dataset and explanation algorithms can be found in the supplement.\\

\begin{figure}[t]
\centering
     \begin{subfigure}[b]{0.23\textwidth}
         \centering
         \includegraphics[width=\textwidth]{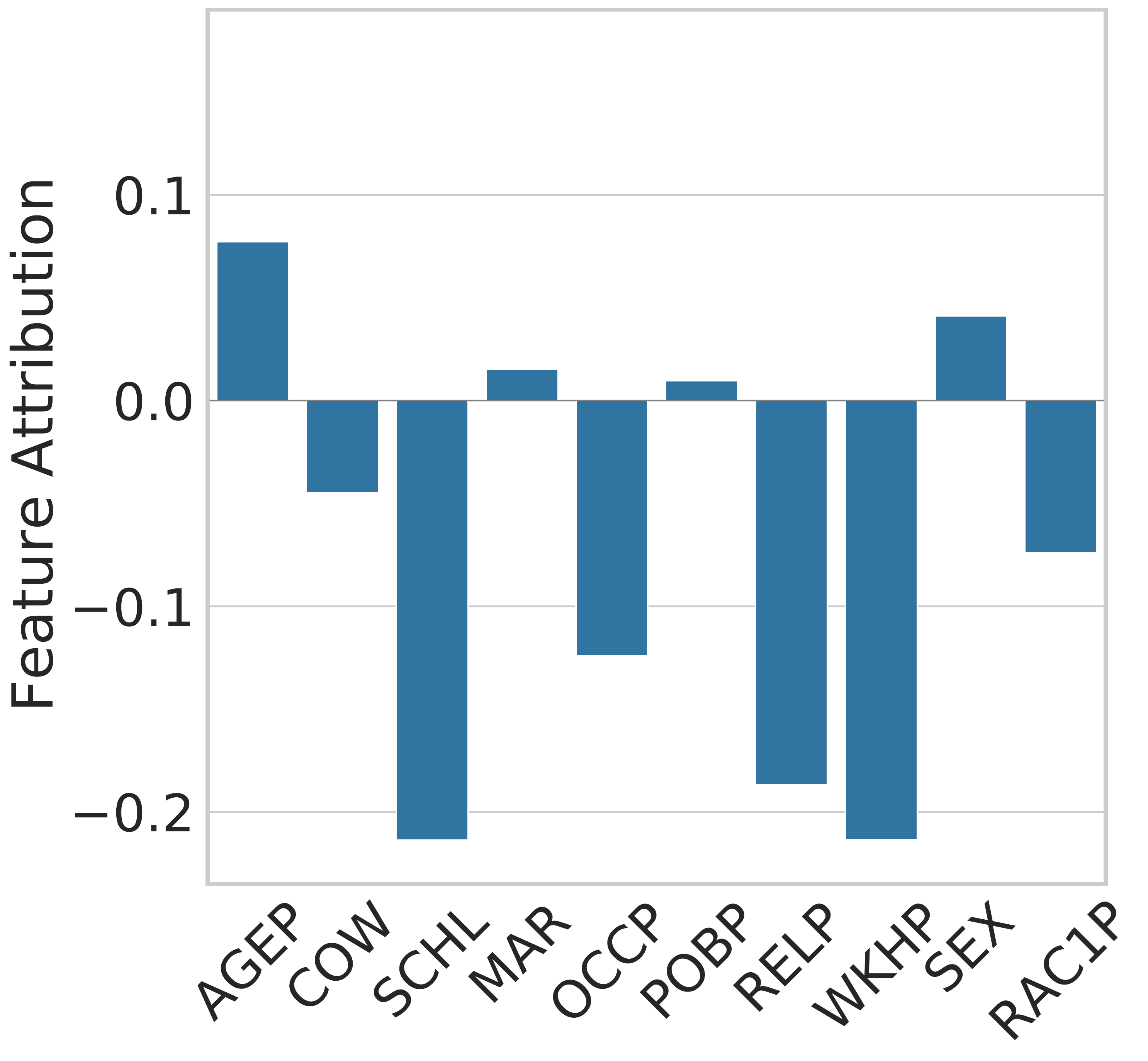}
         \label{fig:different_explanations_shap}
         \caption{SHAP}
     \end{subfigure}
     \hfill
     \begin{subfigure}[b]{0.23\textwidth}
         \centering
         \includegraphics[width=\textwidth]{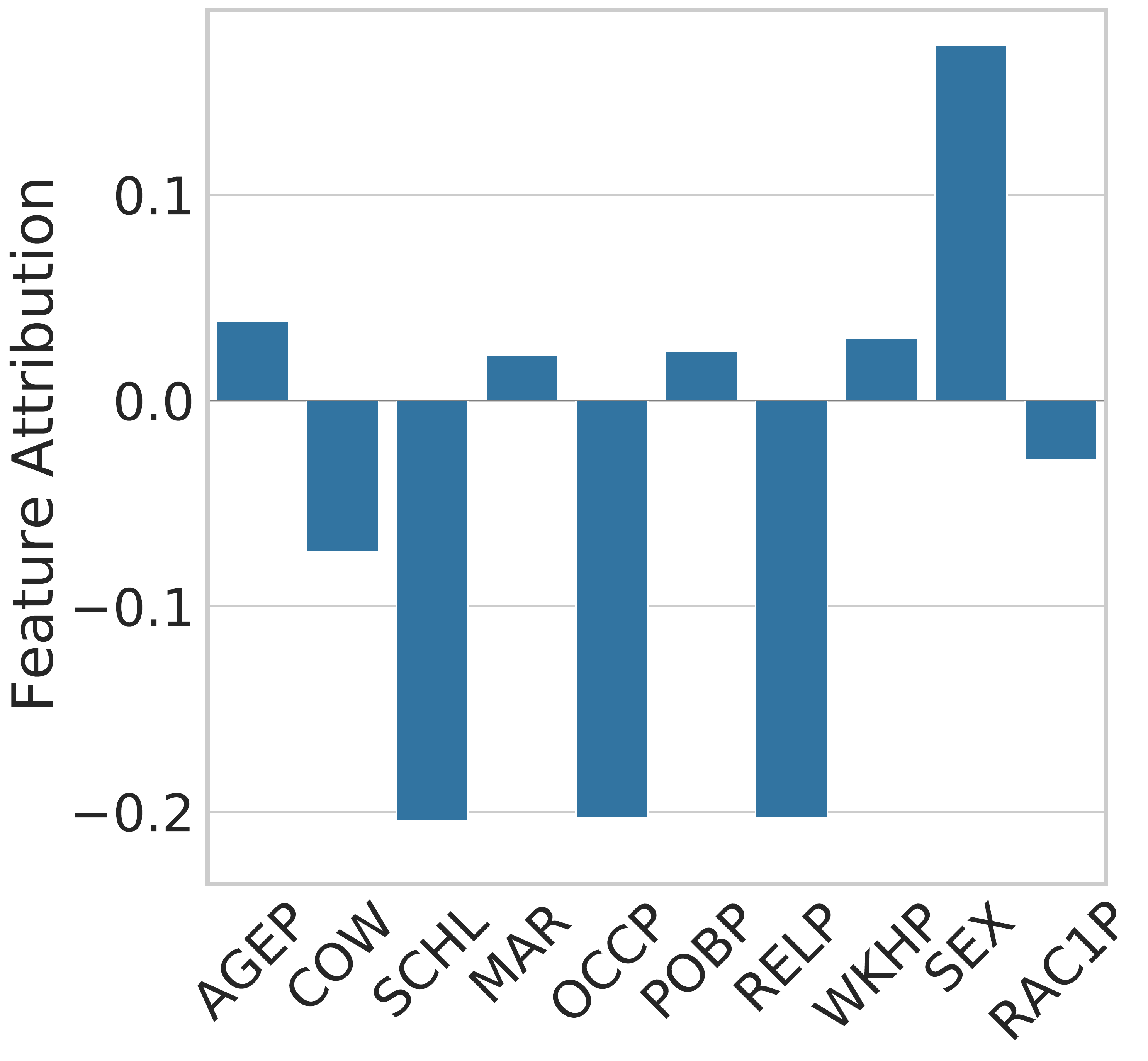}
         \label{fig:different_explanations_lime}
         \caption{LIME}
     \end{subfigure}
     \hfill
        \caption{For any given datapoint, different explanation algorithms might lead to very similar or completely different explanations. In many cases, however, there are both similarities and dissimilarities. The Figure depicts the SHAP and LIME feature attributions for a datapoint in the folktables ACSIncome prediction task \citep{ding2021retiring}: Are these attributions similar or different? More figures showing results for other data points can be found in the supplement. }
        \label{fig:similarities-and-dissimilarities}
\end{figure}

That different explanation algorithms lead to different explanations is also true for counterfactual explanation methods \citep{WachterEtal17,mothilal2020explaining}. Indeed, there is a variety of ways in which the optimization problem can be set up, which in turn leads to different explanations. However, already a single counterfactual explanation method can lead to a large number of counterfactual explanations. In a cooperative context, being able to generate many different counterfactual explanations for the same individual can be beneficial \citep{mothilal2020explaining}. In an adversarial context this is problematic because there is no principled way to choose among different counterfactual explanations, and the adversary is again awarded considerable discretion to determine explanations. In realistic, high-dimensional applications, the number of potential counterfactual explanations can quickly become very large. Let us illustrate this point on the German Credit Dataset. The German Credit Dataset is a 20-dimensional dataset with features on credit history and personal characteristic. The task is to predict credit risk in binary form. How many different counterfactual explanations exist for a single individual? With a common black-box decision function, more than 100 different counterfactual explanations exist for each individual.\\

At its core, the fundamental difficulty of explainable machine learning is then the same as in other fields of unsupervised learning: the lack of a ground truth explanation impedes the development of an algorithmic framework to automatically evaluate explanations. Every explanation algorithm needs to make assumptions about which properties of the decision function it seeks to highlight. As a result, it is possible to develop sanity checks for explanation algorithms and exclude unreasonable approaches \citep{AdebayoEtal18, camburu2019can}, but not to discern whether any of two post-hoc explanations is ``more correct'', which would be equivalent to discussing whether any of two different clusterings is ``more correct'' \cite{LuxWilGuy12} .

\subsection{The explanation provider can choose between a large number of possible explanation algorithms and parametrizations }
\label{sec-cs-parameters}

\begin{figure*}[t]
     \centering
     \begin{subfigure}[b]{0.24\textwidth}
         \centering
         \includegraphics[width=\textwidth]{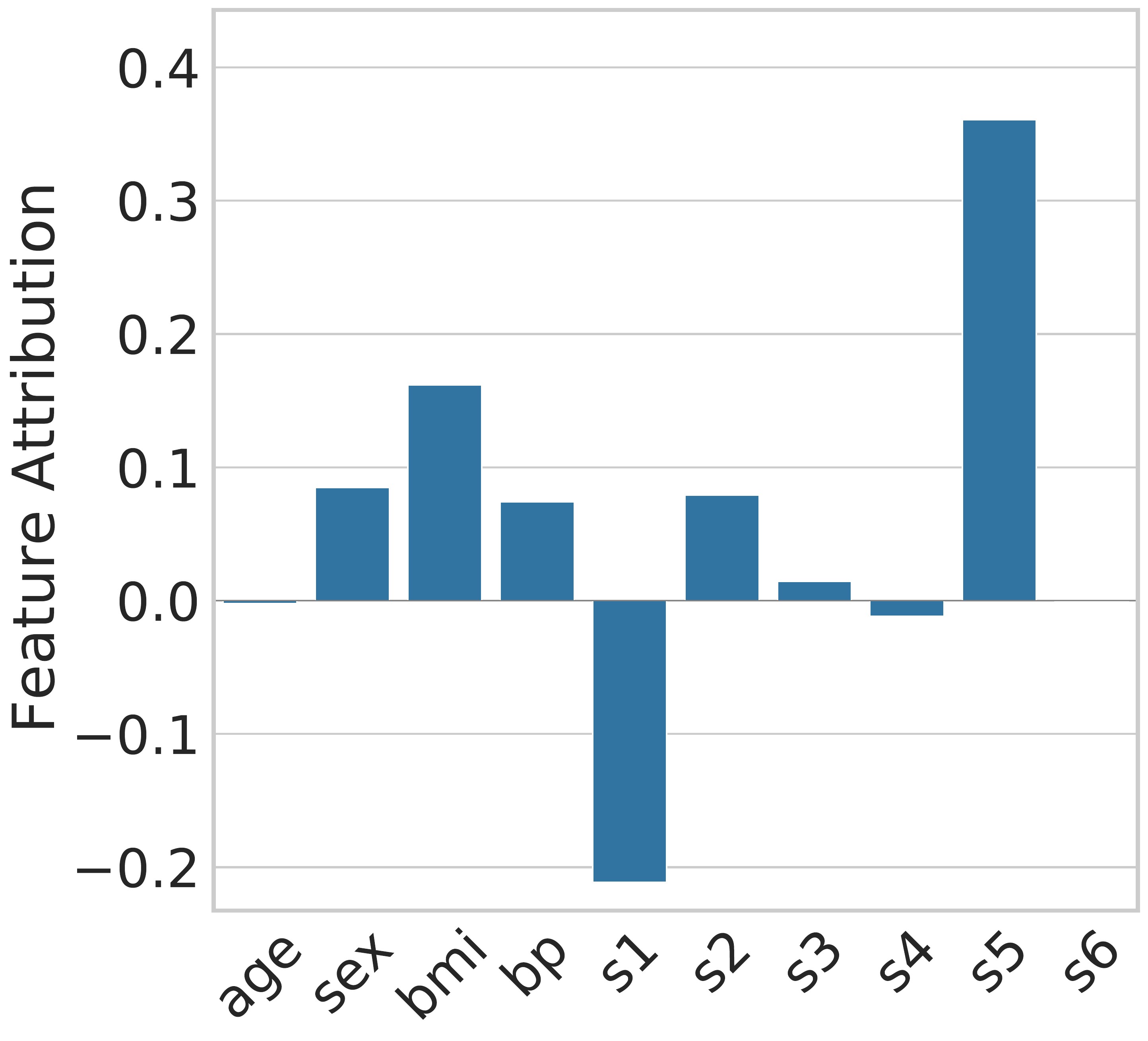}
         \caption{Diabetes, Lin. Regr.}
         \label{fig:diabetes_linear}
     \end{subfigure}
     \hfill
     \begin{subfigure}[b]{0.24\textwidth}
         \centering
         \includegraphics[width=\textwidth]{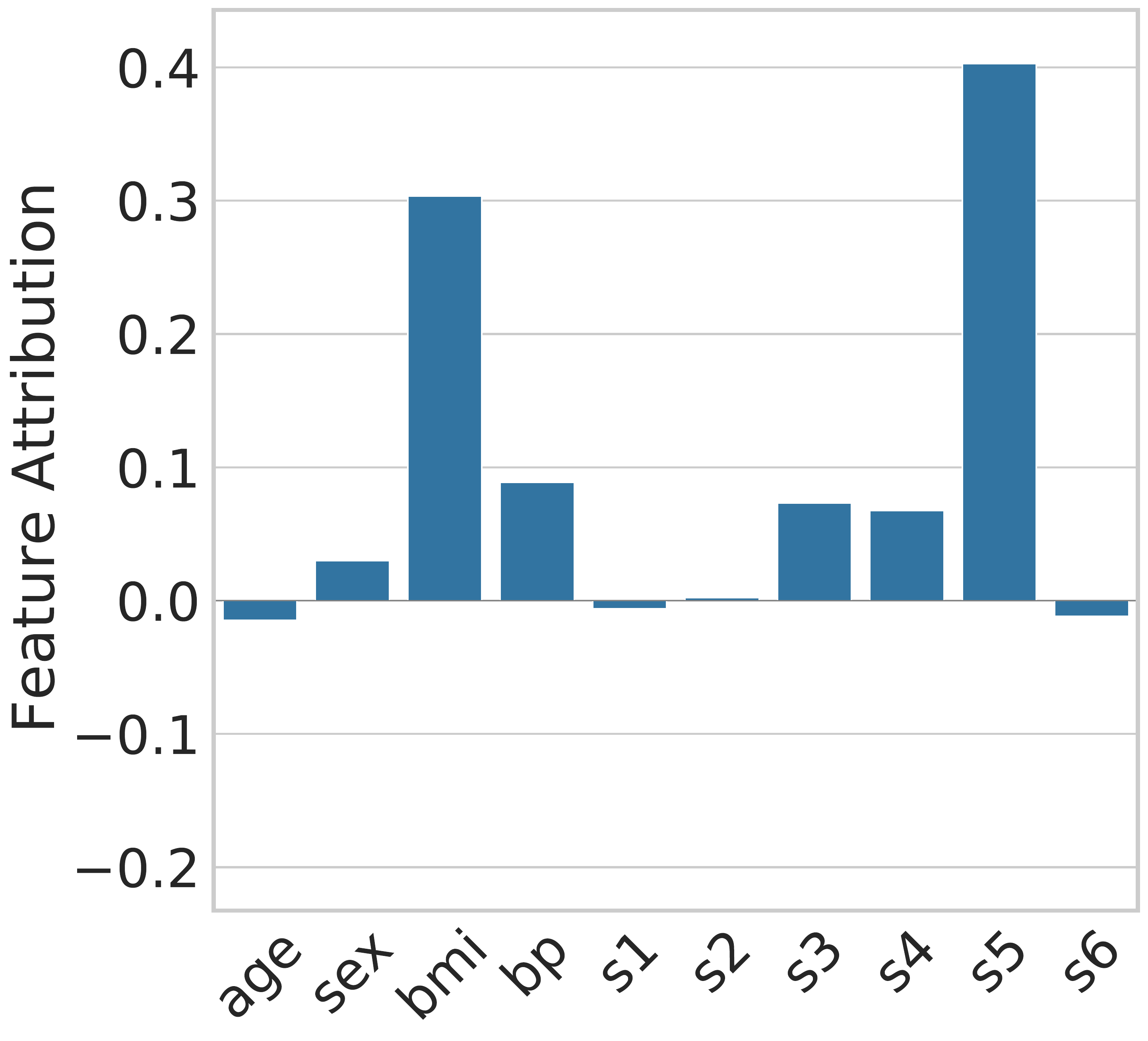}
         \caption{Diabetes, Random Forest}
         \label{fig:diabetes_shap}
     \end{subfigure}
     \hfill
     \begin{subfigure}[b]{0.24\textwidth}
         \centering
         \includegraphics[width=\textwidth]{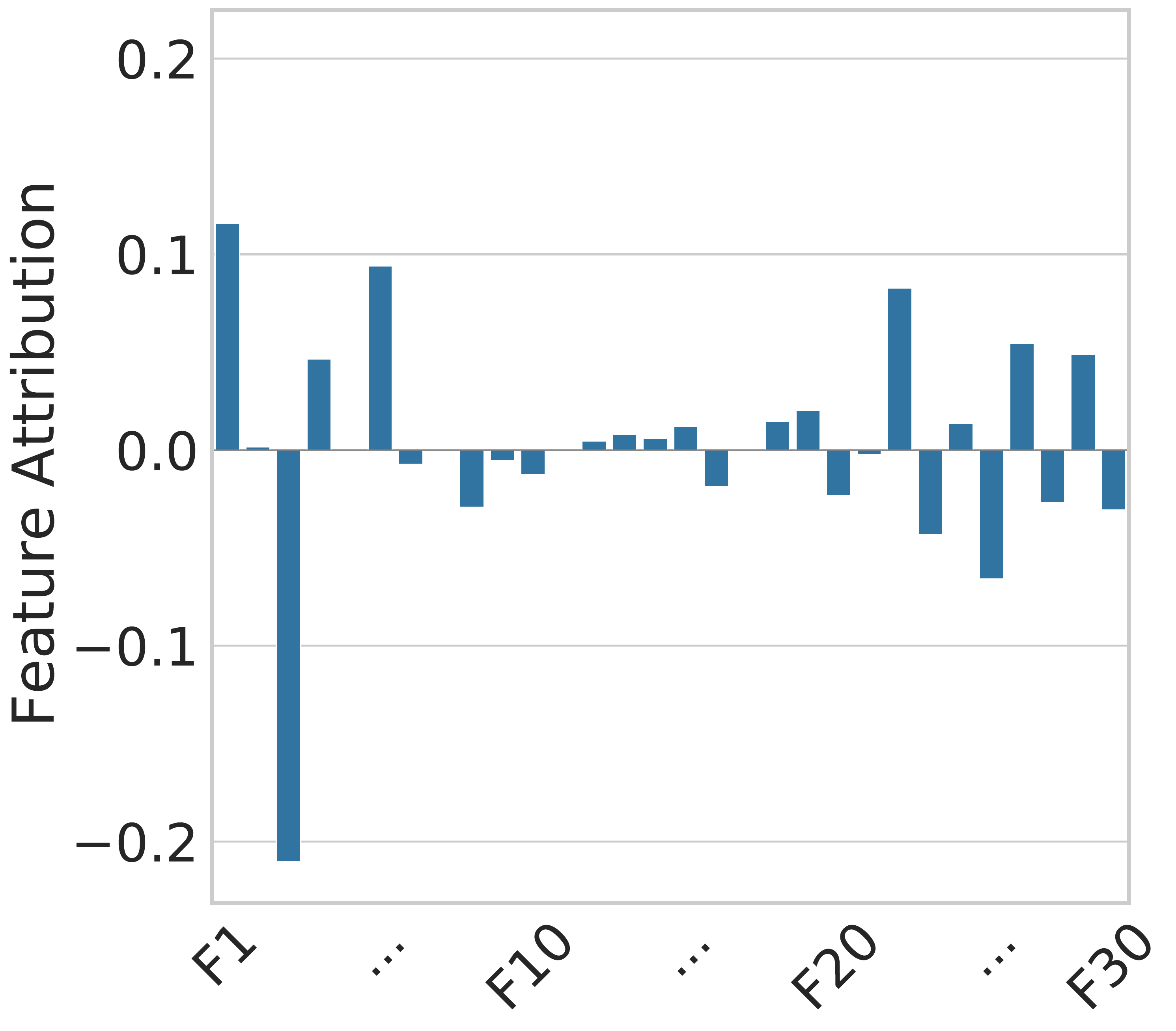}
         \caption{Cancer, 36\% Accuracy}
         \label{fig:cancer_random}
     \end{subfigure}
    \hfill
    \begin{subfigure}[b]{0.24\textwidth}
         \centering
         \includegraphics[width=\textwidth]{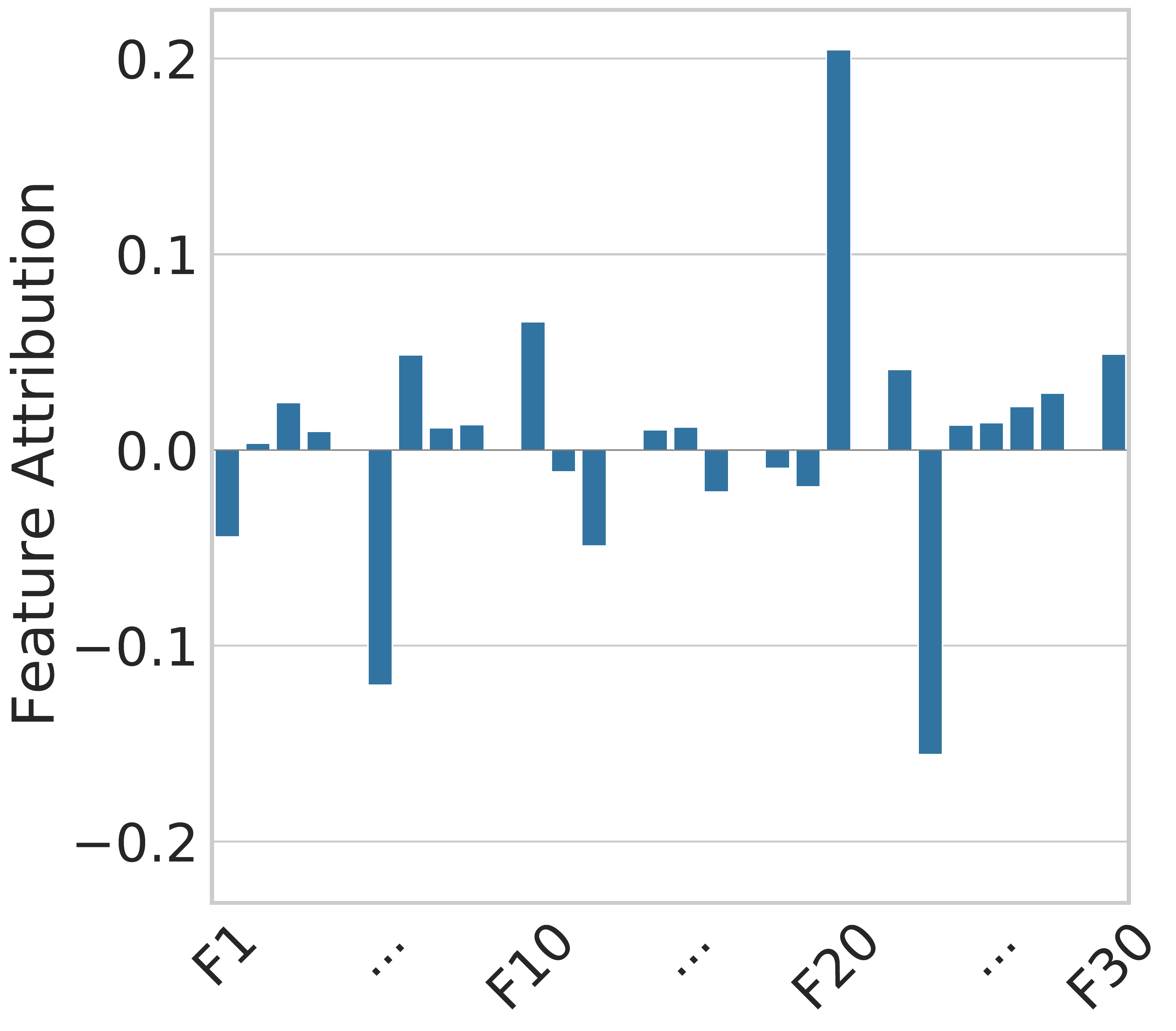}
         \caption{Cancer, 96\% Accuracy}
         \label{fig:cancer_95}
    \end{subfigure}
        \caption{Explanations depend on the exact shape of the classifiers high-dimensional decision boundary. Panel (a) and (b): On the diabetes dataset, linear regression and a random forest agree for 94\% of their predictions. Shown are the SHAP explanations on a data point where the prediction of both methods agree. As we can see, the explanations differ. Panel (c) and (d): the dependence on the decision boundary is subtle. It can even be hard to tell from the explanations whether the classifier had been trained trained at all. On the Wisconsin Breast Cancer dataset,  the SHAP explanations of a classifier trained to achieve an accuracy of 96\% are hard to distinguish from those of the same classifier trained on random labels. More figures showing results for other data points can be found in the supplement. }
        \label{fig:decision-boundary}
\end{figure*}

Even for a single explanation algorithm, there can be many different parameter choices that all lead to different explanations. LIME explanations, for example, depend on the bandwidth and the number of perturbations \citep{GarLux20,LeeEtal19,SlackEtal21a}. The uniqueness properties of Shapley values non-withstanding, there is a multiplicity of ways in which Shapley values can be operationalized to generate explanations \citep{sundararajan2020many}. Counterfactual explanation algorithms depend on the underlying metric chosen to represent closeness (e.g. Euclidean distance vs. $L1$-norm)\footnote{This originates in the philosophical account: counterfactuals depend on the way one measures proximity between facts and alternative counter-facts \cite{Lewis1973}.} as well as additional hyperparameters to weight-off between closeness and prediction, and, at least in principle, any number of additional penalty terms \cite{mothilal2020explaining}.  
In certain cases, it might be possible to come up with good default parameter choices. For example, recent work has demonstrated how to choose the bandwidth parameter of LIME in a principled way or quantify uncertainty in the resulting explanations \citep{ZhangEtal19,LeeEtal19,SlackEtal21a}. It is also possible to exclude explanation algorithms and parametrizations that are completely unreasonable, for example because they are not sensitive to the decision function \citep{AdebayoEtal18, camburu2019can}. This nevertheless leaves an ever-increasing number of plausible explanation algorithms and corresponding parametrizations. Quite generally, different explanation algorithms vary among many different dimensions, and there is an ever increasing number of suggestions as to how black-box functions might be explained. This can be seen, for example, in the recent work of \citet{covert2021explaining}, who summarize 25 existing methods in a unified framework. As already discussed above, there are no fundamental reasons that impede us from using any particular method.\footnote{The distinction between two ``different'' explanation algorithms and different parameter choices for the ``same'' explanation algorithm is of course a matter of perspective: We might consider the question of distributional versus intervential Shapley values as a question of how to use ``the'' SHAP method \cite{janzing2020feature}, but we might as well perceive it as a discussion as to which of two different methods to use.}

\subsection{Explanations depend on the exact shape of the high-dimensional decision boundary }
\label{sec-cs-decision-boundary}

\begin{figure*}[t]
     \centering
     \begin{subfigure}[b]{0.3\textwidth}
         \centering
         \includegraphics[width=0.7\textwidth]{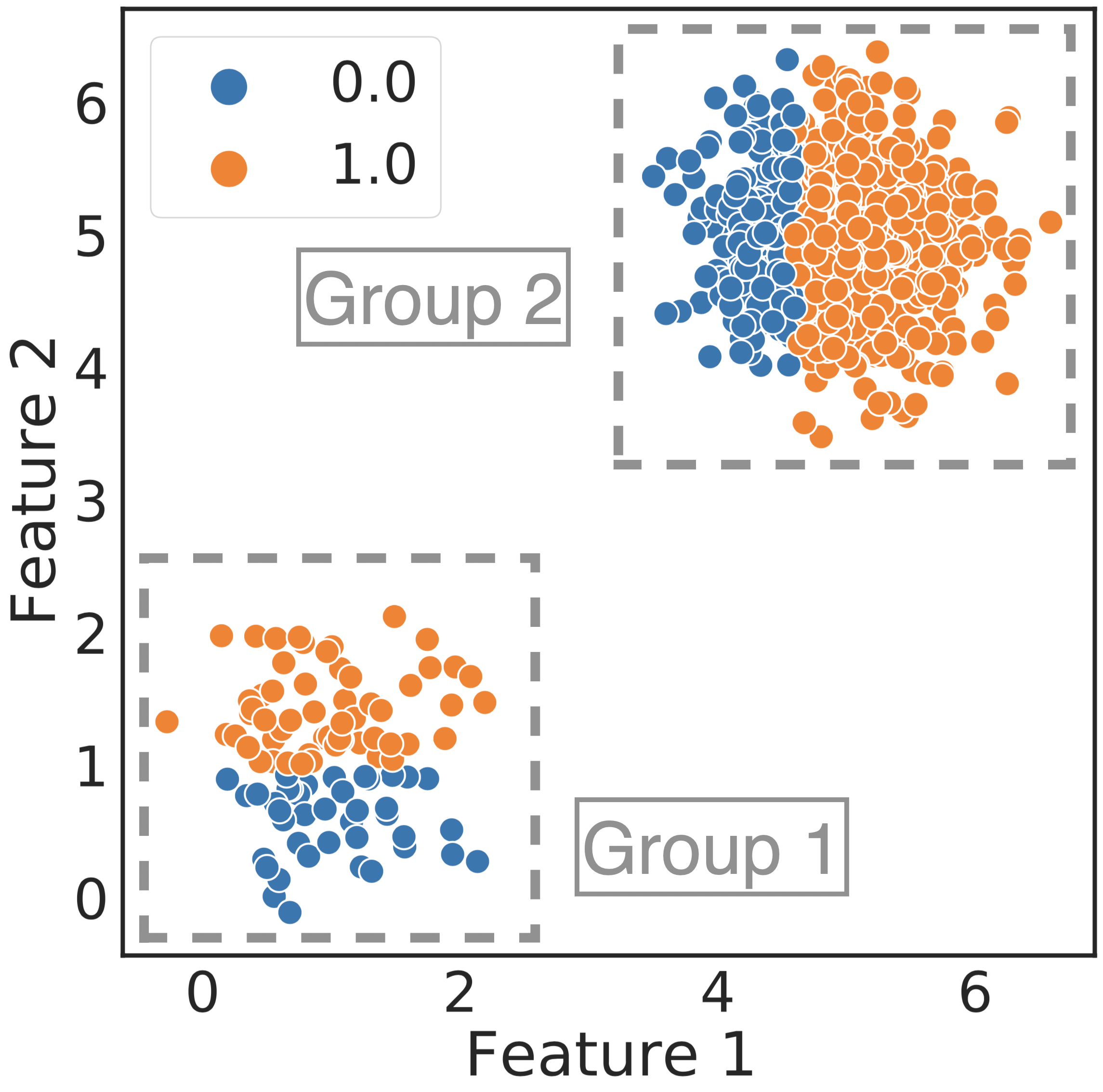}
         \caption{Dataset}
         \label{fig:two_groups_dataset}
     \end{subfigure}
     \hfill
     \begin{subfigure}[b]{0.3\textwidth}
         \centering
         \includegraphics[width=0.8\textwidth]{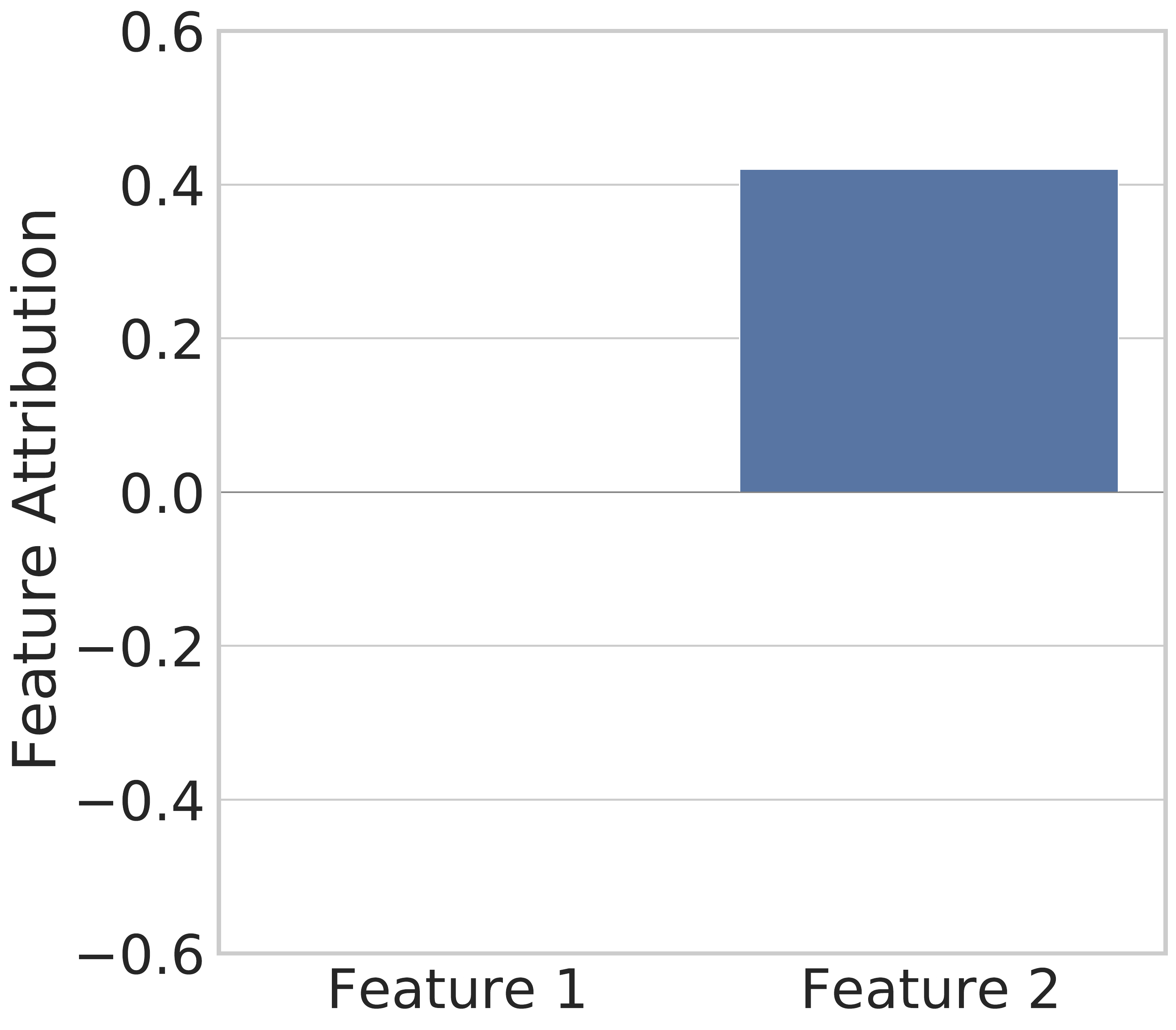}
         \caption{Reference data set: Group 1 only}
         \label{fig:only_first_group}
     \end{subfigure}
     \hfill
     \begin{subfigure}[b]{0.3\textwidth}
         \centering
         \includegraphics[width=0.8\textwidth]{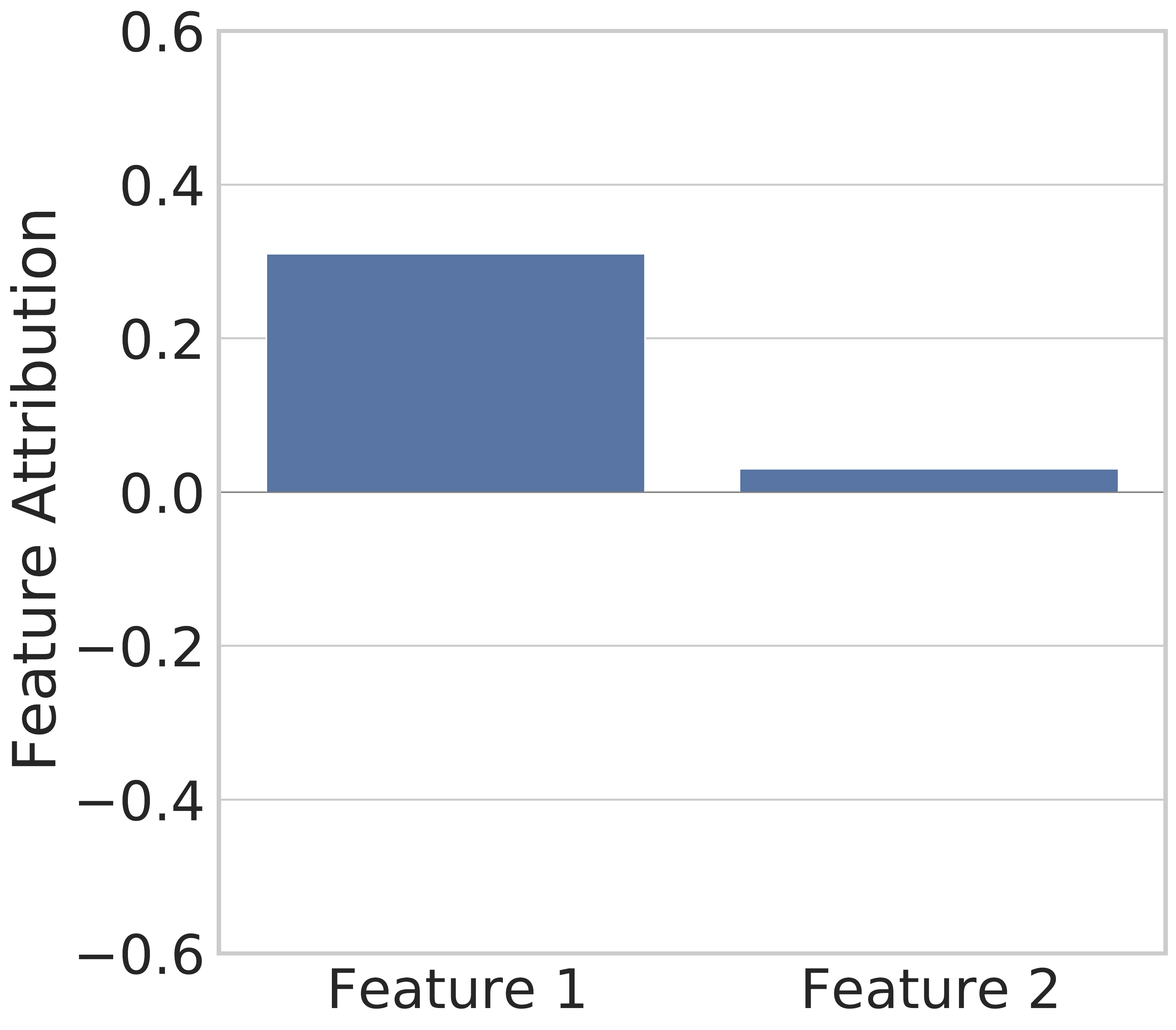}
         \caption{Reference data set: entire dataset}
         \label{fig:with_second_group}
     \end{subfigure}
        \caption{A simple toy example of how the choice of the explanation's reference dataset can influence the resulting explanations. The dataset in Panel (a) consists of two different population groups. The blue and orange color depicts the binary label that the classifier is supposed to predict at each data point (to get an intuition, you might think of the groups as ``male'' and ``female'', and the label as ``is awarded the credit'' or ``is not awarded the credit'').  
        Panels (b) and (c) depict the interventional SHAP feature attributions \citep{janzing2020feature} for the {\it same} data point in Group 1. In Panel (b), the explanation's reference dataset consists of the observations of Group 1 only. In Panel (c), the reference dataset is the entire dataset. The example shows that changing the reference dataset can almost completely change the feature attribution from one feature to another.}
        \label{fig:explanations_dataset}
\end{figure*}

Even if we fix a particular explanation method and its parameters, the generated explanations still depend on the {\it exact} shape of the learned decision boundary. In high dimensions, there are often many different black-box functions that solve a particular classification problem to a desired accuracy, that is they represent the data sufficiently well.  However, these functions often lead to different explanations. To a certain extent, we may say that the exact shape of the learned decision boundary is arbitrary, but since the explanations depend on it, these turn out to be arbitrary as well. One of the reasons for the sensitivity of the explanation to the function's shape is that many explanation methods evaluate the function $f$ at datapoints that are outside the data distribution or at points that are unlike most points from the data distribution. In the adversarial scenario, this is problematic because \textit{the adversary can freely modify the values of the function $f$ outside the data distribution} without changing the classification behavior. Recent work has demonstrated that this property can be used to explicitly manipulate and attack explanation methods \citep{SlackEtal20, SlackEtal21}. But even without explicit attacks, there are many different choices, in particular hyperparameter and architecture choices, that influence the shape of the decision boundary, and thus the resulting explanations. For an external examiner, this presents a challenging problem: while certain explicit attacks on explanation methods could in principle be detected through code review (see also Section \ref{sec:transparent-scenario}), it is far less clear how one would argue about choosing one classifier over another, or any particular choice of hyperparameters. This problem is illustrated in Figure \ref{fig:decision-boundary}. Here, we solved the same machine learning problem both with linear regression and a random forest. The two methods have comparable performance on the test set, where 94\% of their predictions agree. Nevertheless, the explanations obtained for the two different decision functions can be quite different -- even for points that receive the same prediction.\\

Turning to counterfactual explanations, it is well-known that these depend on the exact shape of the decision boundary. Let us give an example, again using the German Credit Dataset. Consider two different decision functions, a gradient boosted tree and logistic regression. If we generate a number of diverse counterfactual explanations \citep{mothilal2020explaining} for a typical individual with respect to one decision function, are these also counterfactual explanations with respect to the other decision function (at least as long as both functions arrive at the same decision)? In this simple experiment less than 50\% of counterfactual explanations that work for the gradient boosted tree also work for logistic regression. As discussed above, the fact that the explanations depend on the exact shape of the decision boundary is problematic because it allows the creator of the system to influence the resulting explanations. The particular choice of the decision function can even determine whether certain types of counterfactual explanations exist at all. Let us give an example on the Wisconsin Breast Cancer Dataset. To demonstrate the dependence on the decision boundary, we consider again two different decision functions, linear regression and a random forest. For linear regression, there exist a large number of counterfactual explanations that modify only a single variable. For the random forest, it is impossible to find any such counterfactual explanations. This is despite the fact that both classifiers exhibit similarly low test error.\\

\subsection{It is unclear how to choose the reference dataset that many explanations depend on}
\label{sec-cs-reference-dataset}

In recent years, there has been an increased focus on the composition of datasets, for example on the representation of different sociodemographic groups in machine learning datasets \citep{paullada2021data,BarHarNar19}. In many real-world problems such as credit lending, the criteria for choosing an appropriate dataset are not clear. %
In both cooperative and adversarial contexts, the creator of the system has to make numerous choices, many of which can have significant effects on both the shape of the learned decision boundary and the generated explanations. For example, \citet{AndersEtal20} have shown that gradient-based explanations can be manipulated by adding additional variables to the dataset. In this section, we highlight the additional role that the dataset can have on algorithmic explanations, even when \textit{keeping the learned decision boundary constant}. Indeed, while some explanation algorithms such as LIME only rely on the learned decision boundary, other methods such as SHAP and some counterfactual explanation methods make additional use of the data in order to generate explanations. The relevant dataset could be the training data, but it could also be a different dataset. We refer to it as the \textit{reference dataset}. While the usage of such a dataset to generate explanations can be seen as a remedy to the vagaries of high dimensions, or as a possibility to generate counterfactual explanations that look like they come from the data, this approach is problematic as long as the adversary determines the composition of the dataset. The reason is that whether certain datapoints are included in the dataset or not can determine whether an explanation algorithm provides one or another explanation. Figure \ref{fig:explanations_dataset} illustrates this with a simple example: By deciding between two different reference datasets, one can effectively decide whether one ore another feature was relevant to the decision. %

\subsection{Bottom line: Post-hoc explanations are highly problematic in an adversarial context}

It is extremely important to understand that an explanation algorithm is based on many human choices that are shaped by human objectives and preferences. While many choices are plausible, there is no objective reason to prefer one algorithm over the other, or one explanation over the other. Apart from the explanation algorithm and its particular parameters, explanations are influenced by human choices such as the selection of the classifier and the composition of the dataset. In adversarial contexts it implies that the adversary can choose, among many different plausible explanations, one that suits their incentives. 
This complicated situation makes it particularly difficult for external observers, including judges and regulatory bodies, to determine whether an explanation is acceptable. Explanation algorithms appear to provide objective explanations, yet as explained above this is not the case (compare Section \ref{sec-explanations-algorithms-own-world}).

\section{Once an examiner is allowed to assess the provided post-hoc explanations, she'd better investigate the decision function directly} 
\label{sec-testing}

So far we have discussed explainability obligations in European Union law and their motivation (Section \ref{sec-legal}), and 
pointed out theoretical (Sections \ref{sec-algorithm-incomplete}-\ref{sec-explanations-algorithms-own-world}) and practical (Sections \ref{sec-cs-different-algorithms-different-explanations}-\ref{sec-cs-reference-dataset}) shortcomings of post-hoc explanations.
In this section, we add yet another component to our argument. 
In an adversarial setting, it is not only the AI decision system itself but also the corresponding explanation algorithm which might need to be examined by a third party. Even if the examiner only attempts to assess the most basic consistency properties of the provided explanations, that is to check whether the explanations relate to the AI decision system at all, this  necessarily requires that the examiner is able to query the AI system. But then, the  explanations become entirely redundant: Rather than relying on explanations to enable risk management, provide trust or bias and discrimination detection (compare Section \ref{sec-rationales-behind-norms}), the examiner could directly query the AI system for problematic decision behavior. Because the creator of the system and the examiner have competing interests, it is important to distinguish degrees of transparent interaction between the two. Naturally, the examiner would like to have access to as much information as possible, whereas the adversary creator wants to disclose as little information as possible. We distinguish between a minimal and a fully transparent scenario of information disclosure (Sections \ref{sec:minimalist-scenario}-\ref{sec:transparent-scenario}).

\subsection{Minimalist scenario where decision function and explanation algorithm can be queried}
\label{sec:minimalist-scenario}

To determine whether the adversary's explanations actually correspond to the used decision function $f$ instead of being arbitrary justifications not related to the decision process, the examiner needs to be able to query the decision function and the generated explanations.\footnote{This means that for any possible datapoint (or individual) $x$, the examiner is allowed to ask the adversary: ``For this hypothetical datapoint $x$, what would be the decision $y=f(x)$, and what would be the corresponding explanation $E(x,y)$? The adversary would then privately compute both quantities and make them available to the examiner, but not tell the examiner how the computation was performed.}  This includes a fair amount of related knowledge, such as which variables are input to the algorithm, but excludes explicit access to the decision function, explanation algorithm, source code and training dataset. A related but slightly more limited version of this scenario arises when individuals jointly collect the decisions and explanations from the creator of the system. %
In this \textit{minimalist scenario}, the examiner can validate the internal consistency of the provided explanations. Researchers have proposed a number of criteria that the examiner can test for such as faithfulness to the model, robustness to local perturbations, as well as necessity and sufficiency notions for individual feature attributions \citep{AdebayoEtal18, kommiya2021towards, Vilone2021}. The examiner might also want to perform tests as to whether the provided explanations have been manipulated \citep{SlackEtal21}. More importantly however, even just with the ability to query the decision function, the examiner can ignore the explanations and directly investigate the decision function for problematic properties. For example, the examiner could conduct a systematic evaluation of, say, fairness metrics such as equal opportunity and demographic parity, based on an independent reference dataset of her choice (see \cite{BarHarNar19} for these and other notions of fairness and discrimination). Indeed, %
because the adversary designing the explanation algorithm has no interest in choosing explanations that highlight any discriminatory behavior of the decision algorithm, the examiner is well-advised to simply ignore the explanations and test the decision algorithm directly. %
Although such tests might be similar to certain explanation algorithms, what is important is that the examiner (as opposed to the creator) designs and implements them. Note that we are {\em not} saying that the minimalist scenario actually allows the examiner to assess all legally relevant properties of the decision function. What exactly can be assessed with querying access is a question that still requires more research. Our point is that once we have querying access, the explanations are useless.

\subsection{Fully transparent scenario where algorithms' source code and training data are disclosed}
\label{sec:transparent-scenario}

At the opposite end of the minimalist scenario is the {\em fully transparent scenario} where the examiner is allowed to investigate the decision function, source code and training data. An examiner could then scrutinize whether the explanation algorithms have been implemented according to the state of the art with sensible parameter choices. This directly rules out the possibility for  the creator of the system to manipulate explanations. Are post-hoc explanations useful in the transparent scenario, perhaps because the examiner now has the tools to verify whether the adversary has chosen the ``correct'' explanations? As we have already discussed above, the problem is that there is no notion of ``correct'' explanation (Sections \ref{sec-explanations-algorithms-own-world} and \ref{sec-cs-different-algorithms-different-explanations}). Thus, except for notions of internal consistency \citep{AdebayoEtal18,kommiya2021towards}, there is, in general, nothing the examiner can say about the explanations. Another issue, already observed in Sections \ref{sec-cs-decision-boundary} and \ref{sec-cs-reference-dataset}, are hyperparameter choices and decisions regarding the composition of the dataset. For these decisions, it is highly non-trivial to come up with uniquely reasonable defaults: If the adversary has found a particular neural network architecture with hyperparameters that generalize well on the adversary's own dataset, %
how exactly could the examiner argue that this is inappropriate? Nevertheless, all of these choices can influence the resulting explanations, even if we fix a particular explanation algorithm. Of course, the examiner could scrutinize the source code, re-train the system with different parameters, perform tests on the data, and generate alternative explanations. Some have argued that this might be sufficient in order to assess a variety of legal requirements \citep{kleinberg2018discrimination}. While we think that more research is needed on what can be realistically achieved in the fully transparent scenario, it is quite clear that the examiner can, at least in principle, perform a variety of powerful tests (whether this is achievable in practice, based on the limited resources of an examiner, is yet a different story). At any rate, just as in the minimalist scenario, the examiner is well-advised to examine and test the system on her own, and to ignore the explanations provided by the adversary creator.

\section{Discussion}
\label{sec-discussion}

Explainability is often praised as a tool to mitigate some of the risks of black-box AI systems. Our paper demonstrates that in adversarial contexts, post-hoc explanations are of very limited use.  From a technical and philosophical point of view these explanations can never reveal the ``unique, true reason'' why an algorithm came to a certain decision. In complicated black-box models, such a true reason simply does not exist. We moreover demonstrated that post-hoc explanations of standard decision algorithms on simple datasets possess a high degree of ambiguity that cannot be resolved in principle. For these reasons, post-hoc explanations of black-box systems are, to a certain degree, incontestable. In the best case, post-hoc explanation algorithms can point out some of the factors that contributed to a decision --- these algorithms are therefore useful for model debugging, scientific discovery and practical applications where all parties share a common goal. In adversarial contexts, in contrast, we demonstrated that local post-hoc explanations are either trivial or harmful. In the worst case, the explanations may induce us into falsely believing that a ``justified'', or ``objective'' decision has been made even when this is not the case. 
\\

It was also seen that it remains unclear how expectations of explainability in the GDPR or the AIA ought to be interpreted. The GDPR does not give rise to a general explainability obligation, and the draft AI Act currently would only require some degree of explainability in relation to high-risk applications of AI. We call on legislators to formulate related provisions with more specificity in order to create legal certainty in this respect. If the final version of the AIA requires a strong version of explainability for high-risk AI systems, black-boxes simply cannot be used: they cannot be explained directly, and the only indirect means of explaining them --- local post-hoc explanations --- are unsuitable. In this case, one would have to resort to the use of simple, inherently interpretable machine learning models rather than black-box models (compare \cite{Rudin19}) although this may impede innovations. We would expect that these algorithms and their explanations are more robust and less susceptible to manipulation, such that large parts of our criticism would not apply to inherently interpretable models. However, future research needs to clarify whether this is the case, because we are not aware of any research that investigates inherently interpretable machine learning in an adversarial setting. If, on the other hand, explainability in the final version of the AIA is to be understood as one of several means to achieve more transparency in machine learning, other methods than post-hoc explanations might be more suitable to achieve the desired goals of transparency. For example, as far as testing for biases and discrimination is concerned, it is unlikely that the creator of the system will choose to generate explanations that can be used to uncover hidden biases. But there is a much more direct route to assess discrimination than implicitly through explanations. Indeed, external examiners could directly test the system for discriminatory properties \cite{kleinberg2018discrimination}. As such, the external examination of black-boxes  may be a more suitable means of enabling more accountable AI systems.\\ 

The current draft of the AIA already requires documentation regarding the functioning of AI systems. However, one has to be aware of the versatile manipulation possibilities that lie in the development process of AI systems itself, through choice of training data, features, algorithms, parameters, and so on. Even in the fully transparent scenario where the entire development pipeline including the source code is open \cite{kleinberg2018discrimination}, a considerable leeway for manipulations remains. In order to address these, an external examiner would  need access to considerable manpower and resources. Even when training data and source code can in principle be examined, algorithms re-applied or even retrained, actually doing so for a system that has been developed by a large team might be very difficult if not impossible. More research is needed to understand exactly which legal objectives can be satisfied by such extended documentation of AI systems, or whether the documentation would again just serve as a means to provide an appearance of objectivity without any real value. \\

Overall, we believe that the question of testing and certifying  machine learning systems in an adversarial scenario is a research direction that is still heavily under-explored. There is no single way to achieve all the desired transparency and control goals for such AI systems. Even complete transparency, open code, open data might not lead to all the desired goals. For this reason, it is important to investigate in more detail what objective can be achieved by which means, and which goals might not be possible to achieve at all. Only then can we engage in a meaningful debate about responsible use of AI systems in social contexts. \\

Finally, we recall that our criticism of explainability, in particular local post-hoc explanations, concerns adversarial scenarios. In cooperative scenarios, many interesting discoveries might be made with the help of explainable machine learning.

\section{Funding Disclosure}

This work has been partially supported by the German Research Foundation
through the Cluster of Excellence “Machine Learning – New Perspectives for Science" (EXC 2064/1 number 390727645), the Baden-W{\"u}rttemberg Foundation (program ``Verantwortliche K{\"u}nstliche Intelligenz''), the BMBF Tübingen AI Center (FKZ: 01IS18039A), the 
International Max Planck Research School for Intelligent Systems (IMPRS-IS) and the Carl Zeiss Foundation. The authors declare no additional sources of funding and no financial interests.

\bibliographystyle{ACM-Reference-Format}
\bibliography{references}

\input{supplement}

\end{document}

%% file: ules_latex_defaults.tex
\usepackage{amsmath}
\allowdisplaybreaks[3] %
\usepackage{bbold} %

\usepackage{flafter} %
\PassOptionsToPackage{pdftex}{graphicx} %

\usepackage[pdftex]{graphicx}

\usepackage{geometry}
\usepackage{fancyhdr} %
\usepackage{ifthen} %
\usepackage{layout} 
\usepackage{xspace}%
\usepackage{afterpage}%
\usepackage{url} 
\usepackage{textcomp} %
\usepackage{pdfpages} %

\usepackage{wasysym} 

\usepackage[weather]{ifsym}

\def\ba#1\ea{\begin{align*}#1\end{align*}} %
\def\banum#1\eanum{\begin{align}#1\end{align}} %

\newcommand{\bi}{\begin{itemize}}
\newcommand{\ei}{\end{itemize}}
\newcommand{\be}{\begin{enumerate}}
\newcommand{\ee}{\end{enumerate}}
\newcommand{\bc}{\begin{center}}
\newcommand{\ec}{\end{center}}

\newcommand{\biq}{%
\begin{list}{$\bullet$}{%
\setlength{\topsep}{0cm}
\setlength{\partopsep}{-\parskip}
\setlength{\leftmargin}{0.8cm}
 \setlength{\itemsep}{0pt}
    \setlength{\parskip}{0pt}
    \setlength{\parsep}{0pt}    
}}
\newcommand{\eiq}{\end{list}}

\makeatletter
\@ifclassloaded{beamer}{}{}
\makeatother

\ifx\lemma\undefined
\newtheorem{theorem}{Theorem} 
\newtheorem{lemma}[theorem]{Lemma}

\else
\fi

\newcommand{\ulesqed}{\hfill\smiley}

\ifx\proof\undefined
\newenvironment{proof}{\par\noindent{\em Proof. }}{\ulesqed\\[2mm]}
\else
\fi

\input{ules_latex_colors.tex}

\usepackage[noend]{algorithmic}

\algsetup{linenosize=\scriptsize\slidegray}
\algsetup{linenodelimiter=}

\newcommand{\balg}{\begin{algorithmic}[1]}
\newcommand{\ealg}{\end{algorithmic}}

\newcommand{\RR}{\mathbb{R}}

\let\R\undefined %
\newcommand{\R}{\RR}

\renewcommand{\epsilon}{\ensuremath{\varepsilon}}
\renewcommand{\phi}{\ensuremath{\varphi}}

\wasyfamily
\DeclareFontShape{U}{wasy}{b}{n}{ <-10> ssub * wasy/m/n
<10> <10.95> <12> <14.4> <17.28> <20.74> <24.88>wasyb10 }{}

\DeclareFontShape{U}{wasy}{m}{n}{ <5> <6> <7> <8> <9> gen * wasy
<10> <10.95> <12> <14.4> <17.28> <20.74> <24.88> <35> <40> <50> <60> wasy10  }{}

\newcounter{ulestheorem}

\newcommand{\ulestheorem}[3]{%
\refstepcounter{ulestheorem}%
\begin{ulesblock}{%
Theorem \arabic{ulestheorem}
\label{#1}%
\ifthenelse{\equal{#2}{}}{}{(#2)}}%
#3 %
\end{ulesblock}%
}

\newcommand{\ulesproposition}[3]{%
\refstepcounter{ulestheorem}%
\begin{ulesblock}{%
Proposition \arabic{ulestheorem}
\label{#1}%
\ifthenelse{\equal{#2}{}}{}{(#2)}}%
#3 %
\end{ulesblock}%
}
\newcommand{\ulescorollary}[3]{%
\refstepcounter{ulestheorem}%
\begin{ulesblock}{%
Corollary \arabic{ulestheorem}
\label{#1}%
\ifthenelse{\equal{#2}{}}{}{(#2)}}%
#3 %
\end{ulesblock}%
}

%% file: ules_latex_colors.tex
\ifx\ulesred\undefined
\usepackage{color}

\definecolor{ulesred}{rgb}{0.65,0.18,0.21} %

\newcommand{\ulesred}{\color{ulesred}}

\definecolor{darkgreen}{rgb}{0,0.5,0.2}

\definecolor{verylightgray}{gray}{0.85} 
\definecolor{mediumgray}{gray}{0.5} 
\newcommand{\mediumgray}{\color{mediumgray}} %
\definecolor{darkgray}{gray}{0.4} 
\newcommand{\slidegray}{\mediumgray} %

\definecolor{lavender}{cmyk}{0,0.5,0,0}

\definecolor{darkblue}{rgb}{0,0,0.5}

\definecolor{orange}{rgb}{1,0.5,0}

\else

\fi

\ifx\black\undefined
\newcommand{\black}{\color{black}}
\fi

\ifx\blue\undefined
\newcommand{\blue}{\color{blue}}
\fi

\ifx\red\undefined
\newcommand{\red}{\color{red}}
\fi

\ifx\green\undefined
\newcommand{\green}{\color{green}}
\fi

\ifx\white\undefined
\newcommand{\white}{\color{white}}
\fi

\ifx\magenta\undefined
\newcommand{\magenta}{\color{magenta}}
\fi

%% file: supplement.tex
\onecolumn
\newpage
\appendix

\setcounter{figure}{0}                                      %
\renewcommand\thefigure{\thesection.\arabic{figure}}

\section{Post-Hoc Explanations Fail to Achieve their Purpose in Adversarial Contexts: Supplementary Materials}

\subsection{Code}

The python code to replicate all results in this paper is available at \url{https://github.com/tml-tuebingen/facct-post-hoc}.

\subsection{Datasets }
\label{apx-datasets-algorithms}

In our experiments, we used the following datasets.\\

{\bf Adult-Income.} This dataset contains information about individuals based on the 1994 US Census. It is available from the UCI machine learning repository. We obtained it from the SHAP package  \url{https://github.com/slundberg/shap}. The dataset contains the 12 features  age, workclass, education-num, marital status, occupation, relationship, race, sex, capital gain, capital loss, hours per week, country. In the figures, the features are numbered F1-F12 in this order. The machine learning problem is to predict whether whether an individual’s income is over \$50,000. We trained a gradient boosted tree which achieved a test accuracy of 87\%.\\

{\bf German Credit.} The German Credit Dataset is a  dataset with 20 different features on individual's credit history and personal characteristic. The machine learning problem is to predict credit risk in binary form. We obtained the dataset from the UCI machine learning repository. We trained a gradient boosted tree which achieved a test accuracy of 76\%. We also trained logistic regression which achieved a test accuracy of 74\%.\\

{\bf Folktables.} Folktables is a Python package that provides access to datasets derived from recent US Censuses \url{https://github.com/zykls/folktables}. We used this package to obtain the data from the 2016 Census in California. The machine learning problem is the ACSIncome prediction task, that is to predict whether an individual's income is above \$50,000, based on 8 personal characteristics. We trained a gradient boosted tree which achieved a test accuracy of 83\%.\\

{\bf Diabetes.} The Diabetes dataset is a dataset of diabetes patient records. It is available from the UCI machine learning repository. We obtained it from the scikit-learn machine learning library \url{https://scikit-learn.org}. The dataset contains 10 features about each individual at baseline: age, sex, body mass index, average blood pressure, and six blood serum measurements. The machine learning problem is to predict disease progression one year after baseline. We converted the scalar outcome into a binary by thresholding at the median. We trained linear regression which achieved a test accuracy of 71\%. We also trained a random forest which achieved a test accuracy of 74\%.\\

{\bf Wisconsin Breast Cancer.} The Wisconsin Breast Cancer dataset is a tabular dataset with features of breast mass images. The dataset contains 30 features that describe the characteristics of the cell nuclei present in the image. The dataset is available from the UCI machine learning repository. We obtained it from the scikit-learn machine learning library \url{https://scikit-learn.org}. The machine learning problem is to predict the binary diagnosis (malignant/benign). We trained linear regression which achieved a test accuracy of 96\%. We also  trained linear regression on random labels which achieved a test accuracy of 36\%.\\

\subsection{Explanation Algorithms}

In our experiments, we used the following explanation algorithms.\\

{\bf SHAP} The SHAP algorithm was proposed by \cite{LundbergEtal2017}. We use it via the accompanying python package \url{https://github.com/slundberg/shap}. With (gradient boosted) trees, we use the exact computation method proposed in \cite{lundberg2020local}. With all other classifiers, we use the Kernel SHAP method. The approach by \citet{janzing2020feature} is also implemented in this package. Whenever available, we use parametrizations proposed in the documentaion of the package. \\

{\bf LIME} The LIME algorithm was proposed by \cite{RibeiroEtal16}. We use it via via the accompanying python package \url{https://github.com/marcotcr/lime}.  Whenever available, we use parametrizations proposed in the documentaion of the package. \\

{\bf DiCE} The DiCE algorithm was proposed by \cite{mothilal2020explaining}. We use it via the accompanying python package  \url{https://github.com/interpretml/DiCE}. To generate counterfactual explanations, we used the model-agnostic randomized sampling method. \\

\subsection{Figures}

To create the figures, we normalized the feature attributions to have $l1$-norm 1.

\subsection{Additional Figures}
\label{apx-figures}

The following pages contain additional figures. These follow the figures in the main paper and depict the first observations from the test set, so they are not hand-selected in any way. The reader might notice that we selected the figures in the main paper from these. Figures for all observations from the test are avaialble with the code that will be made available upon publication.

\newpage

\begin{center} 
\huge \bf Additional Figures Related to Figure \ref{fig:different_explanations} in the Main Paper
\end{center} 
\vspace{0.4cm}

\begin{figure*}[h]

\begin{tabular}{cccc}

\includegraphics[width=0.235\textwidth]{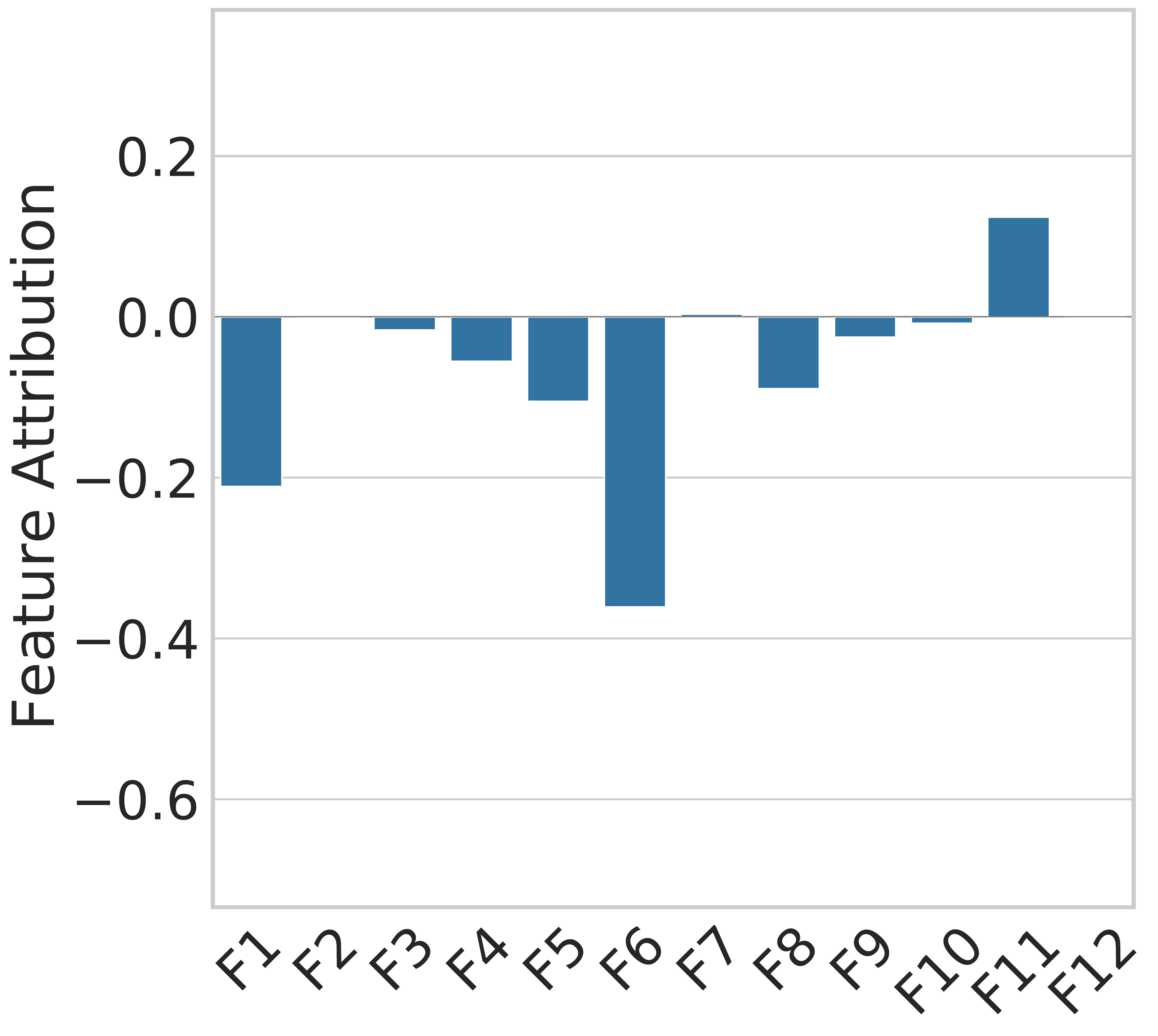} &
\includegraphics[width=0.235\textwidth]{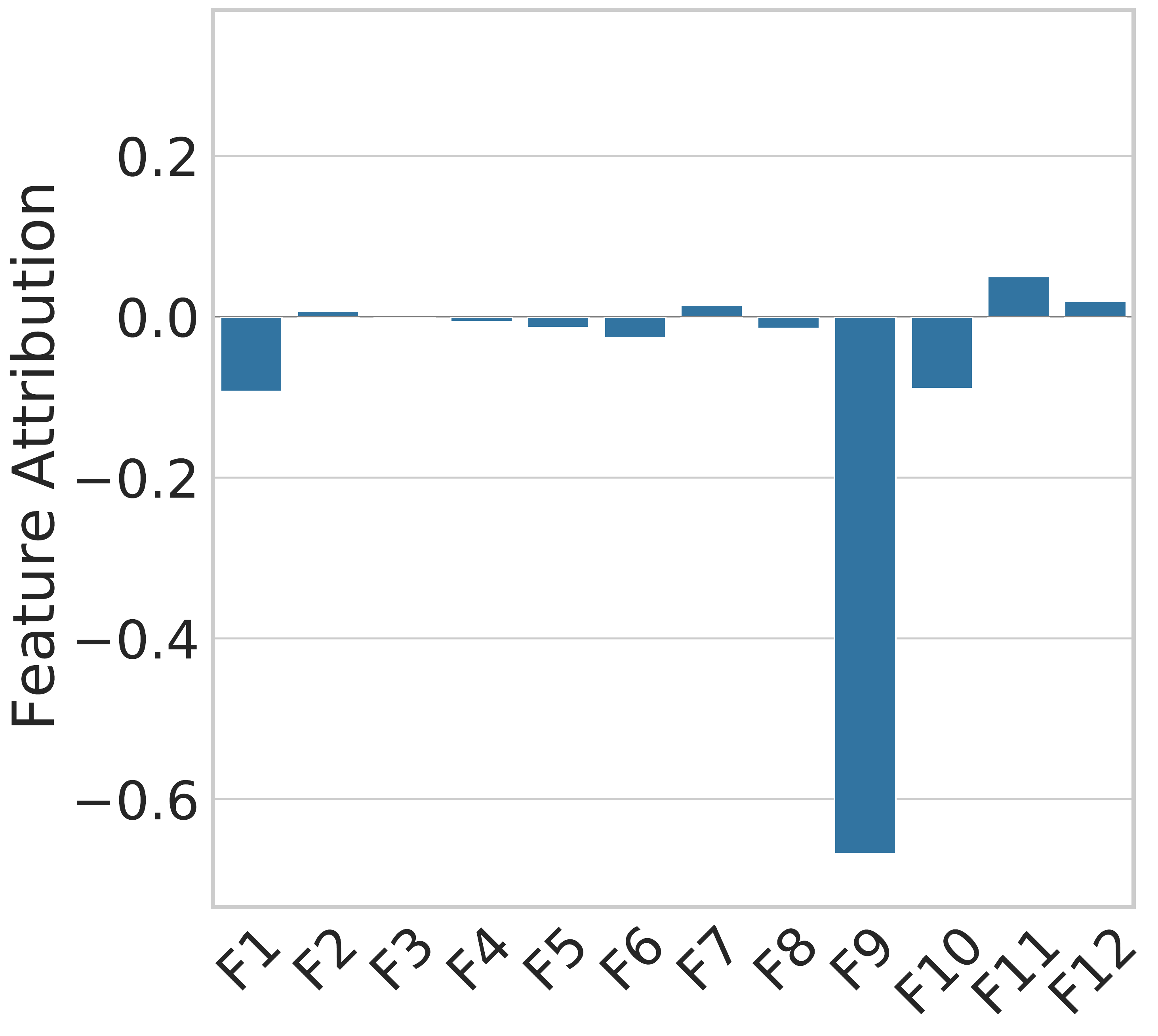} &
\includegraphics[width=0.235\textwidth]{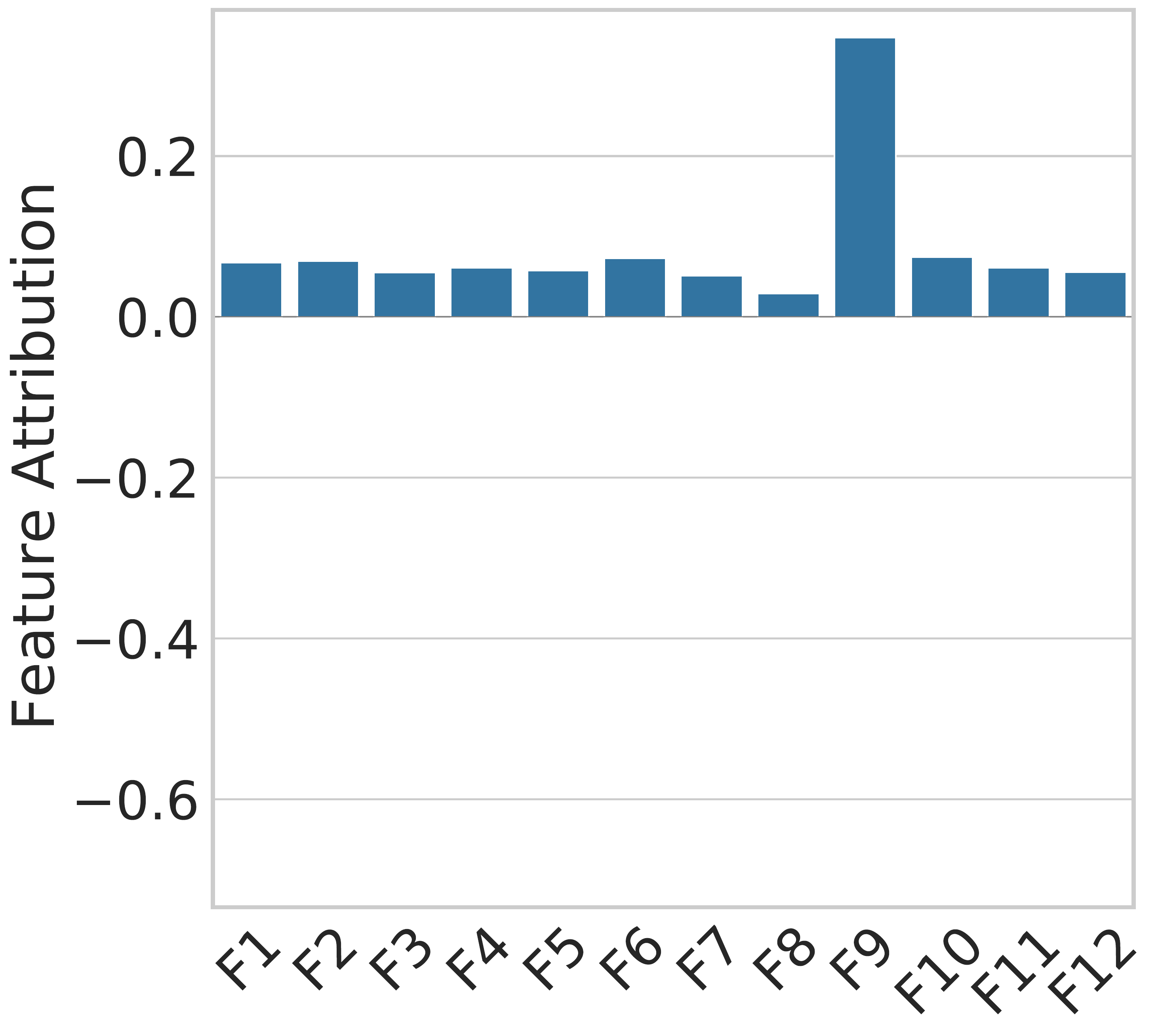} &
\includegraphics[width=0.235\textwidth]{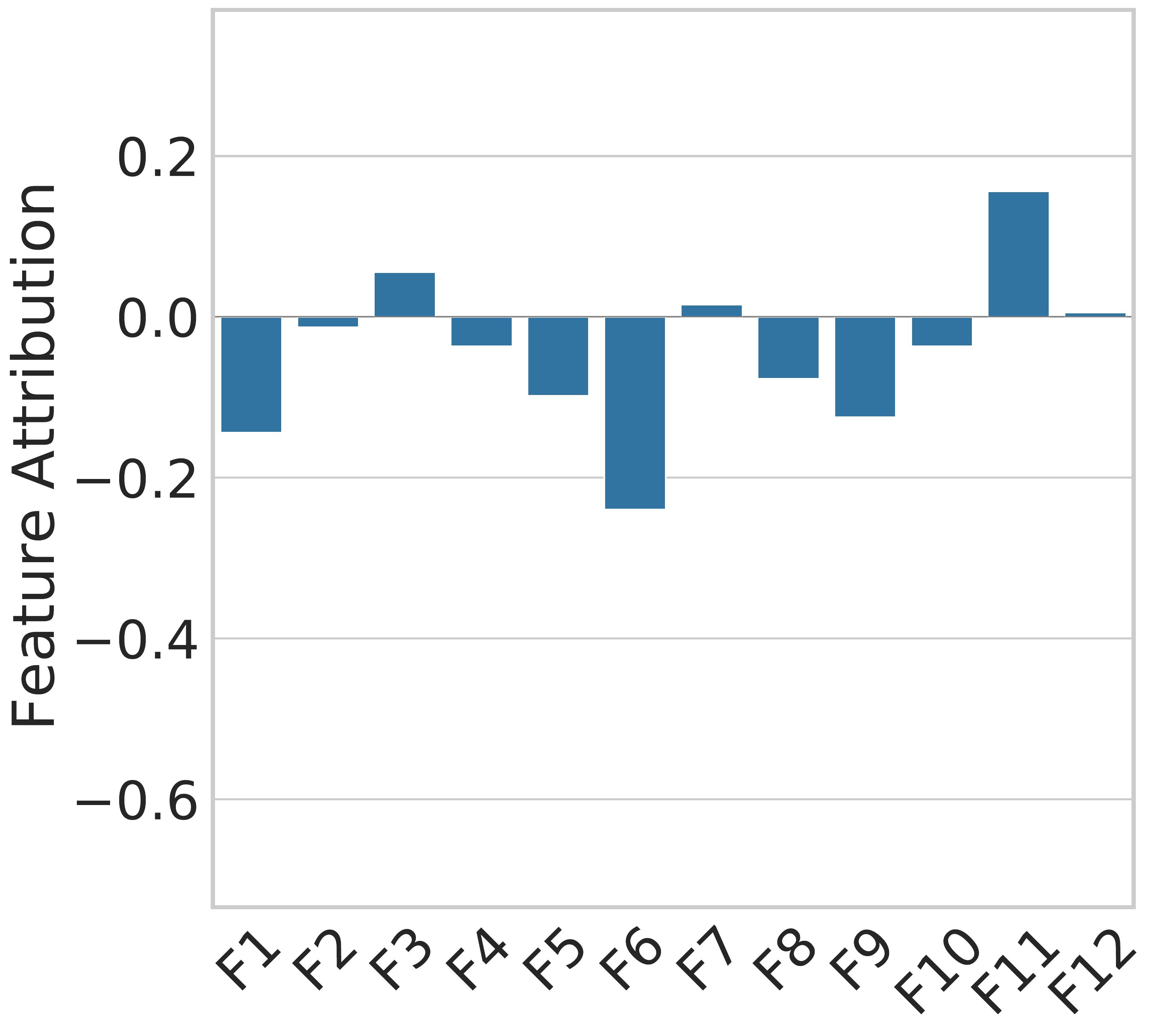} \\
\includegraphics[width=0.235\textwidth]{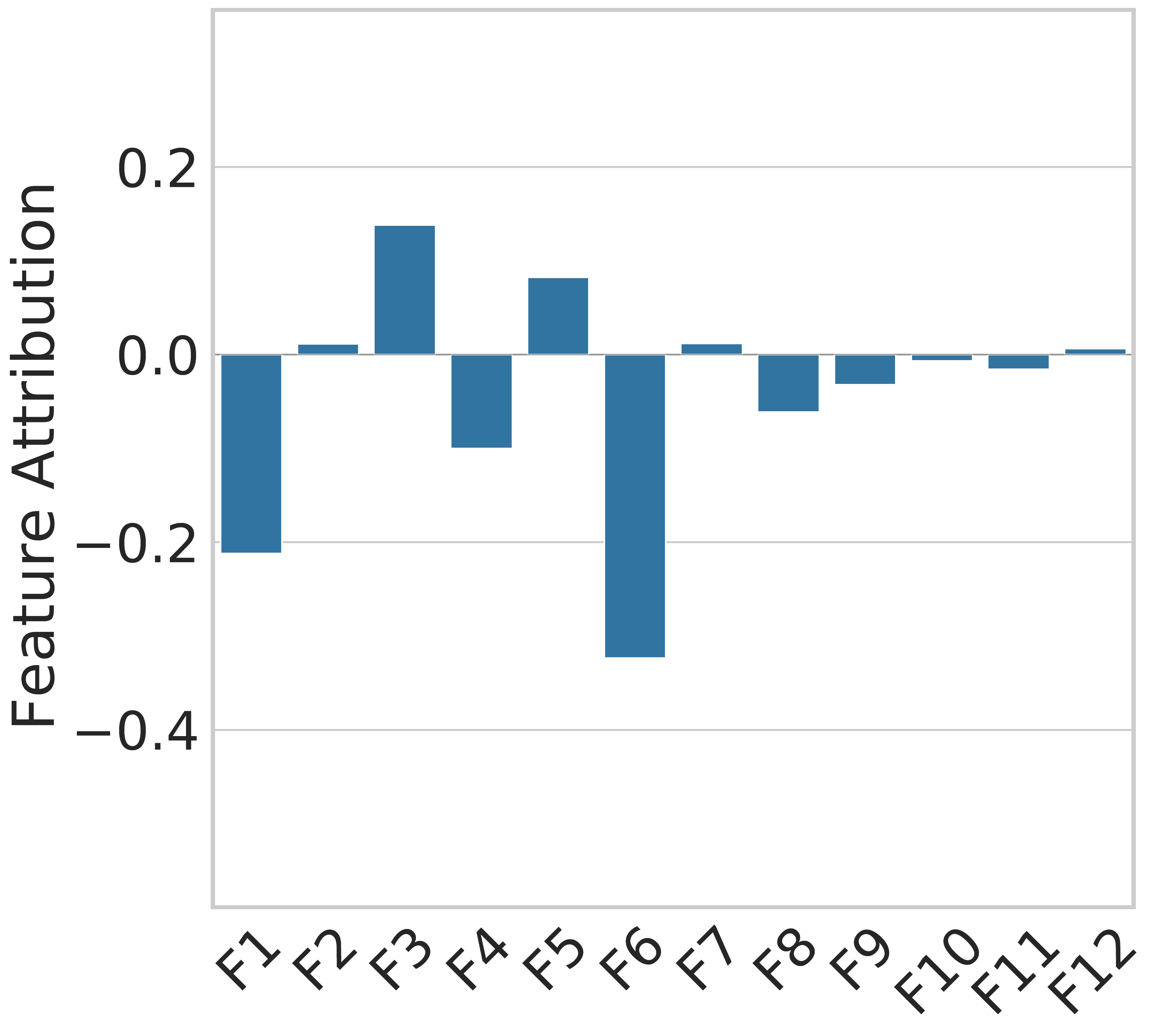} &
\includegraphics[width=0.235\textwidth]{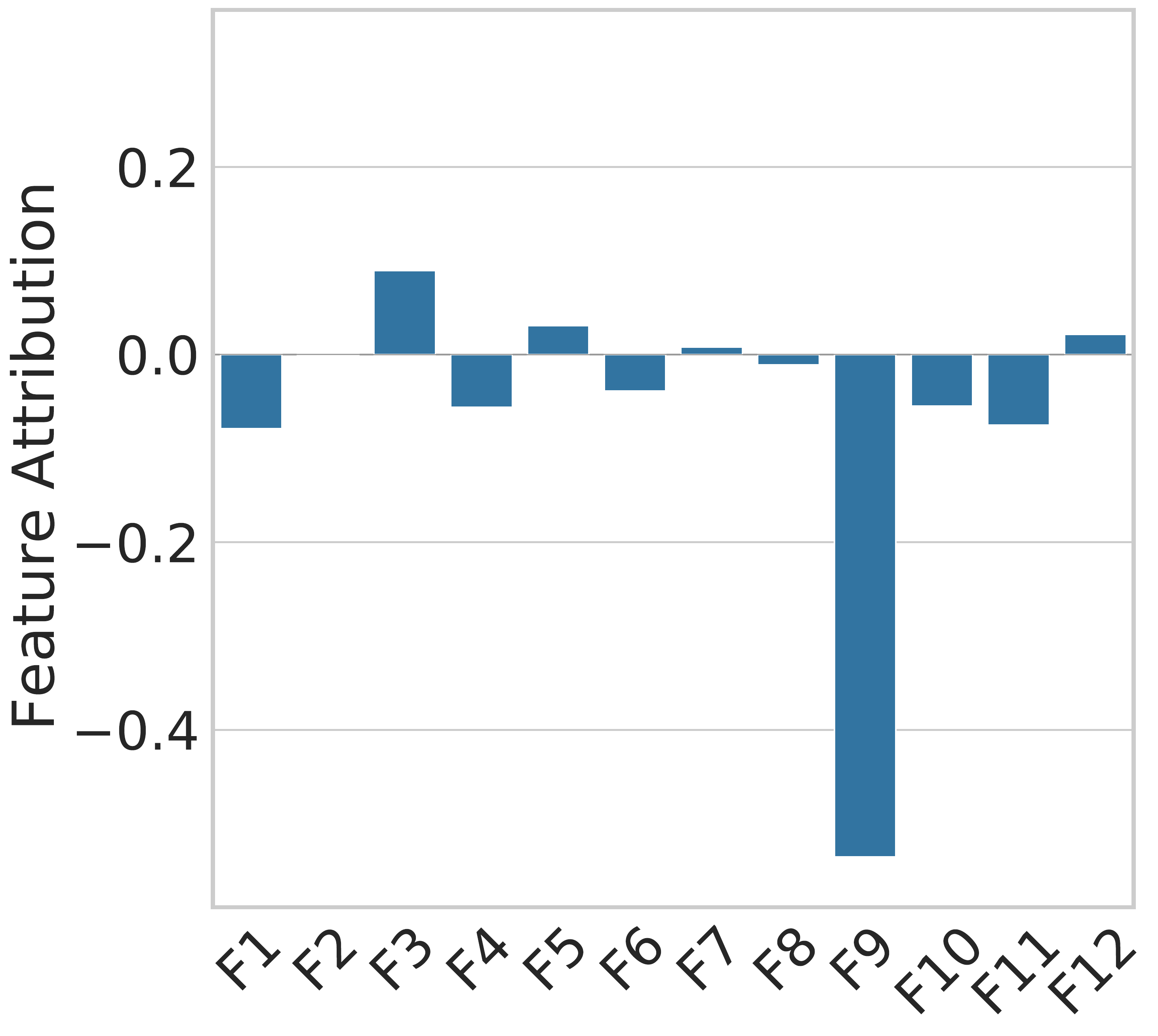} &
\includegraphics[width=0.235\textwidth]{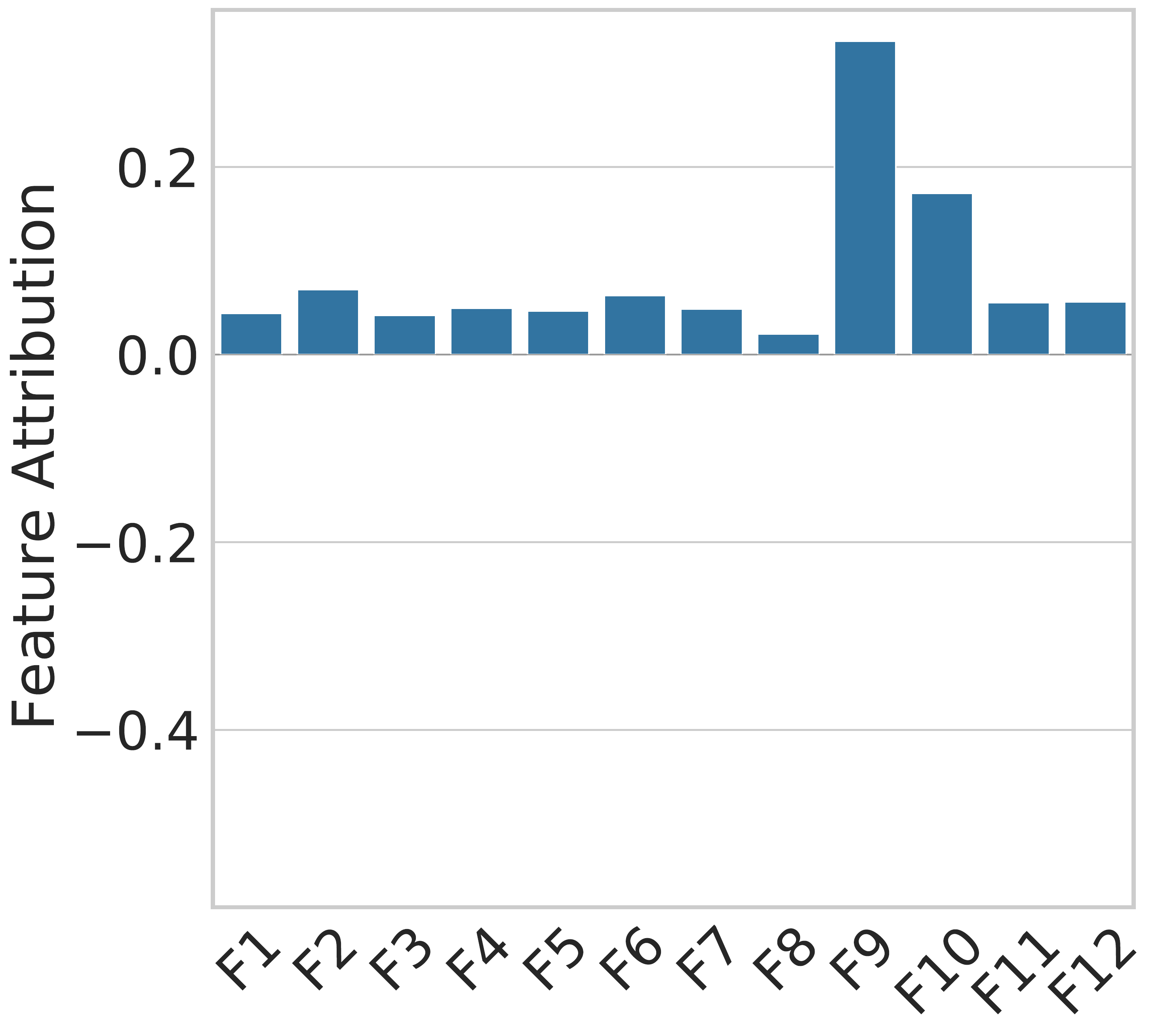} &
\includegraphics[width=0.235\textwidth]{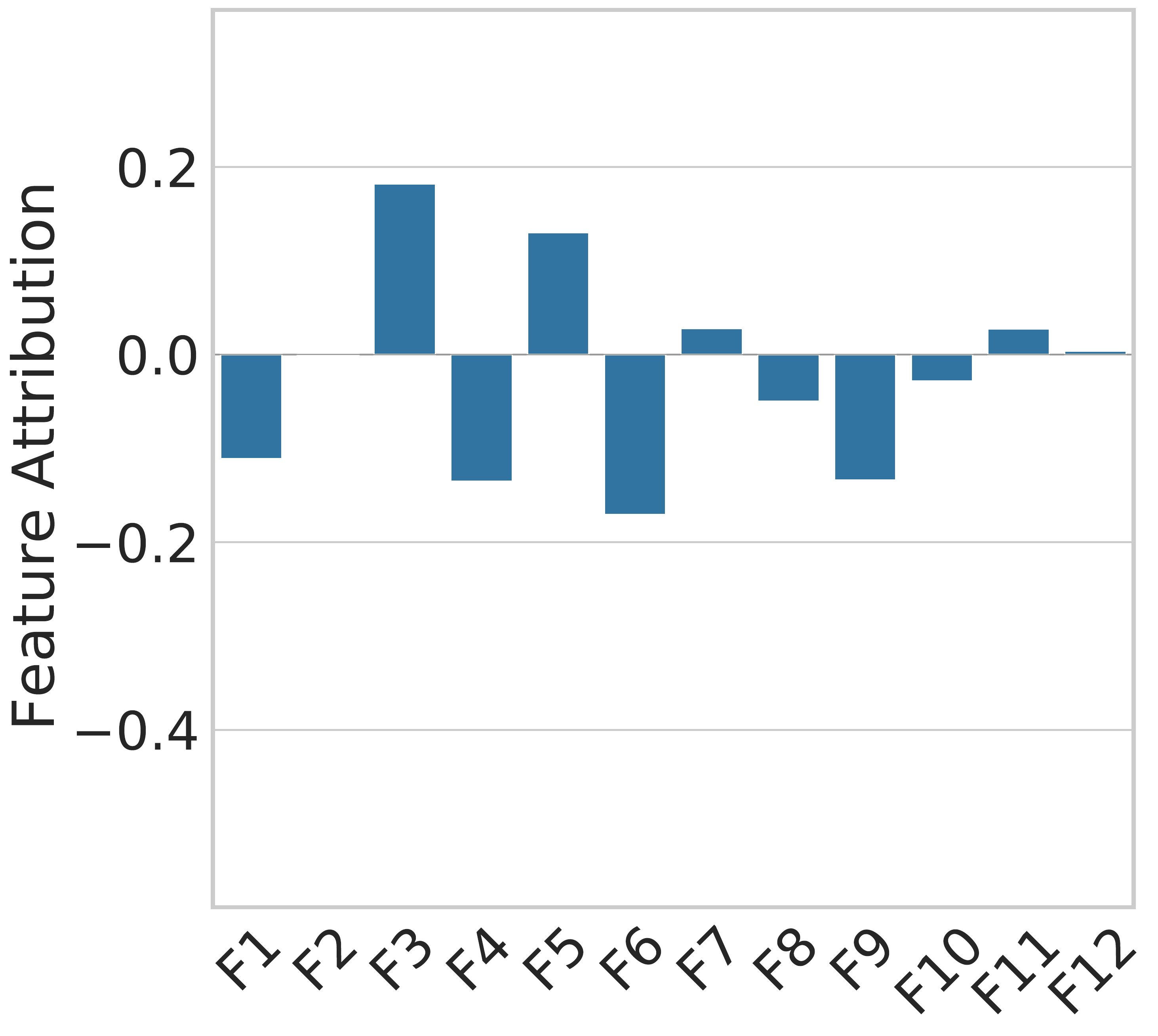} \\
\includegraphics[width=0.235\textwidth]{supplement_figures/adult/2_shap.pdf} &
\includegraphics[width=0.235\textwidth]{supplement_figures/adult/2_lime.pdf} &
\includegraphics[width=0.235\textwidth]{supplement_figures/adult/2_cf.pdf} &
\includegraphics[width=0.235\textwidth]{supplement_figures/adult/2_interventional_shap.pdf} \\
\includegraphics[width=0.235\textwidth]{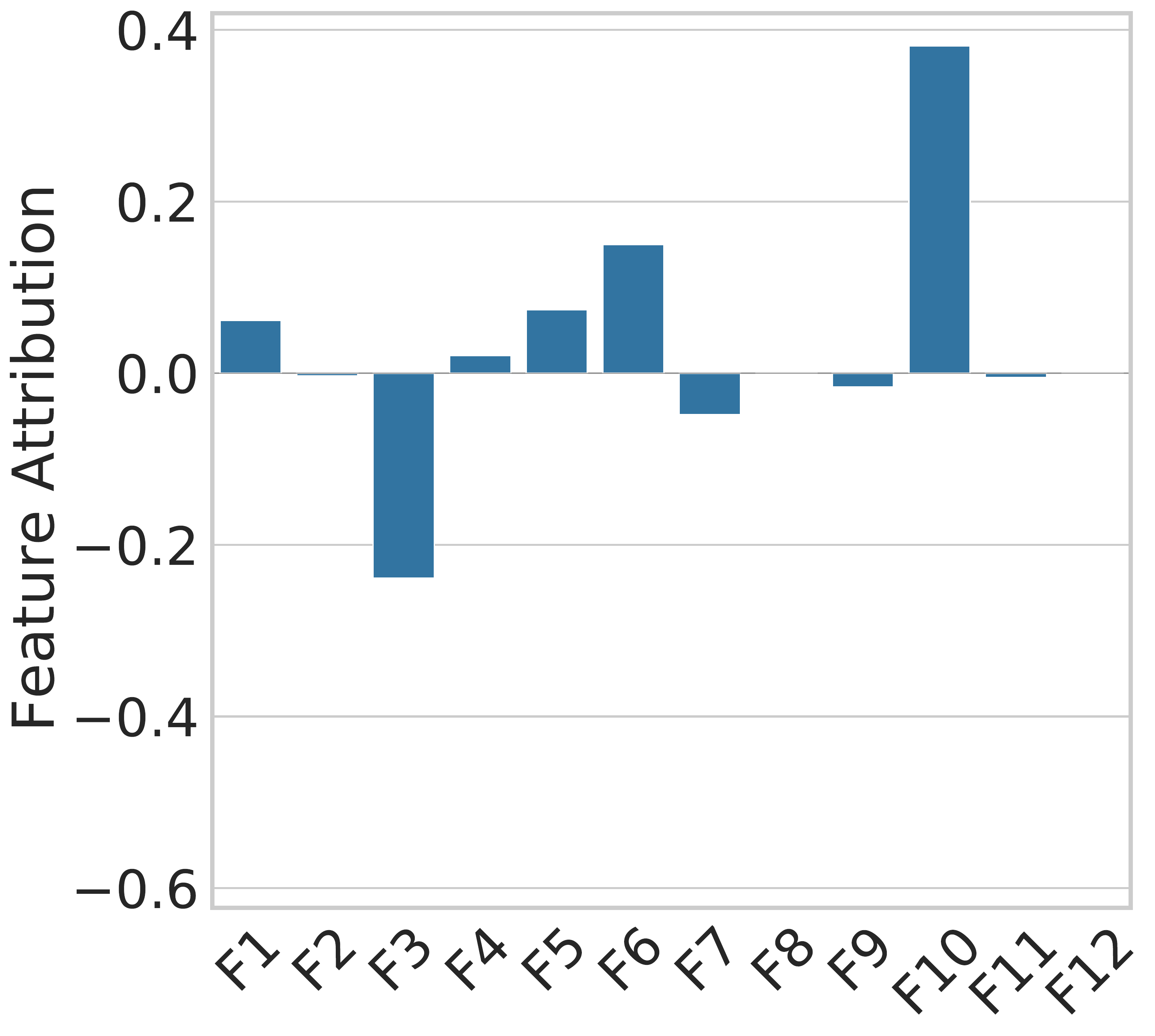} &
\includegraphics[width=0.235\textwidth]{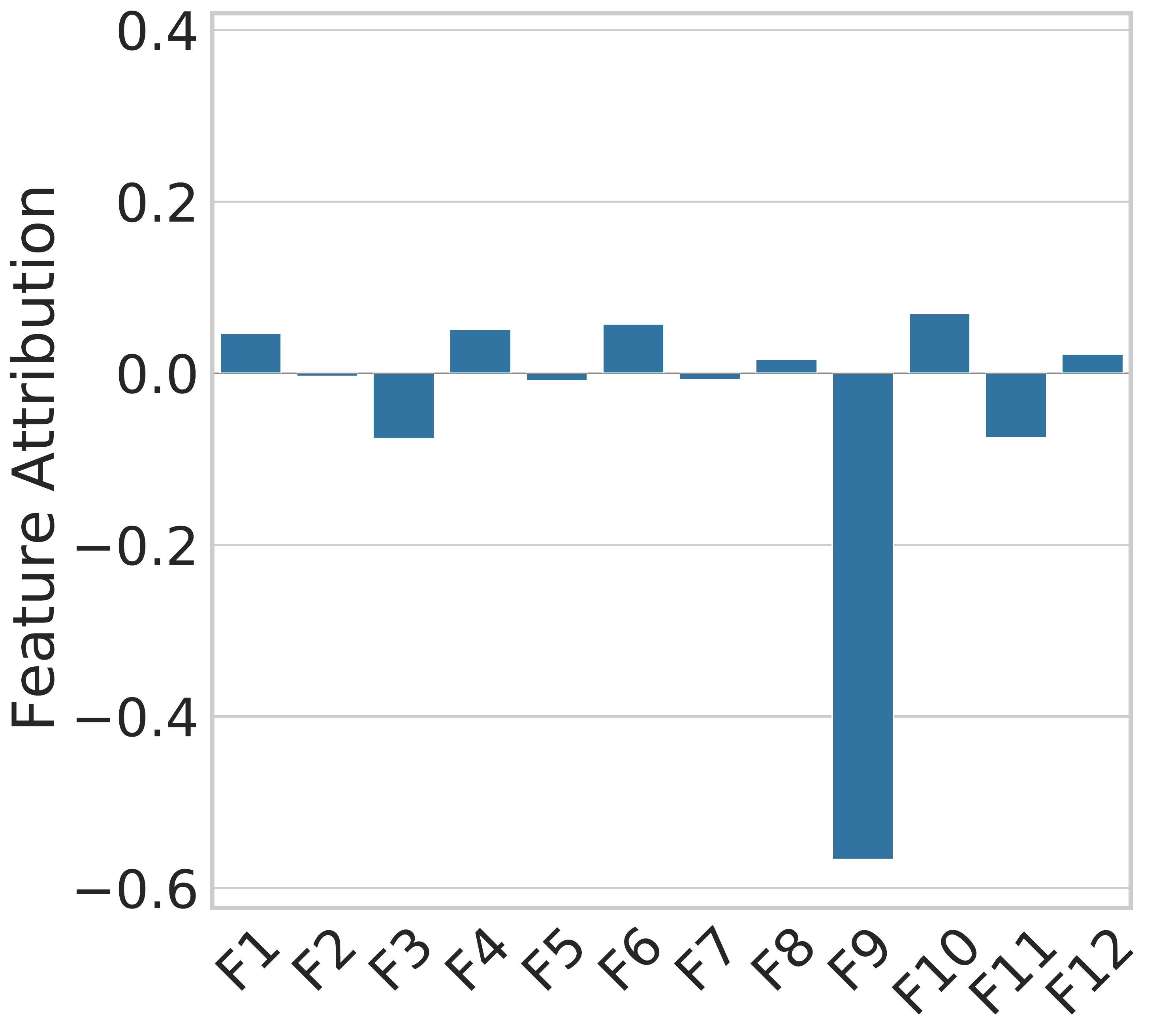} &
\includegraphics[width=0.235\textwidth]{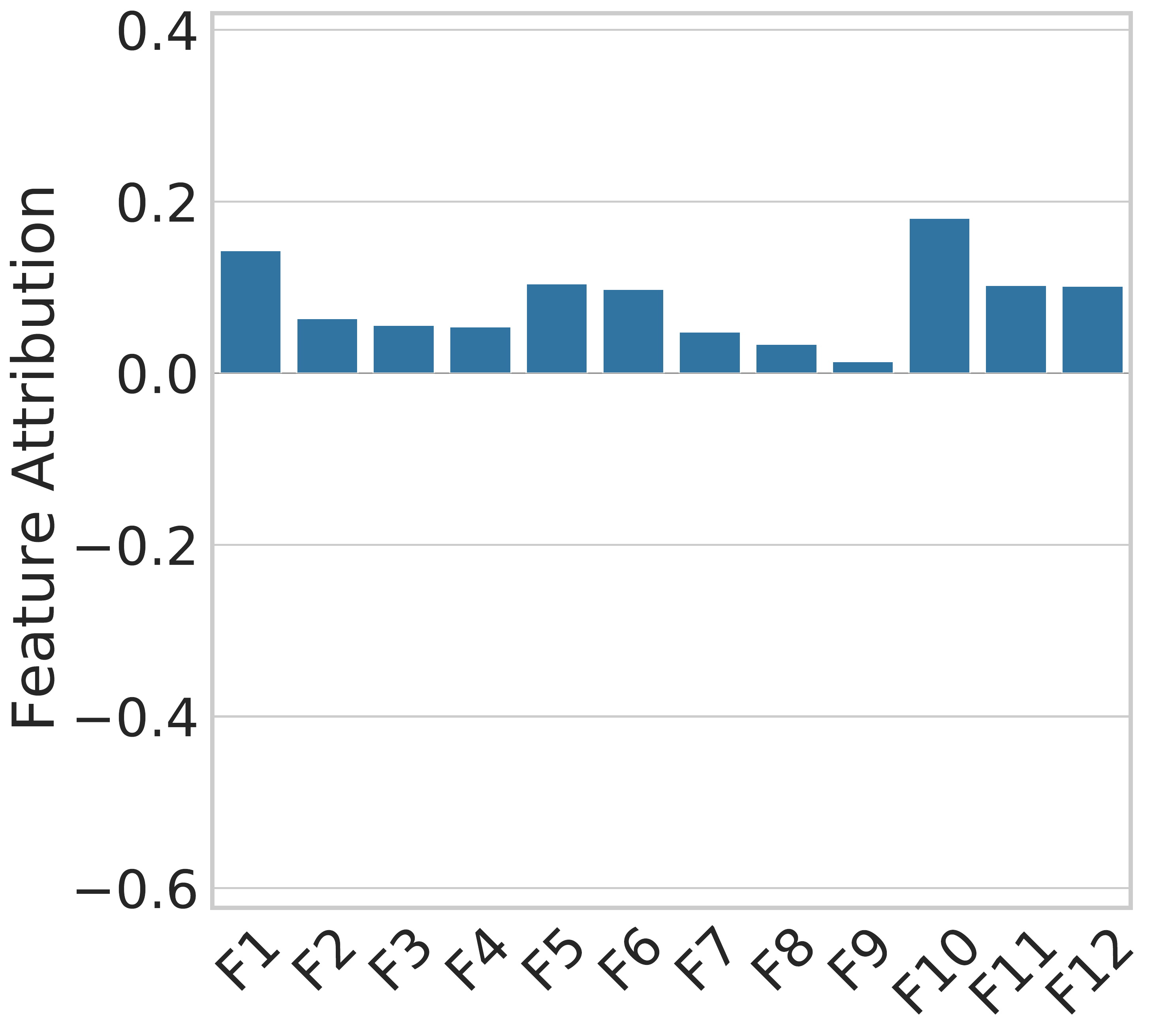} &
\includegraphics[width=0.235\textwidth]{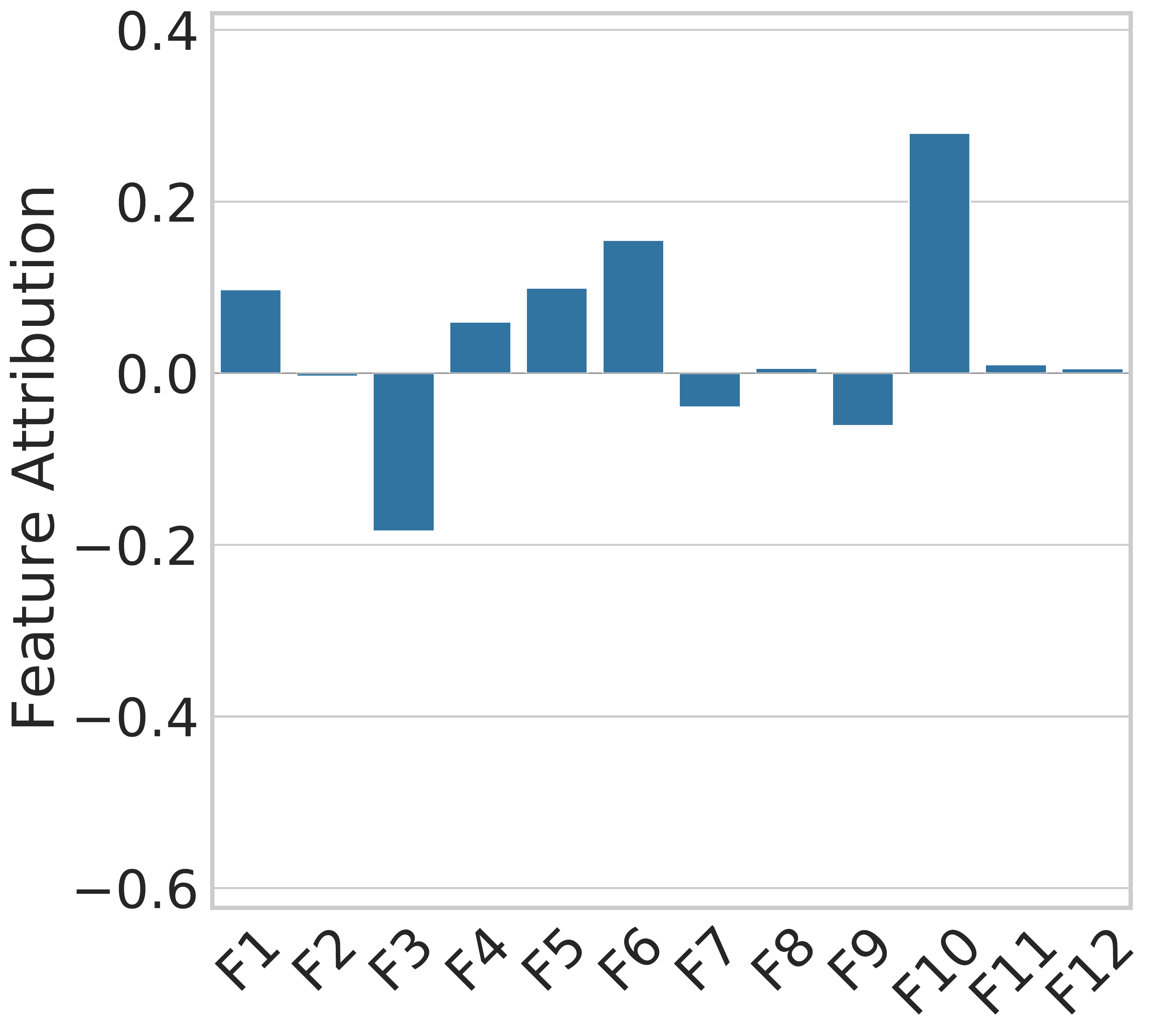} \\
\includegraphics[width=0.235\textwidth]{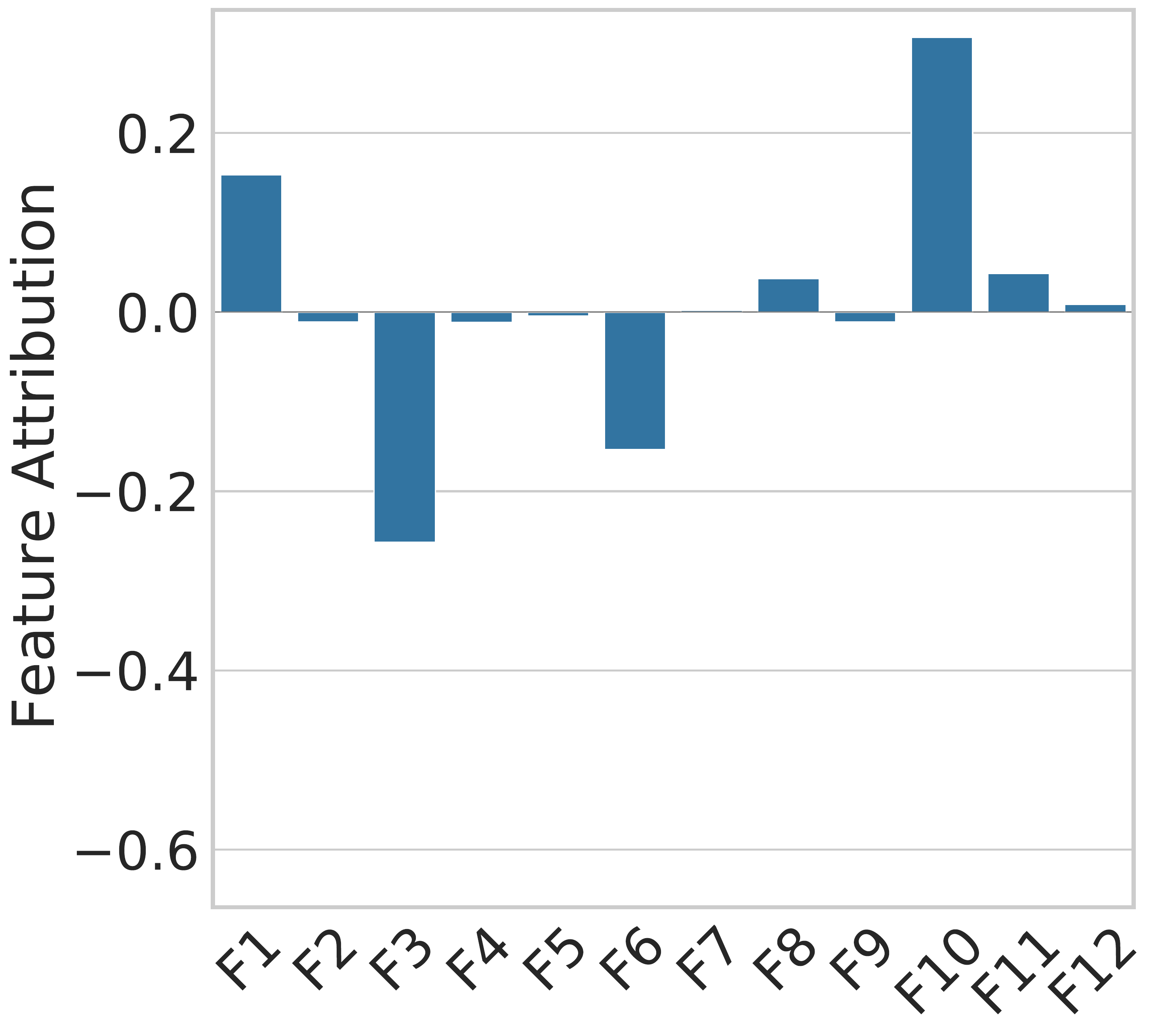} &
\includegraphics[width=0.235\textwidth]{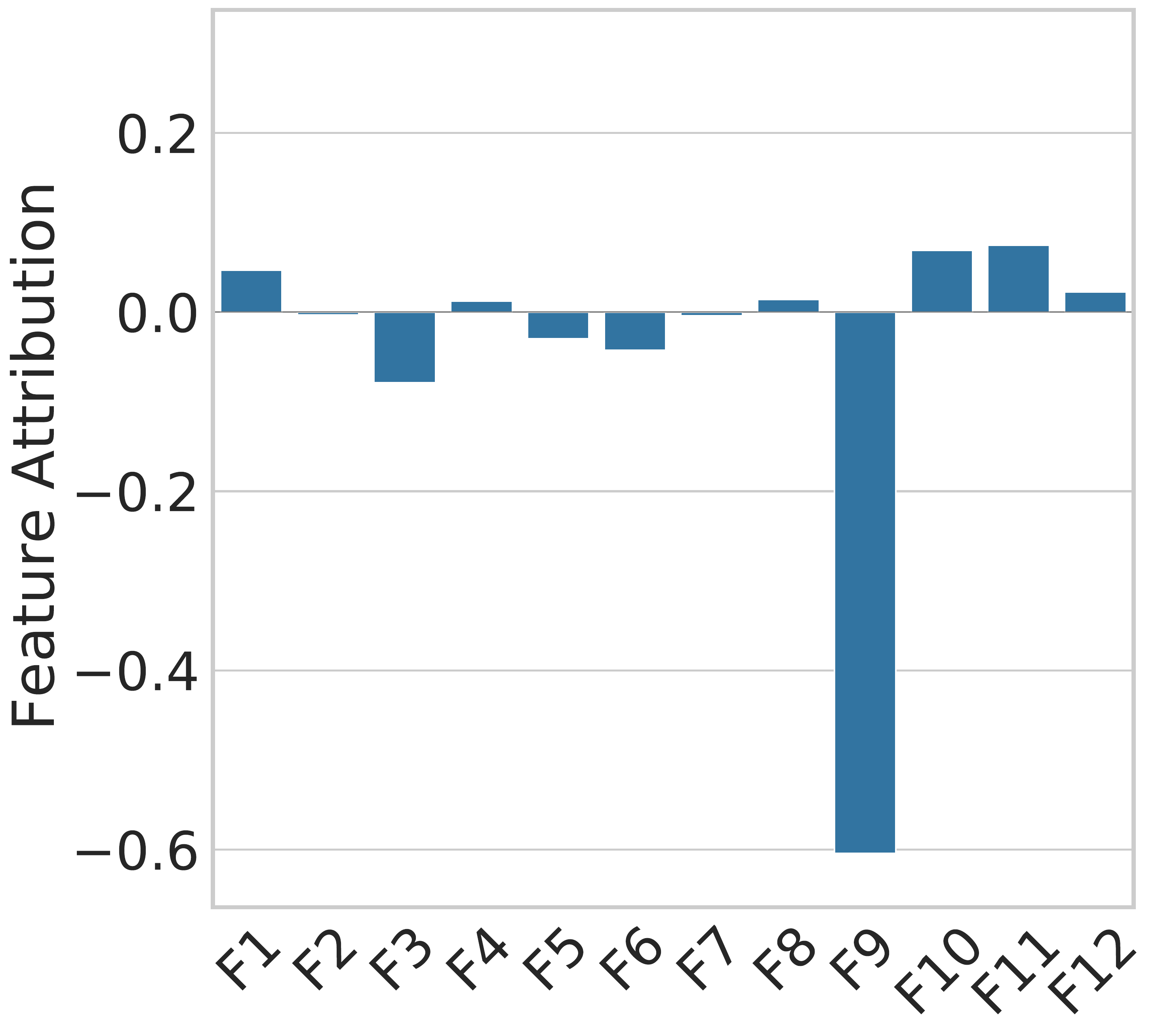} &
\includegraphics[width=0.235\textwidth]{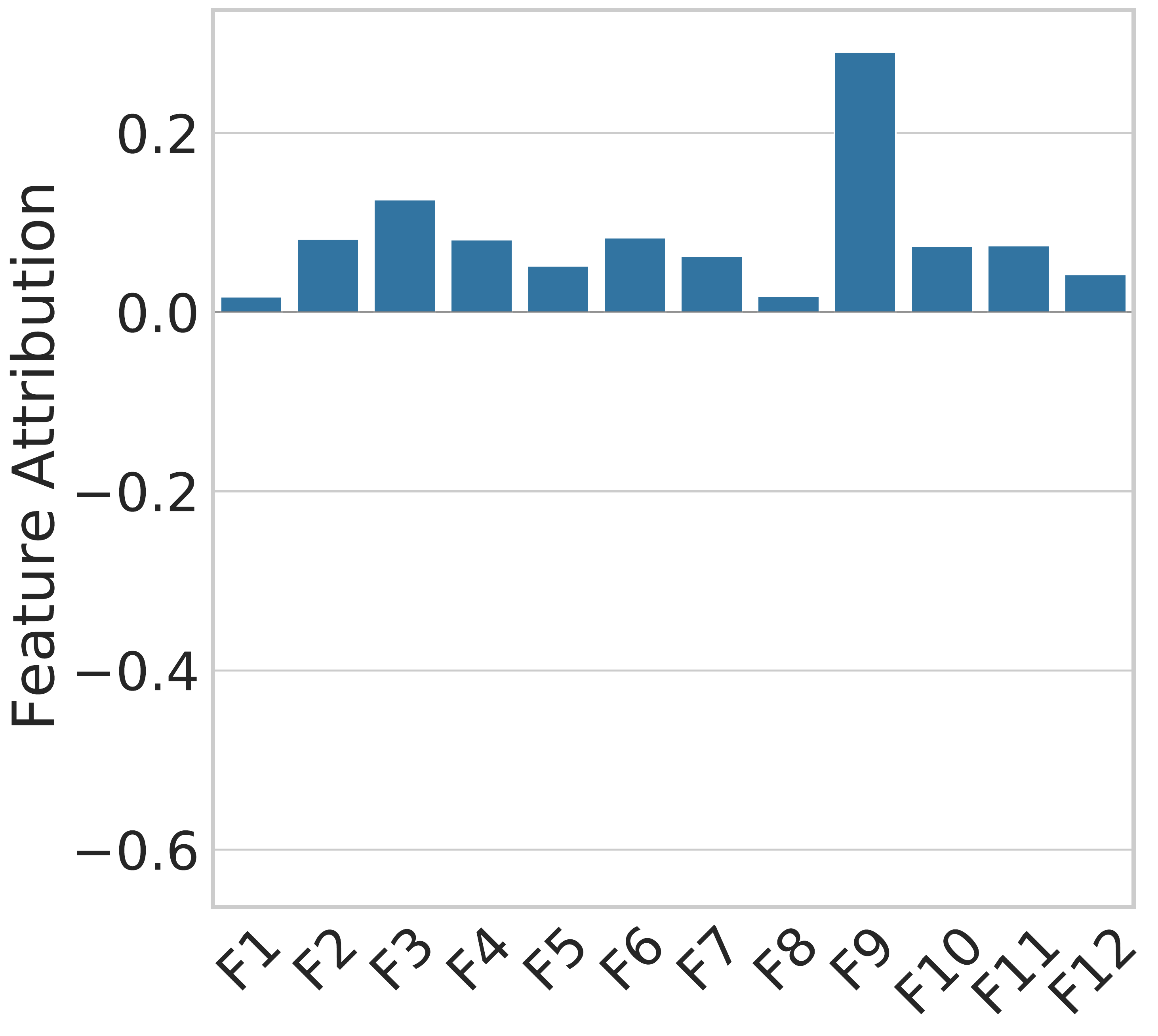} &
\includegraphics[width=0.235\textwidth]{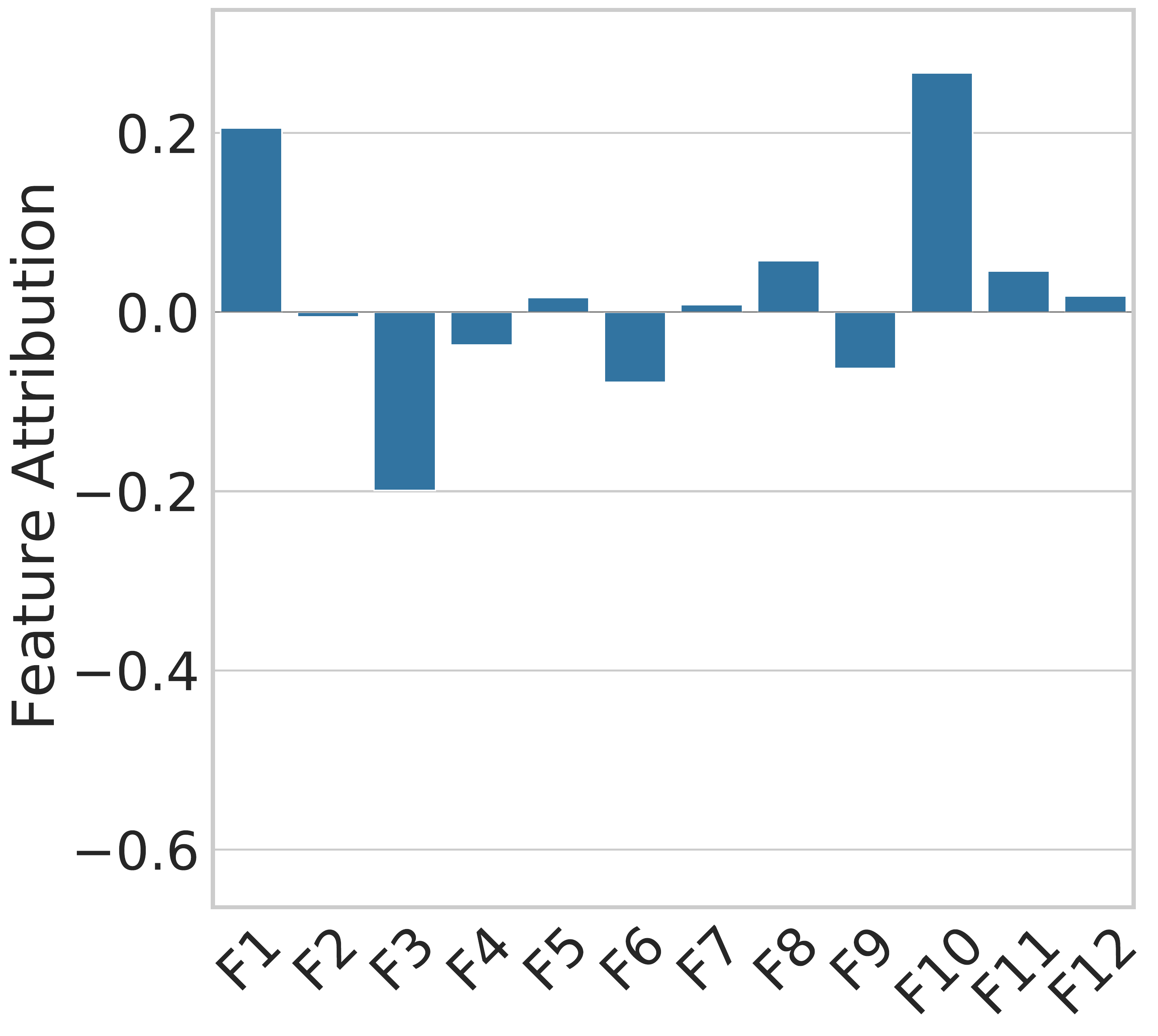} \\
\includegraphics[width=0.235\textwidth]{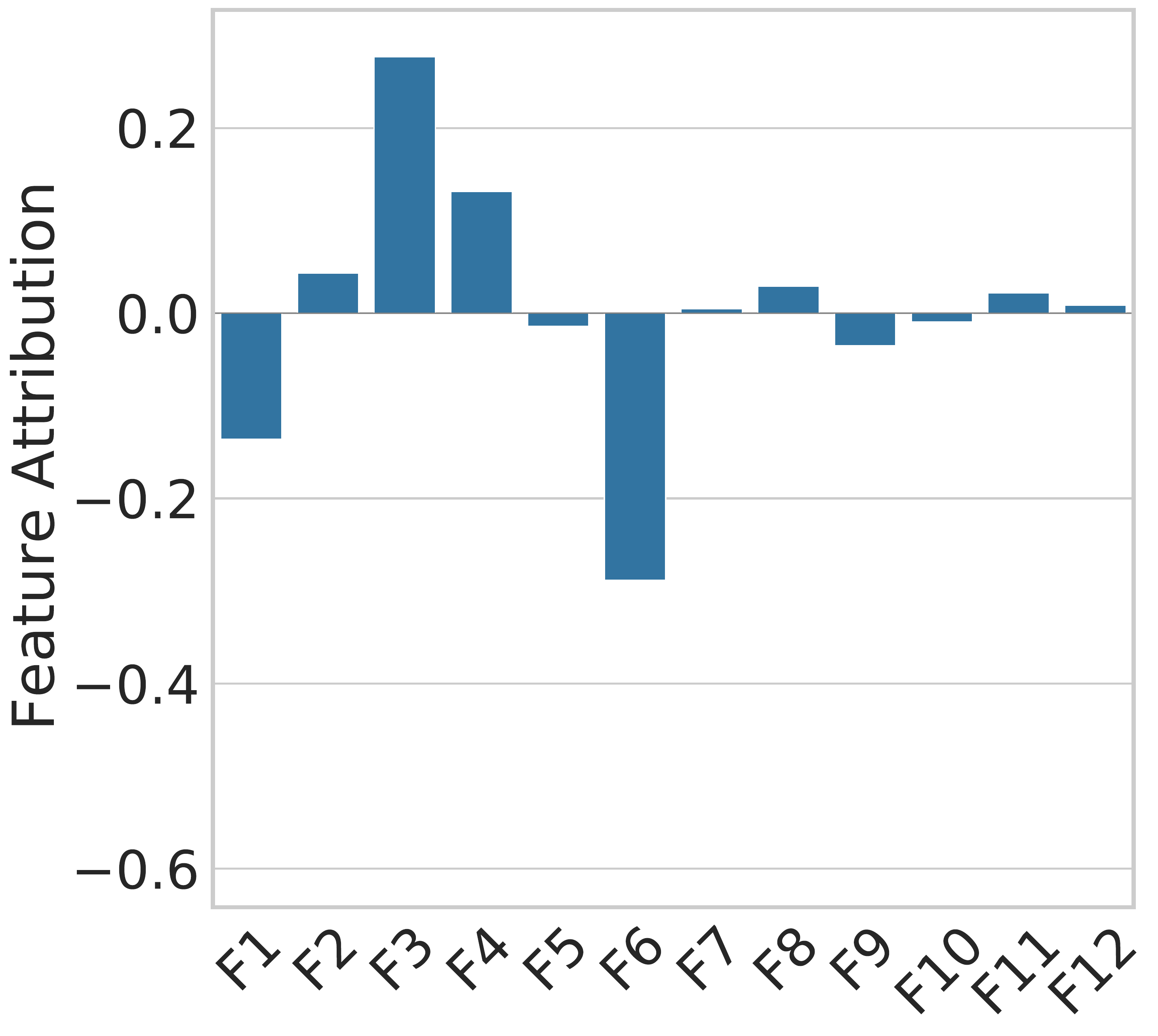} &
\includegraphics[width=0.235\textwidth]{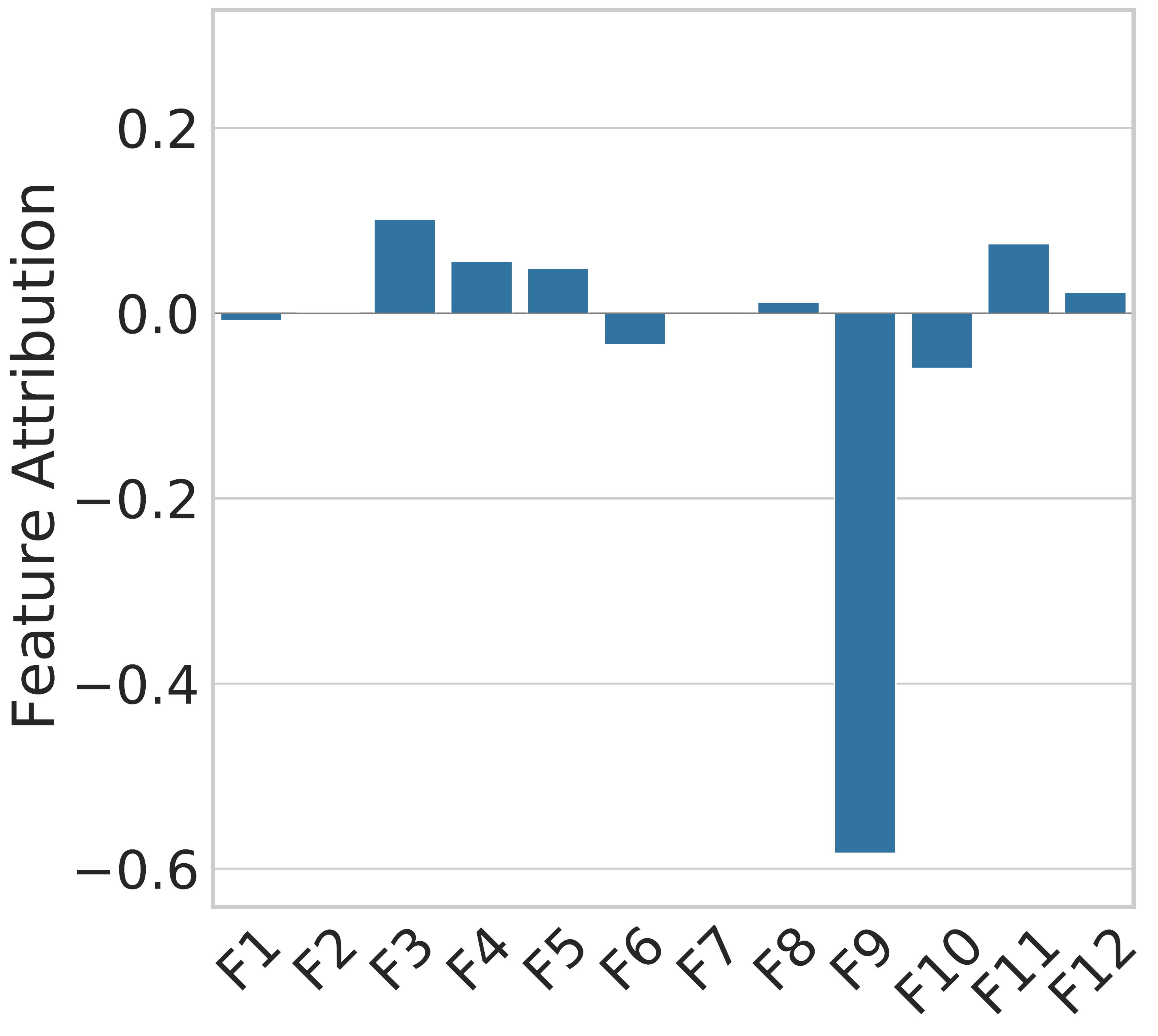} &
\includegraphics[width=0.235\textwidth]{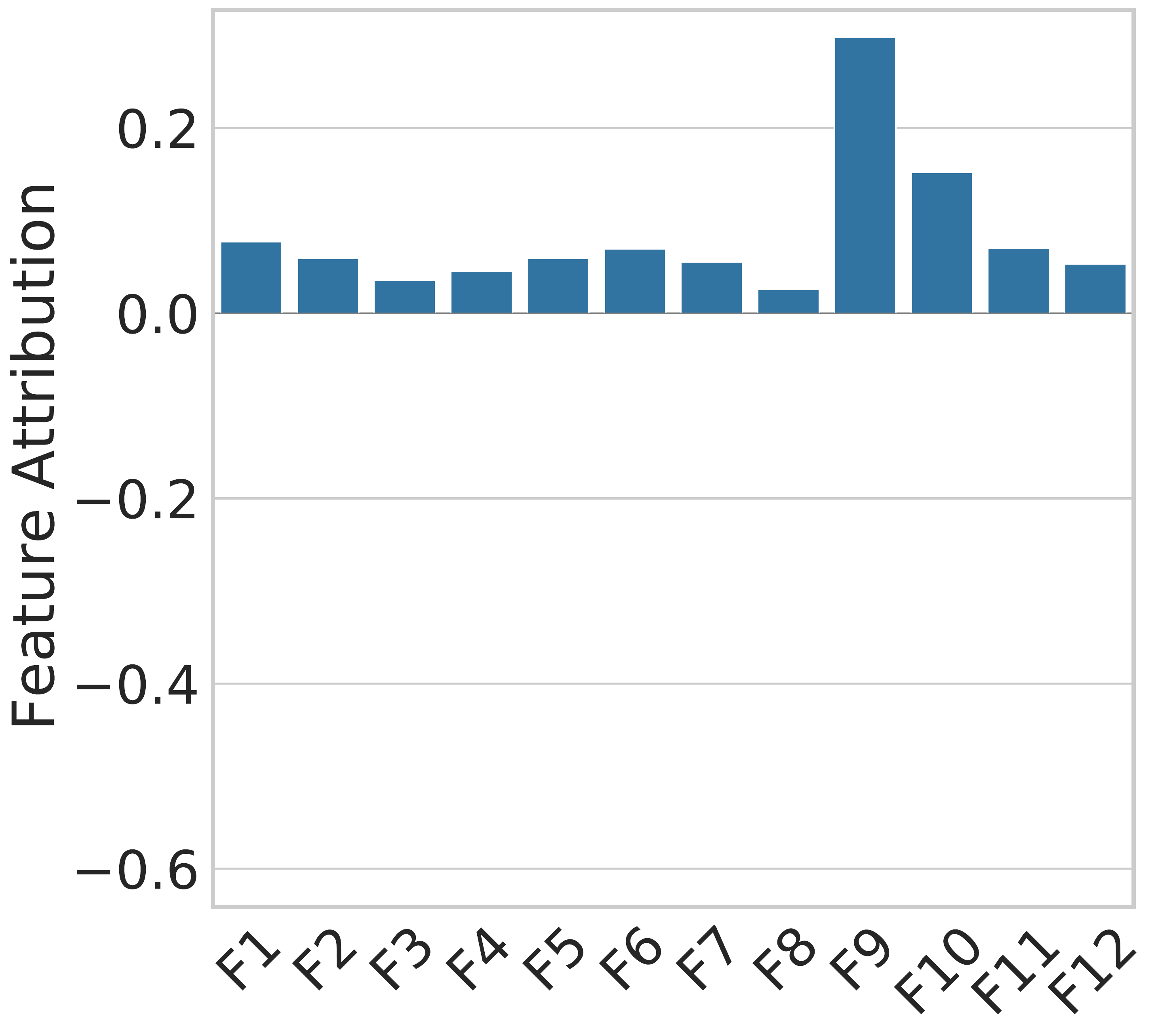} &
\includegraphics[width=0.235\textwidth]{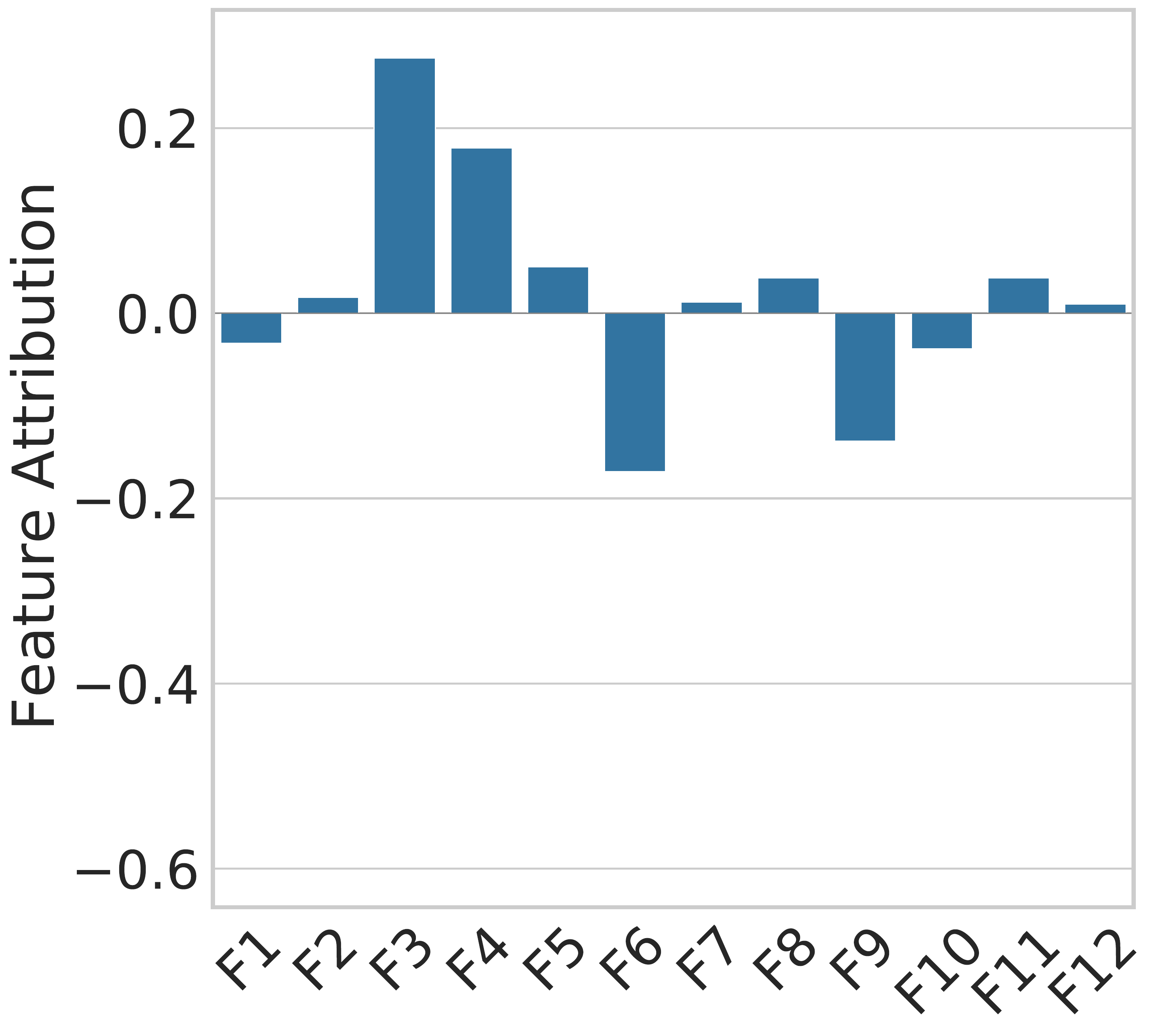} \\

(a) SHAP & (b) LIME & (c) DiCE & (d) Interventional SHAP\\
\end{tabular}

\caption{Different explanation algorithms lead to different explanations (compare Figure \ref{fig:different_explanations} in the main paper). Every row depicts the explanations of the four different explanation algorithms for another individual. The Figure depicts the first 6 observations from the test set.}
\end{figure*}

\newpage

\begin{center} 
\huge \bf Additional Figures Related to Figure \ref{fig:similarities-and-dissimilarities} in the Main Paper
\end{center} 
\vspace{0.4cm}

\begin{figure*}[h]

\begin{tabular}{cc}

\includegraphics[width=0.225\textwidth]{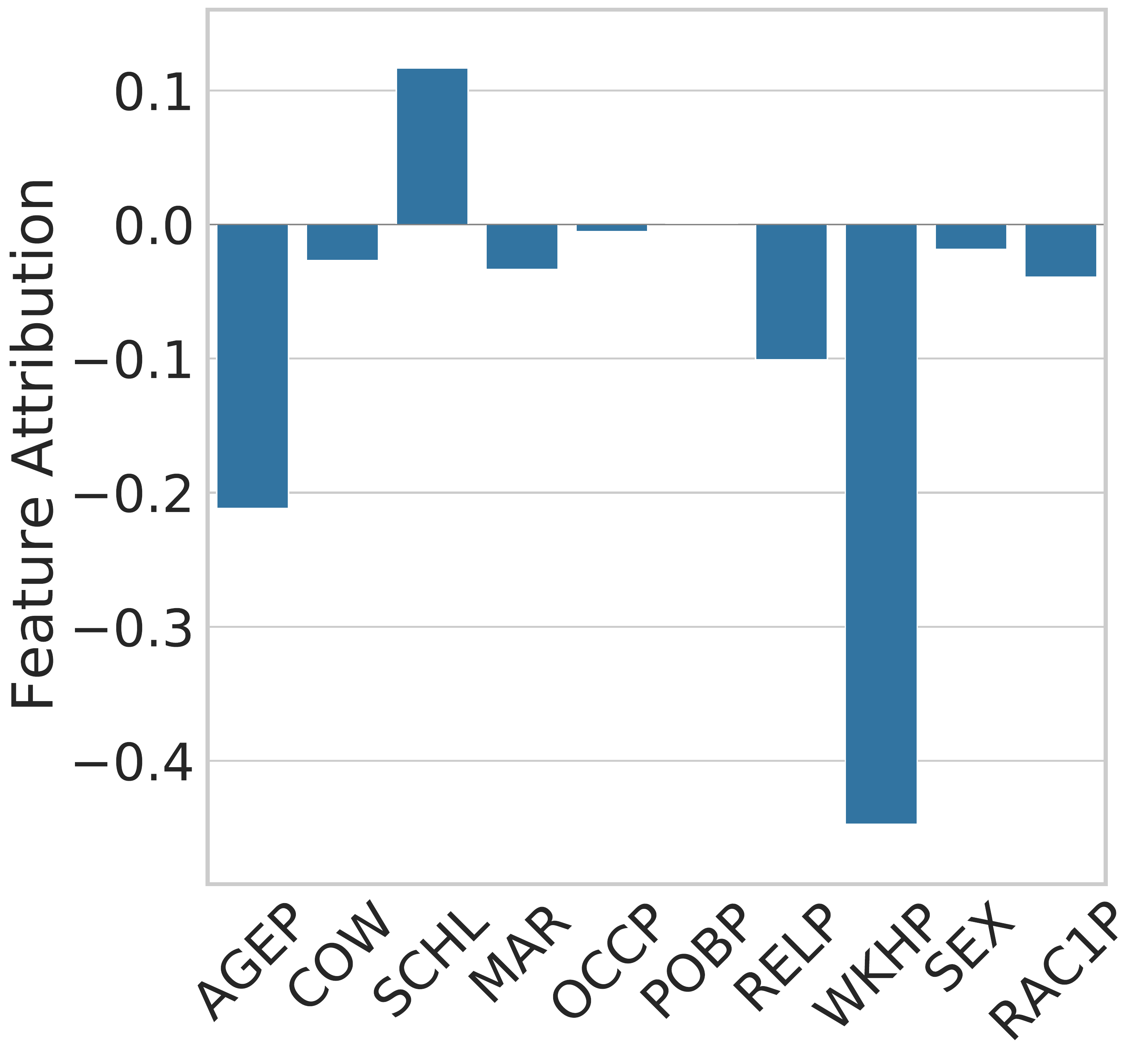} &
\includegraphics[width=0.225\textwidth]{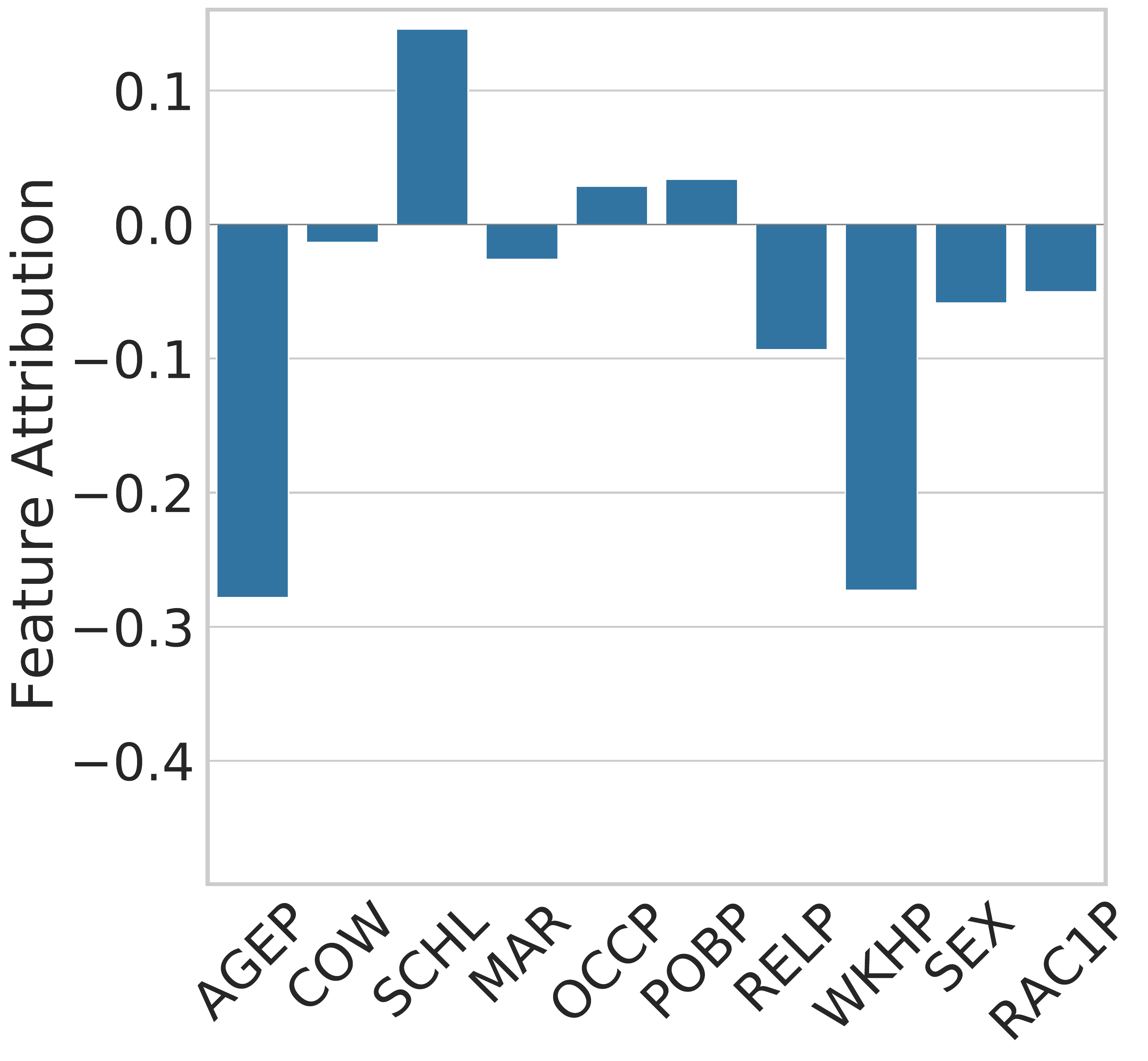}  \\
\includegraphics[width=0.225\textwidth]{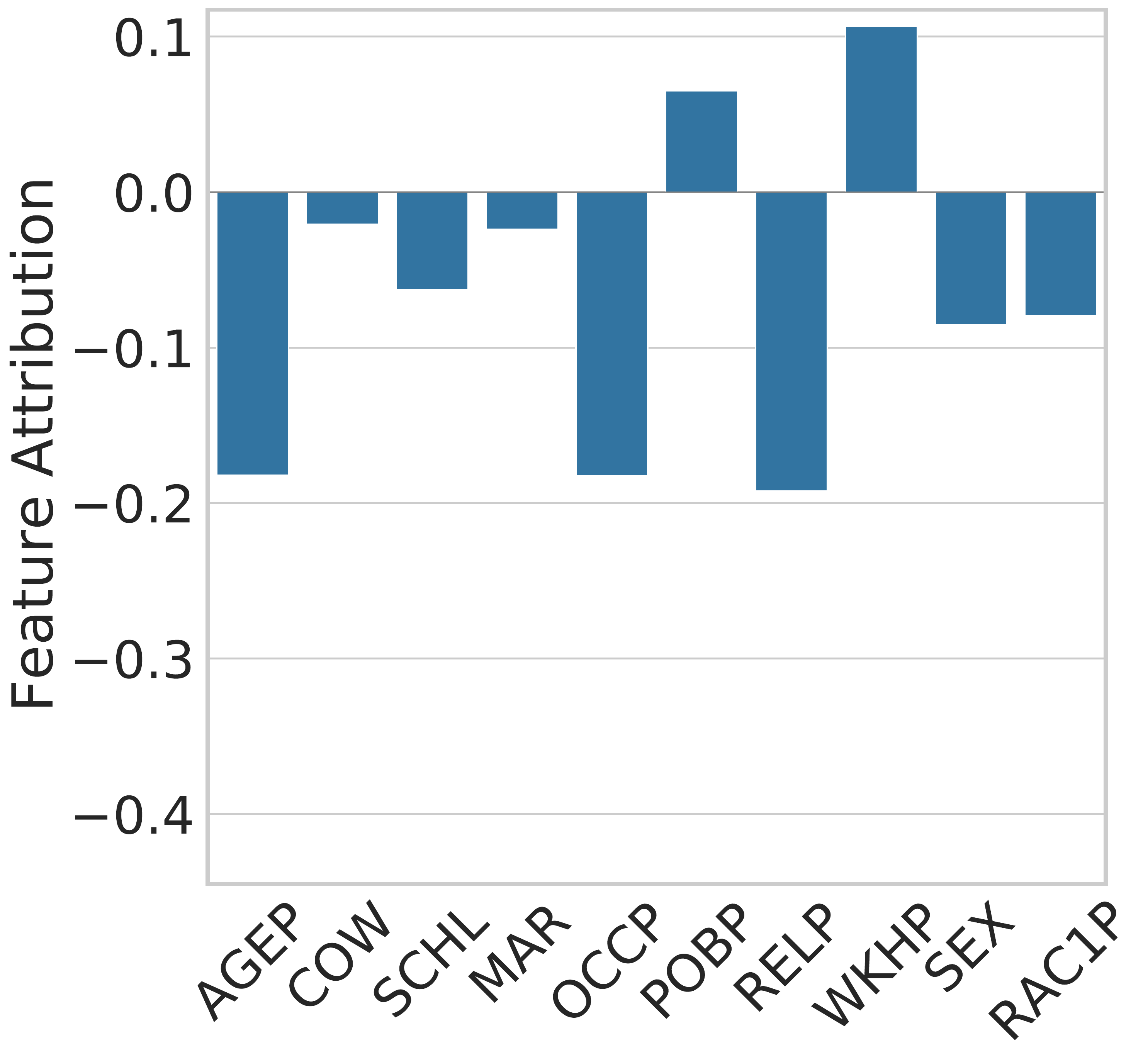} &
\includegraphics[width=0.225\textwidth]{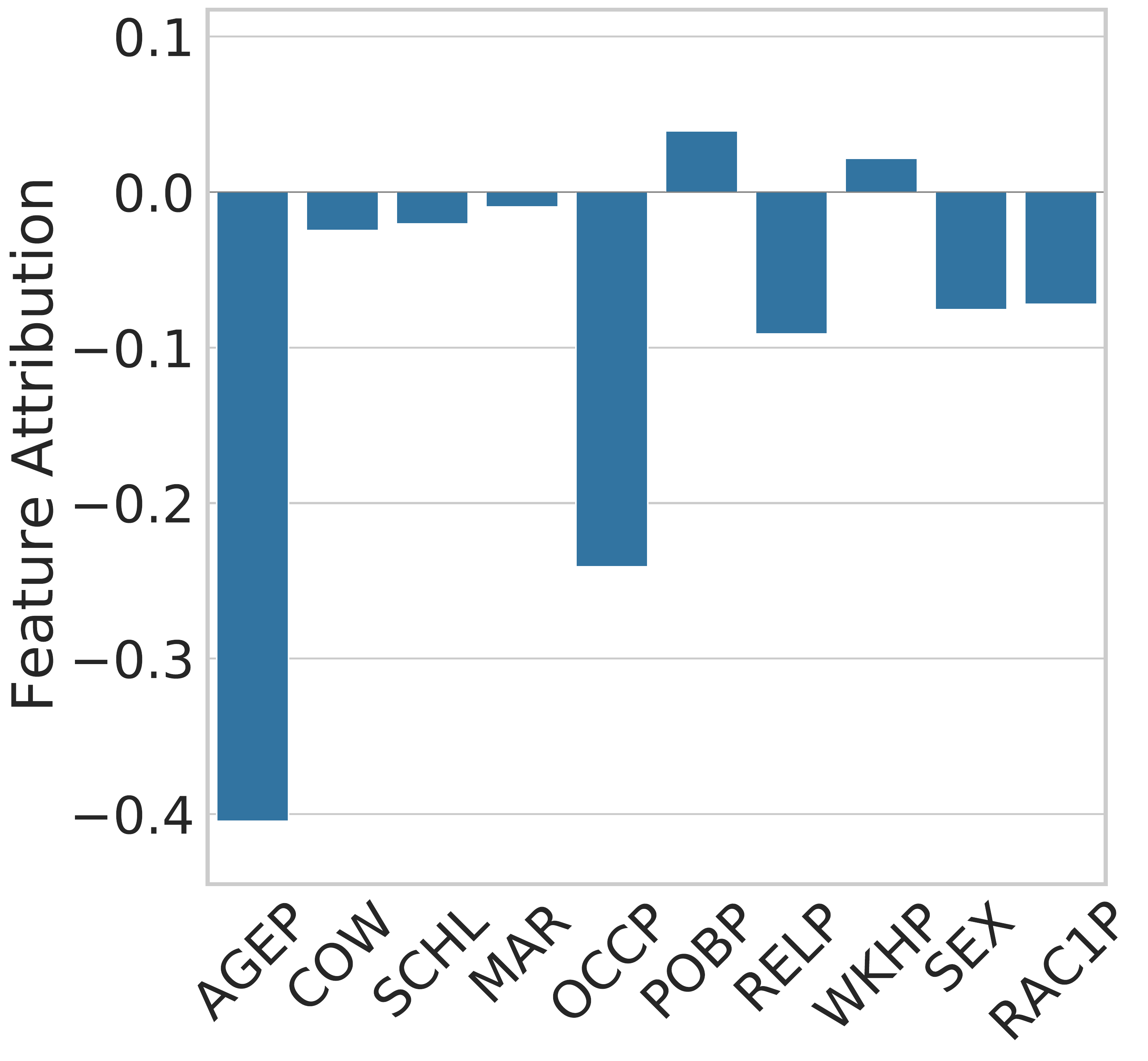}  \\
\includegraphics[width=0.225\textwidth]{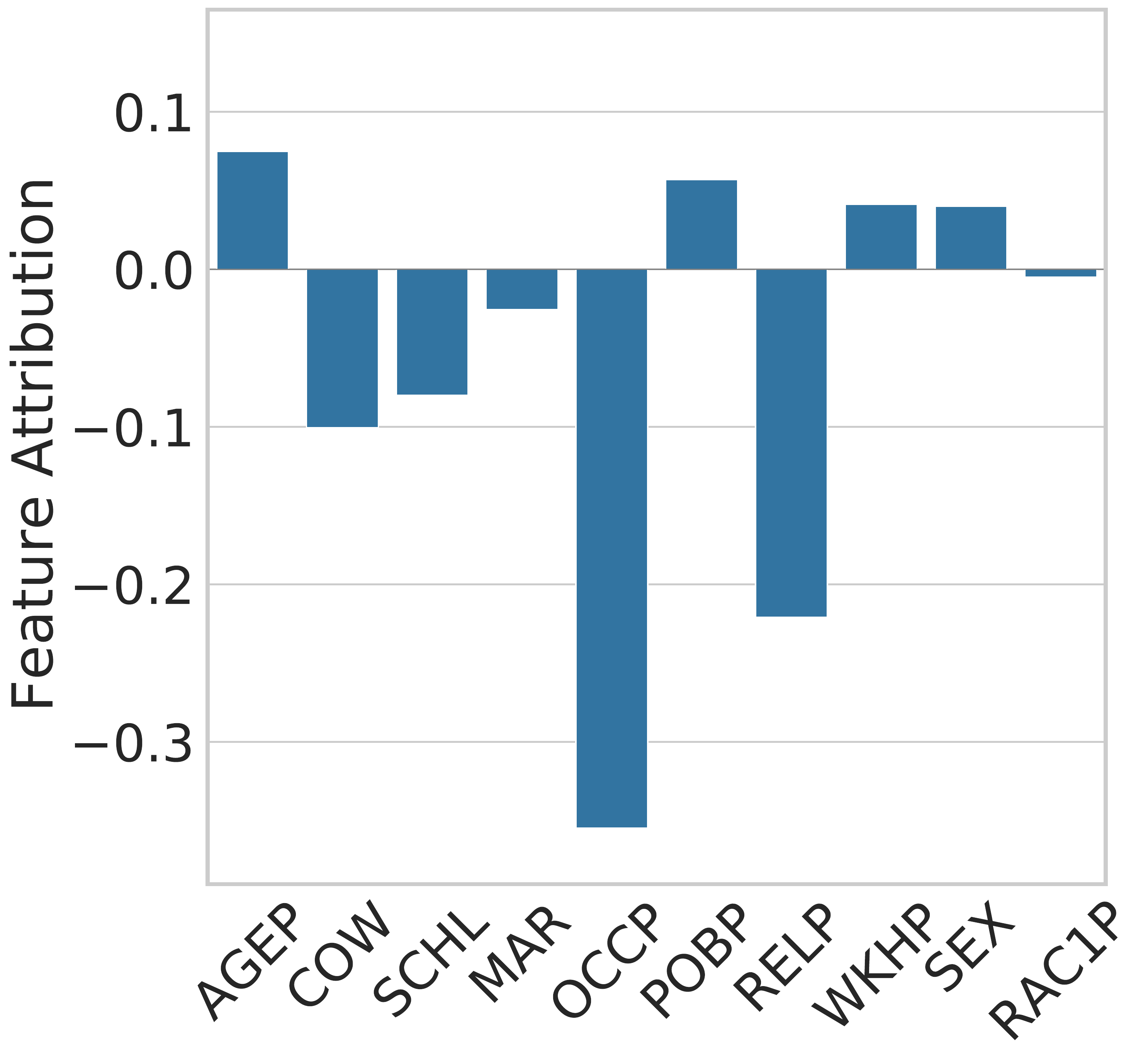} &
\includegraphics[width=0.225\textwidth]{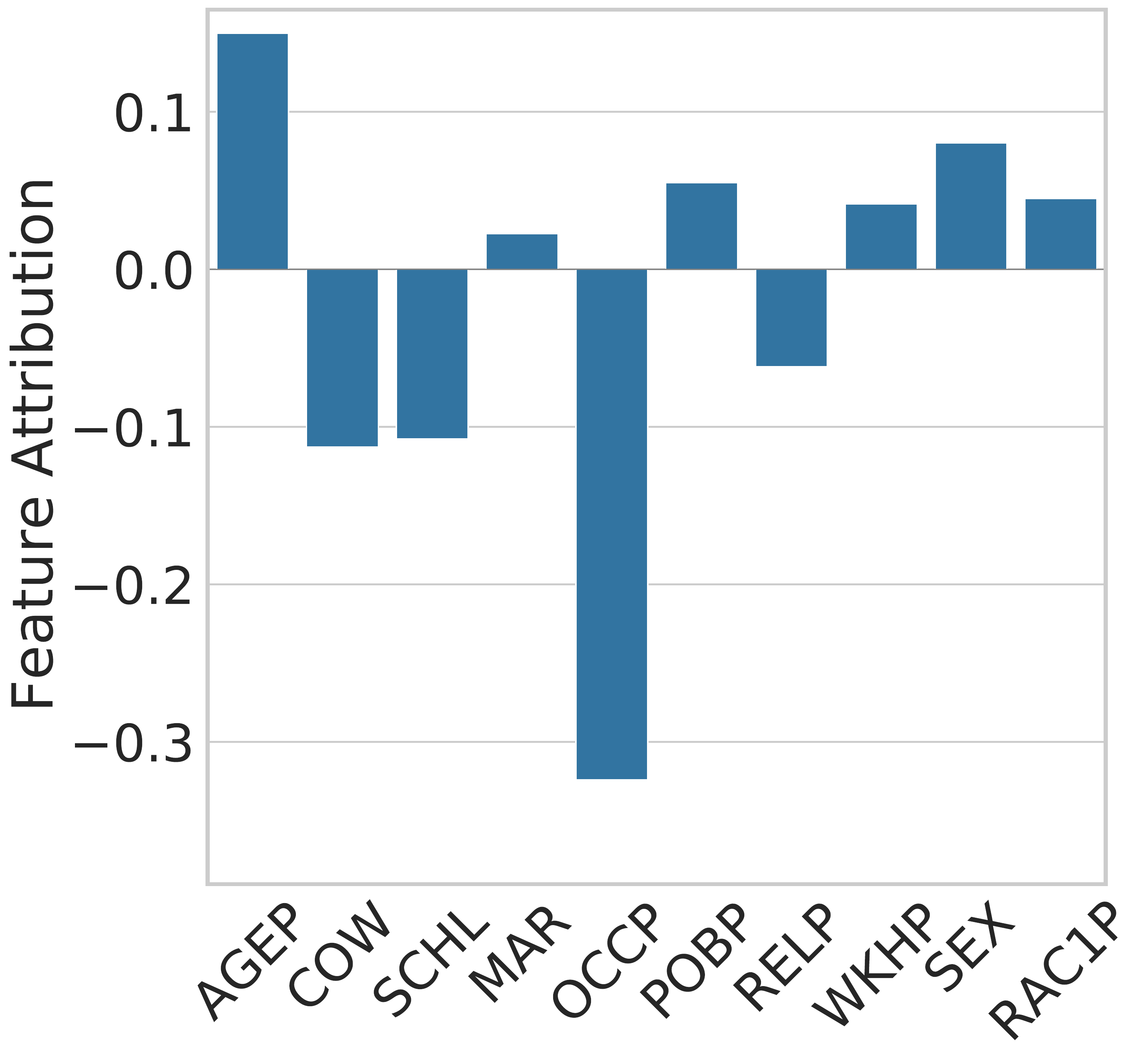}  \\
\includegraphics[width=0.225\textwidth]{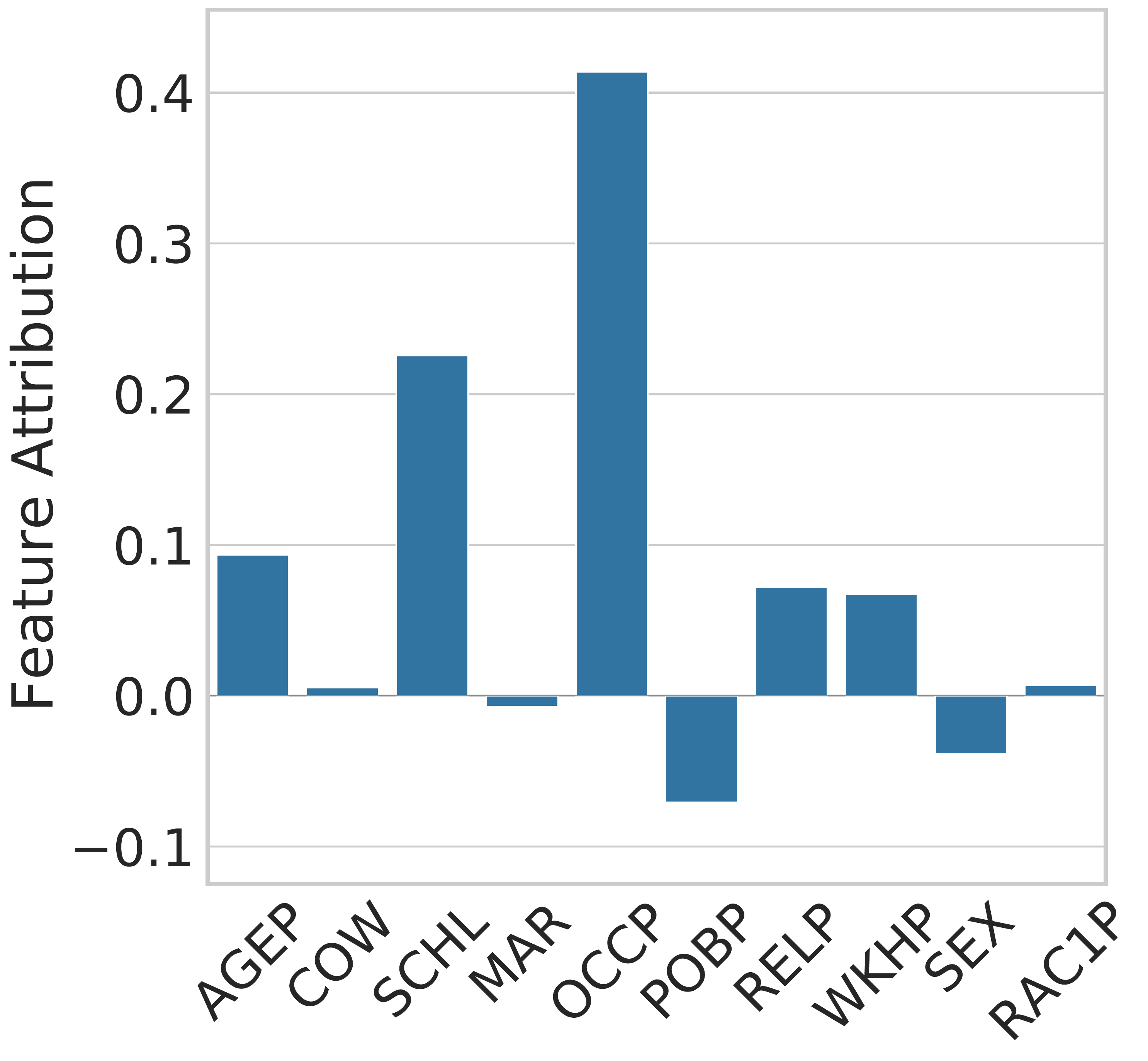} &
\includegraphics[width=0.225\textwidth]{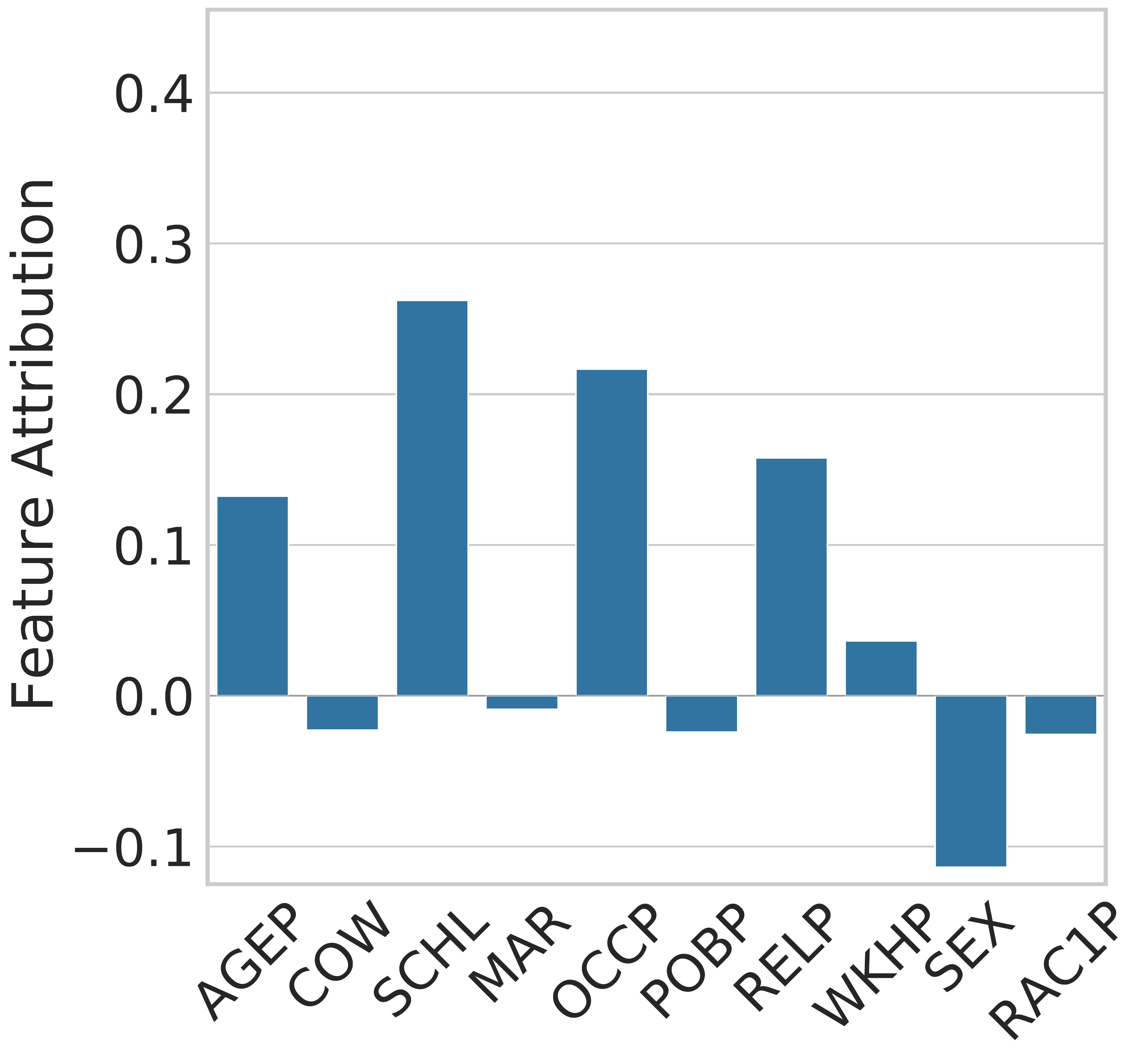}  \\
\includegraphics[width=0.225\textwidth]{supplement_figures/folktables/4_shap.pdf} &
\includegraphics[width=0.225\textwidth]{supplement_figures/folktables/4_lime.pdf}  \\
\includegraphics[width=0.225\textwidth]{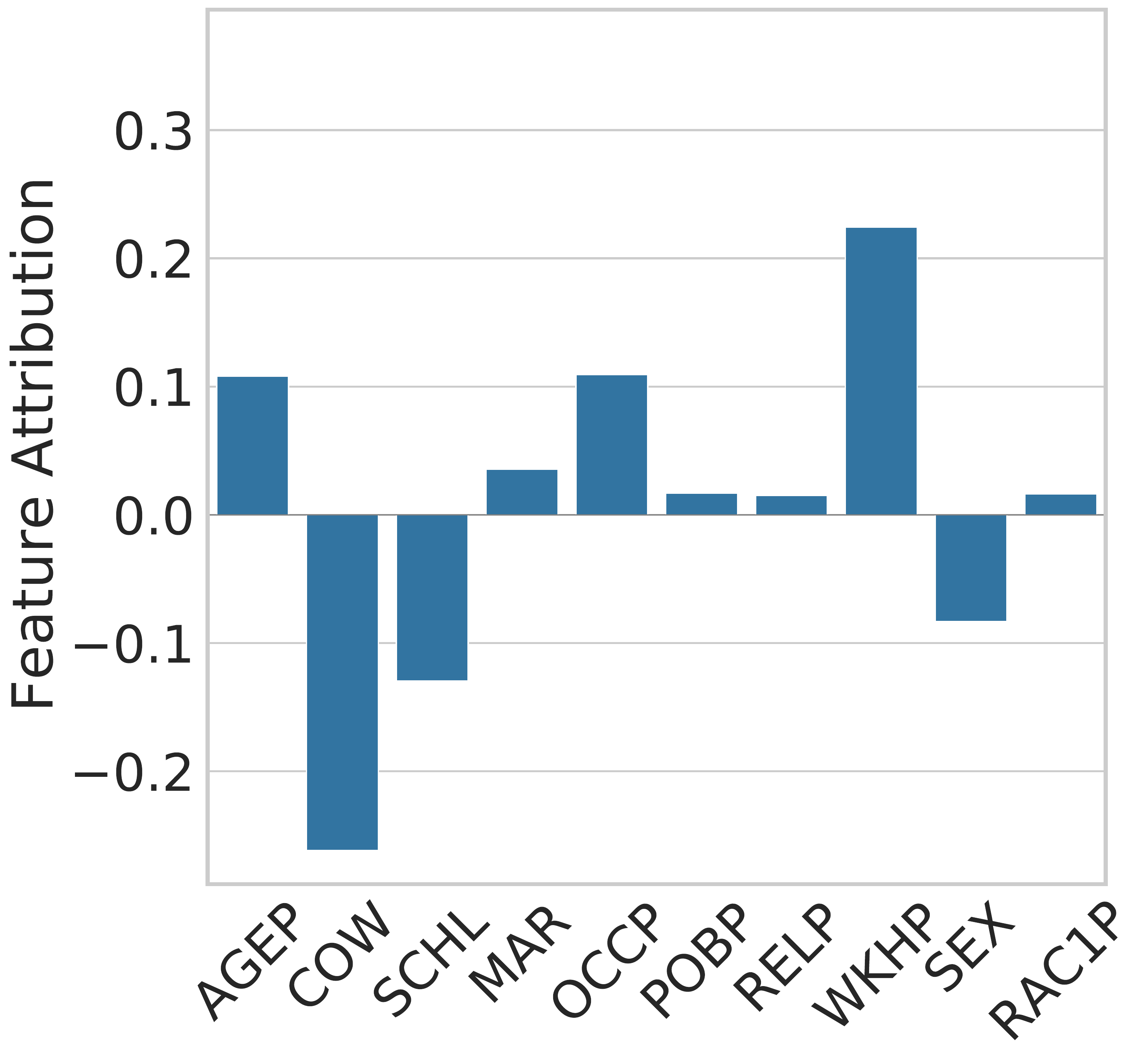} &
\includegraphics[width=0.225\textwidth]{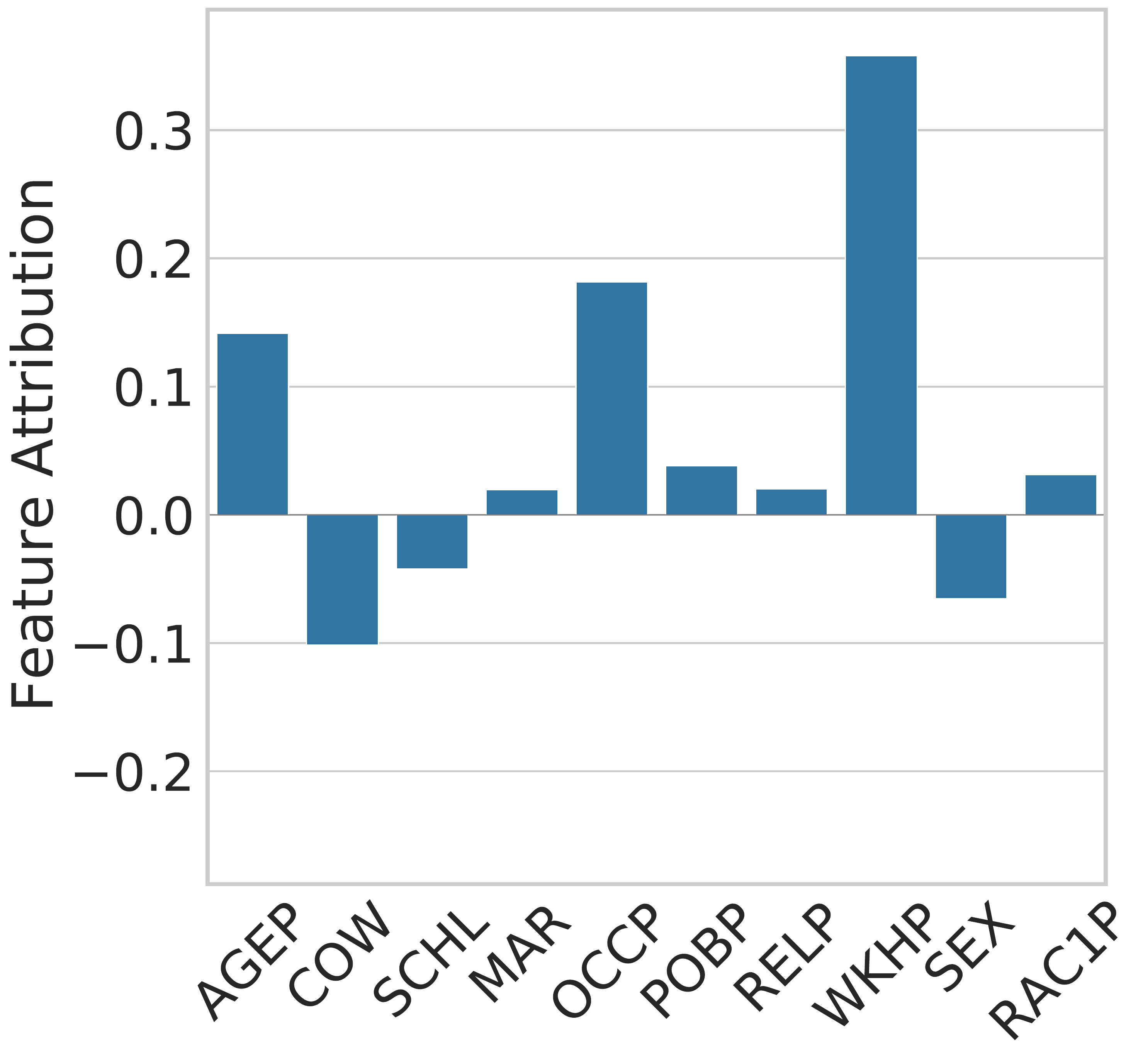}  \\

(a) SHAP & (b) LIME \\
\end{tabular}

\caption{For any given datapoint, different explanation algorithms might lead to very similar or completely different explanations. In many cases, however, there are both similarities and dissimilarities (compare Figure \ref{fig:similarities-and-dissimilarities} in the main paper). Every row depicts the explanations of the two different explanation algorithms for another individual. The Figure depicts the first 6 observations from the test set.}
\end{figure*}

\newpage

\begin{center} 
\huge \bf Additional Figures Related to Figure \ref{fig:decision-boundary} (a), (b) in the Main Paper
\end{center} 
\vspace{0.4cm}

\begin{figure*}[h]

\begin{tabular}{cc}

\includegraphics[width=0.225\textwidth]{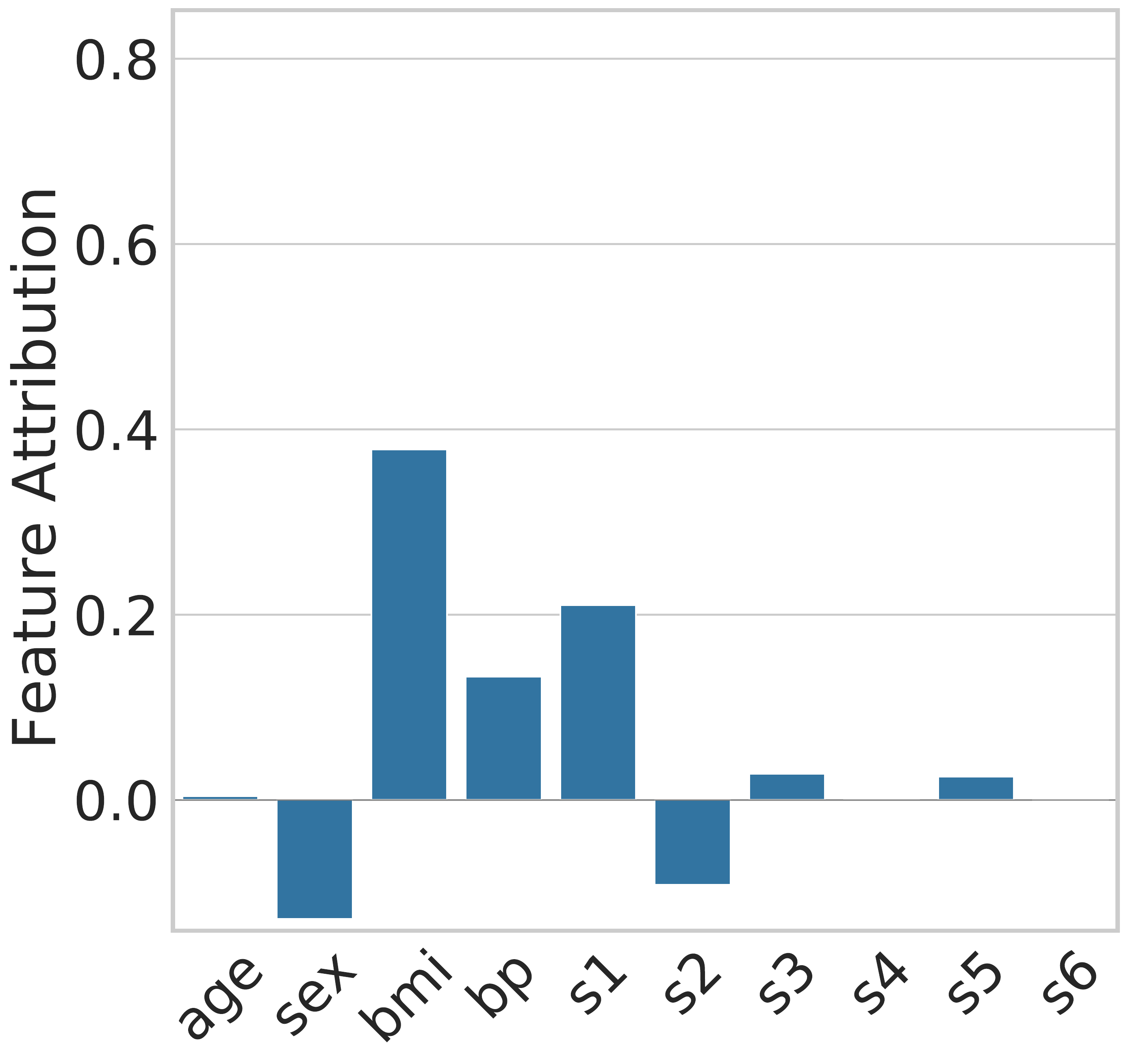} &
\includegraphics[width=0.225\textwidth]{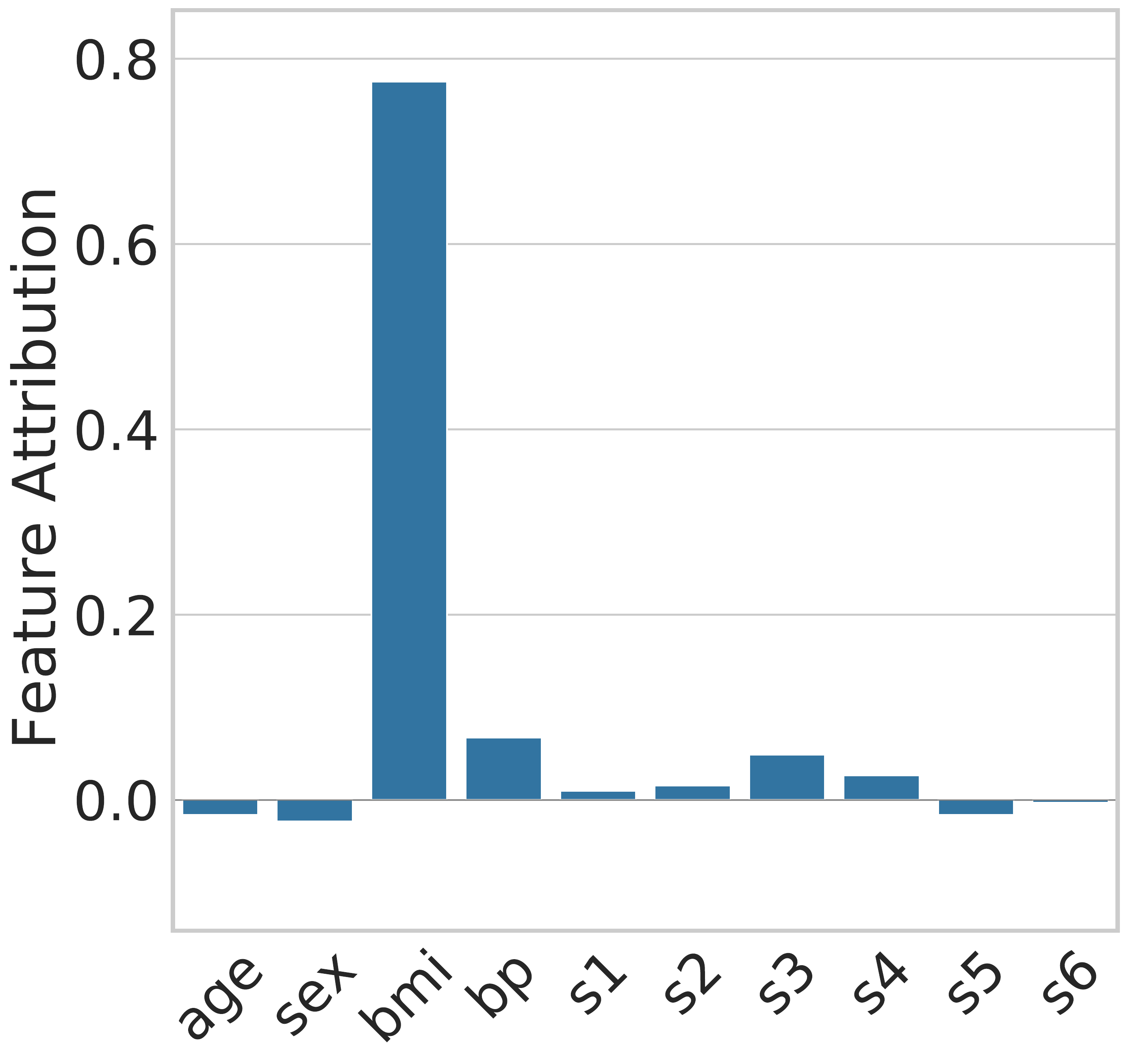}  \\
\includegraphics[width=0.225\textwidth]{supplement_figures/diabetes/1_linear.pdf} &
\includegraphics[width=0.225\textwidth]{supplement_figures/diabetes/1_forest.pdf}  \\
\includegraphics[width=0.225\textwidth]{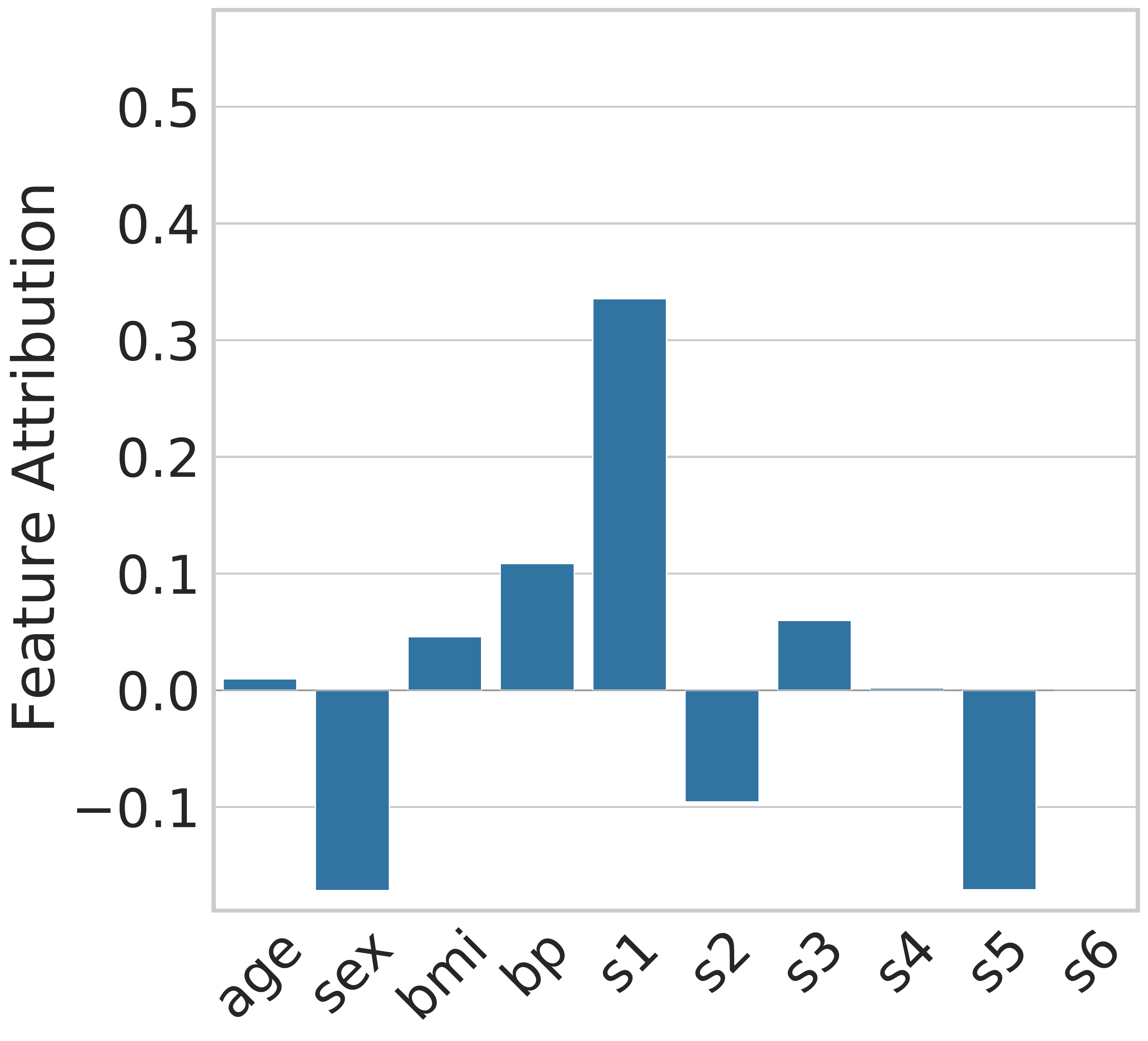} &
\includegraphics[width=0.225\textwidth]{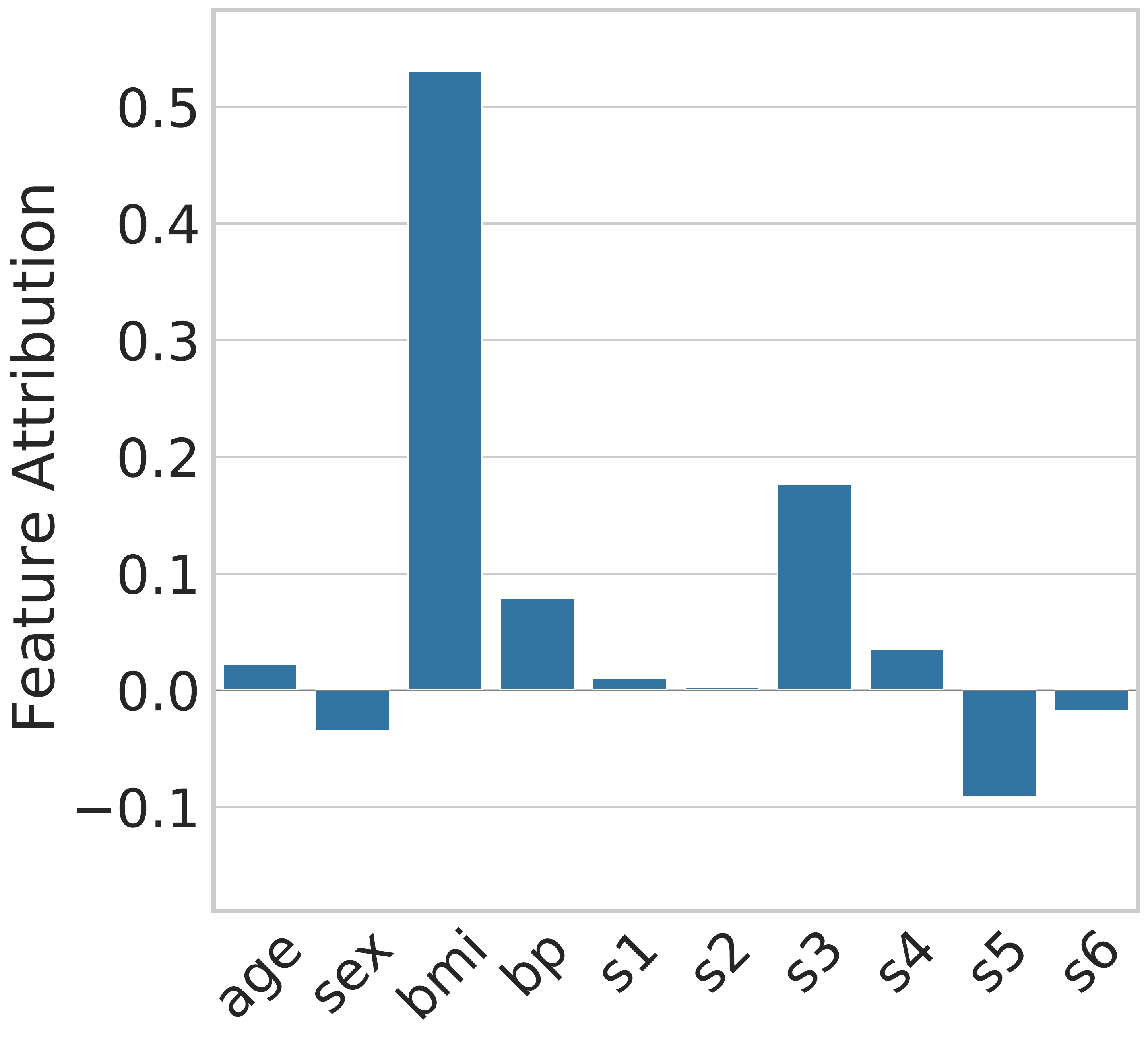}  \\
\includegraphics[width=0.225\textwidth]{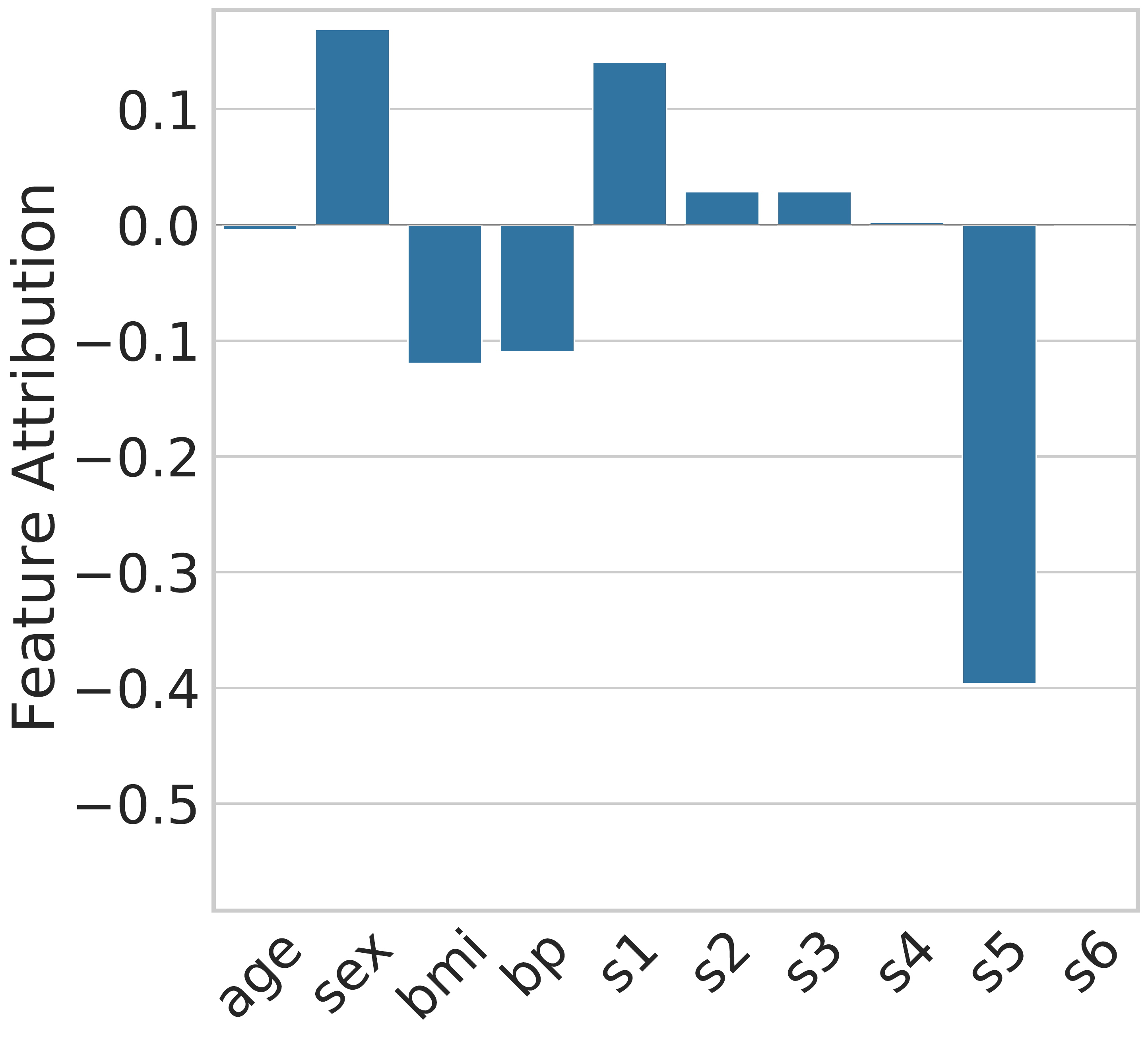} &
\includegraphics[width=0.225\textwidth]{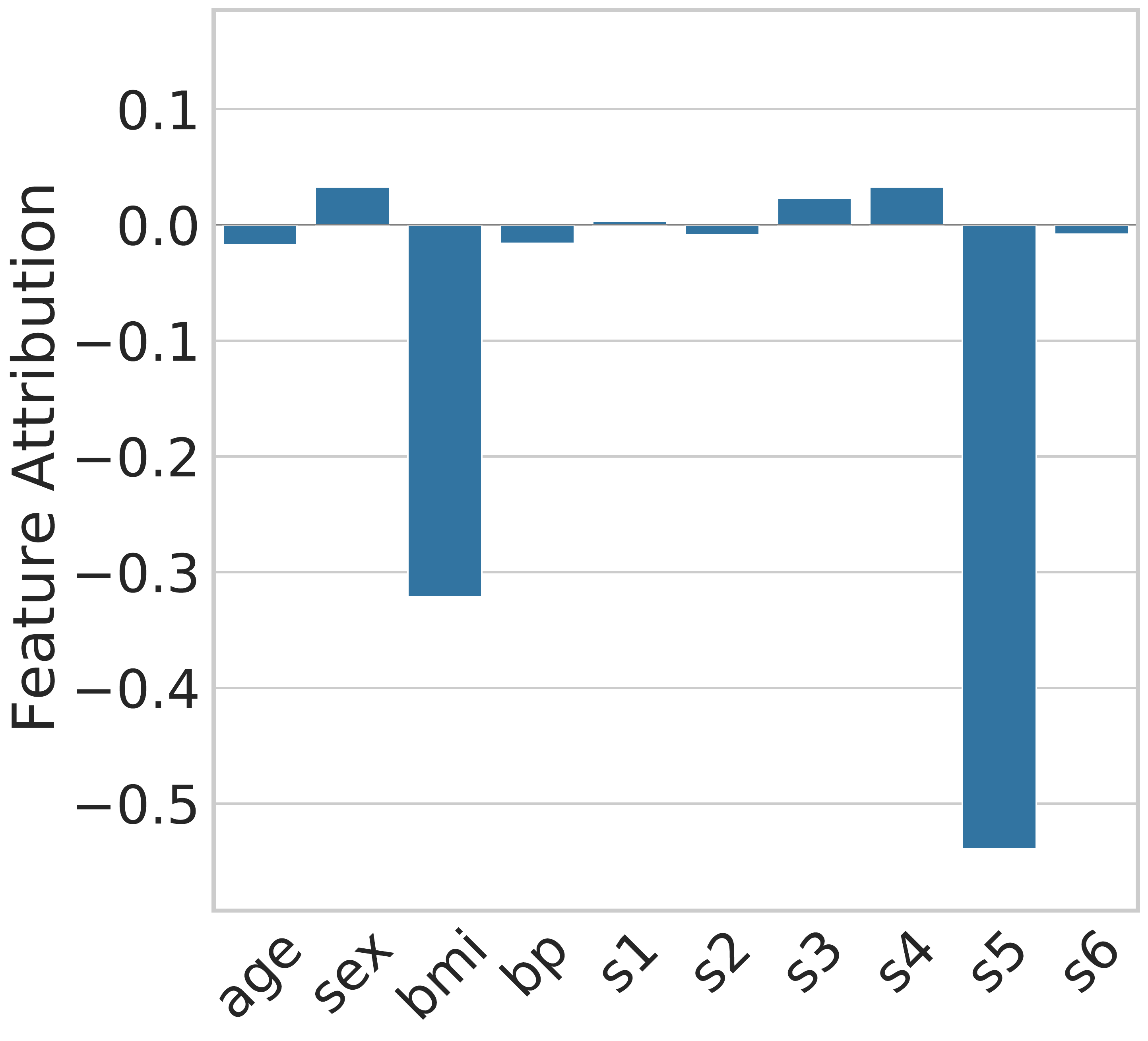}  \\
\includegraphics[width=0.225\textwidth]{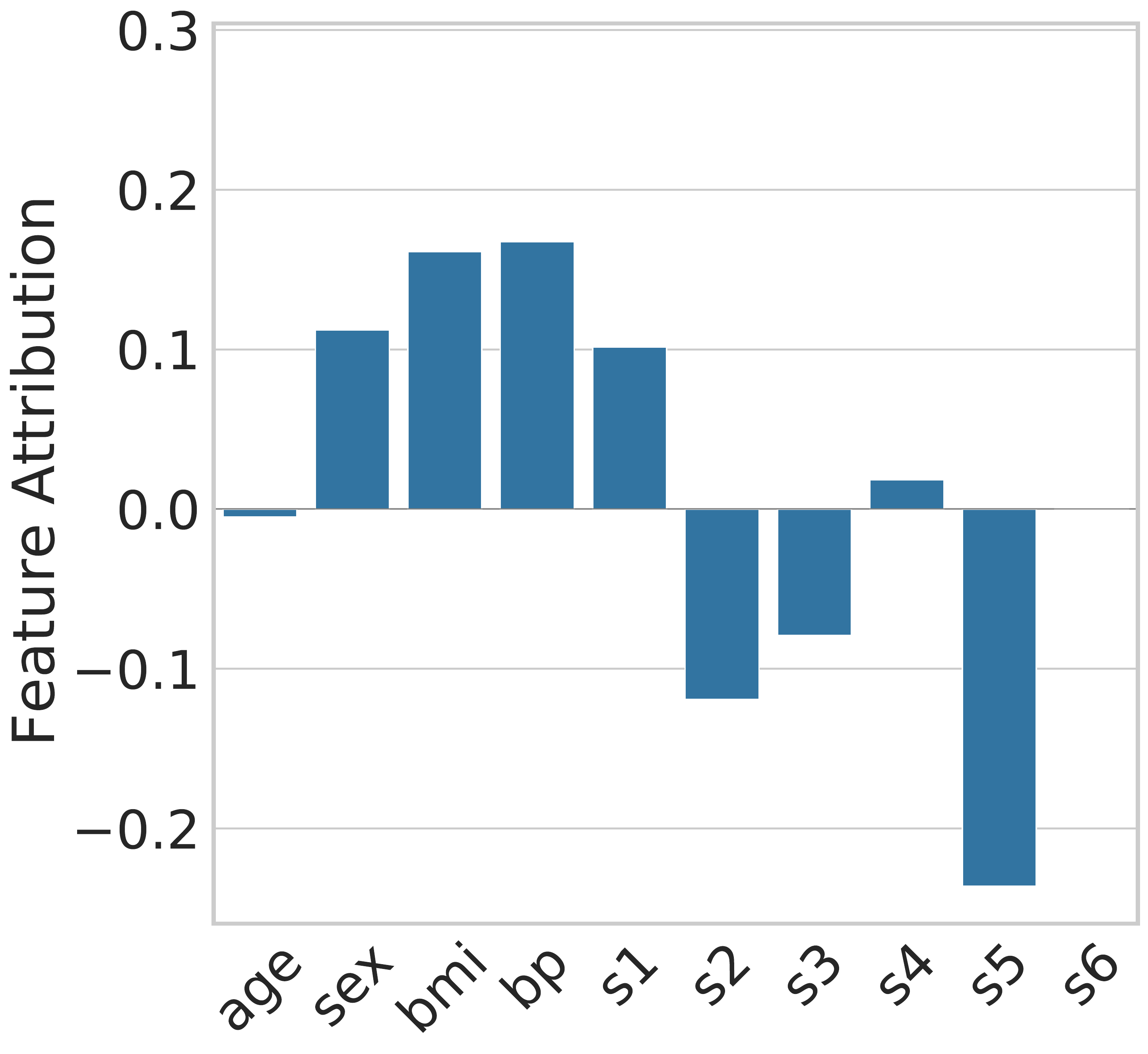} &
\includegraphics[width=0.225\textwidth]{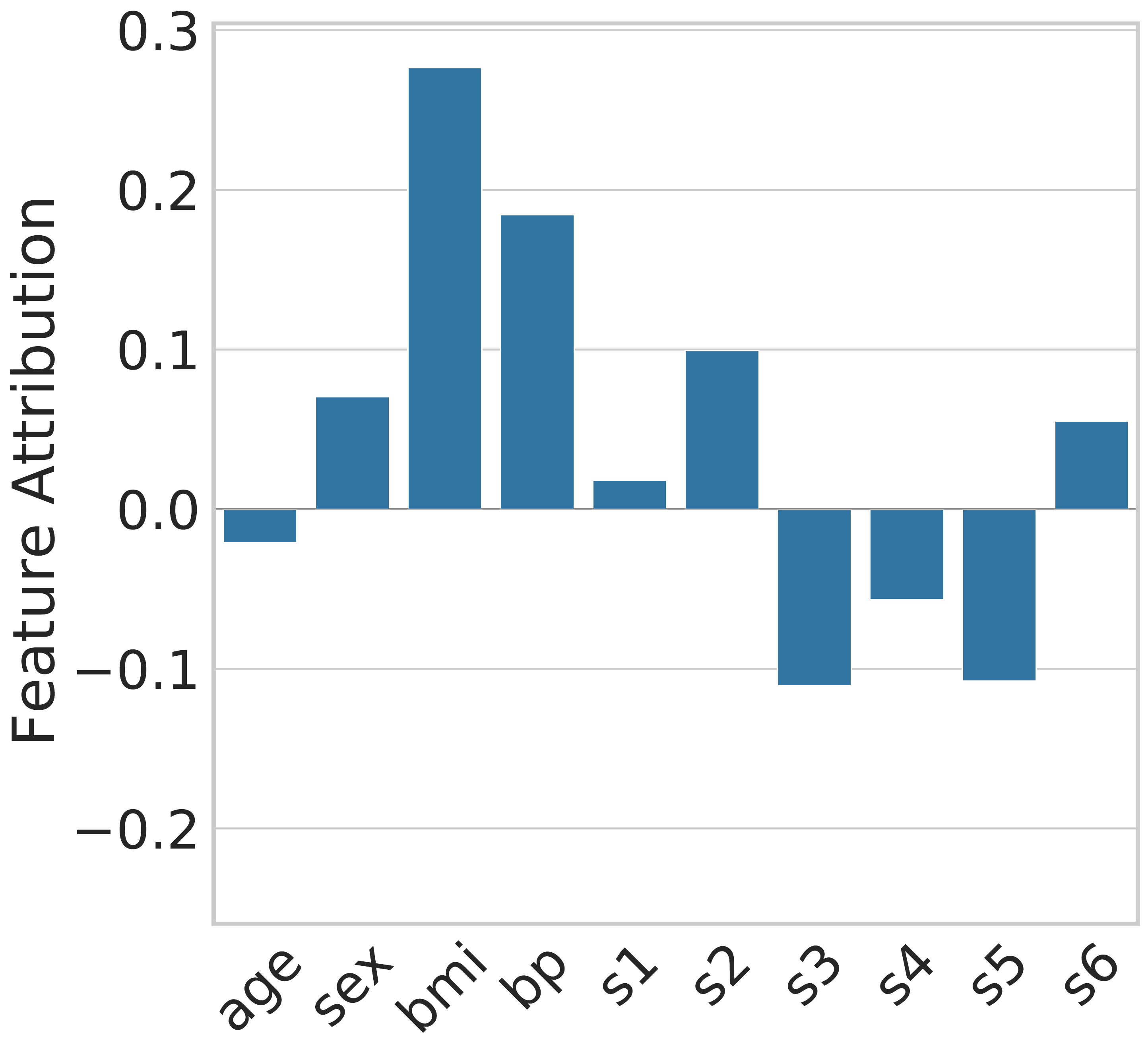}  \\
\includegraphics[width=0.225\textwidth]{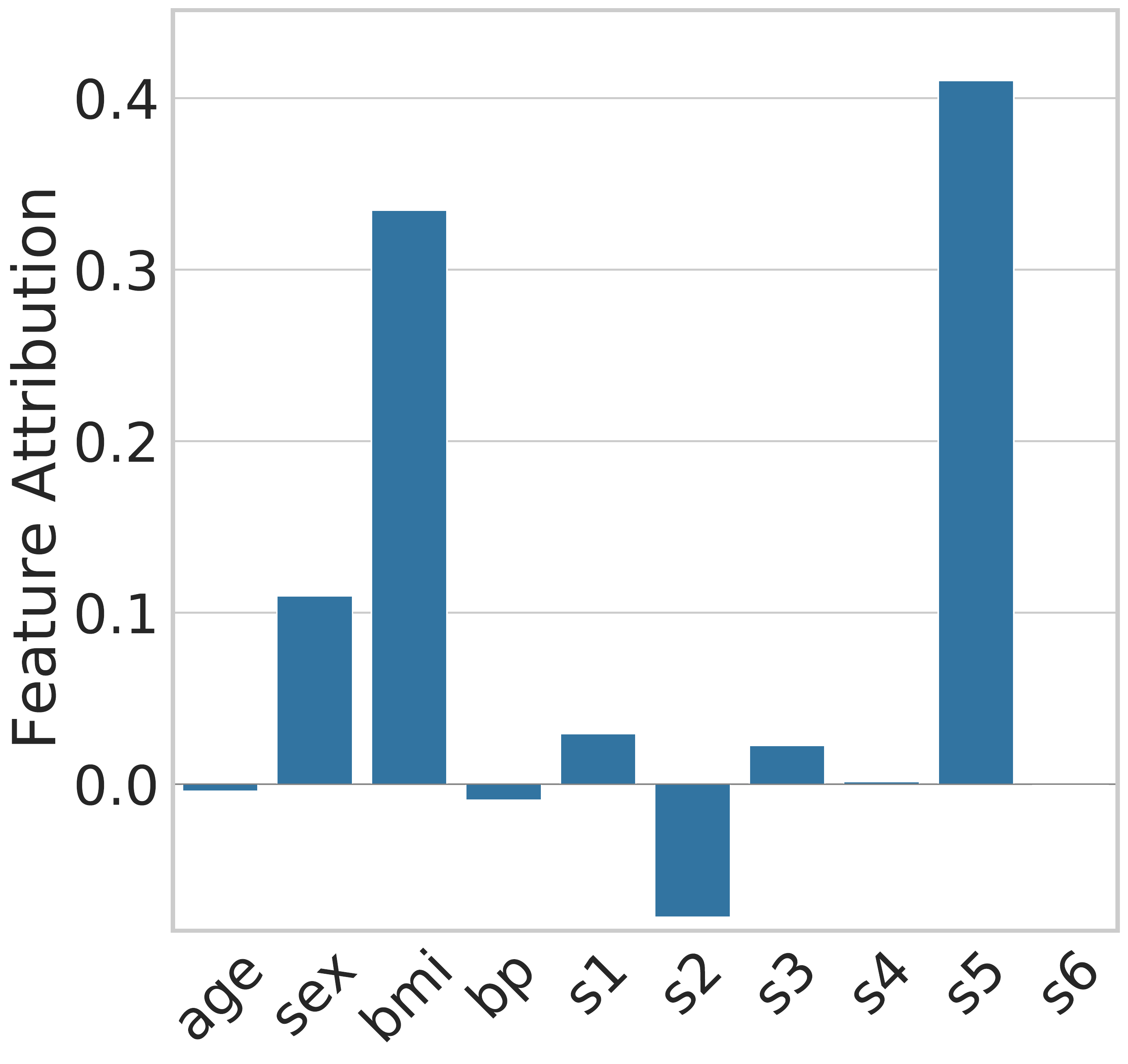} &
\includegraphics[width=0.225\textwidth]{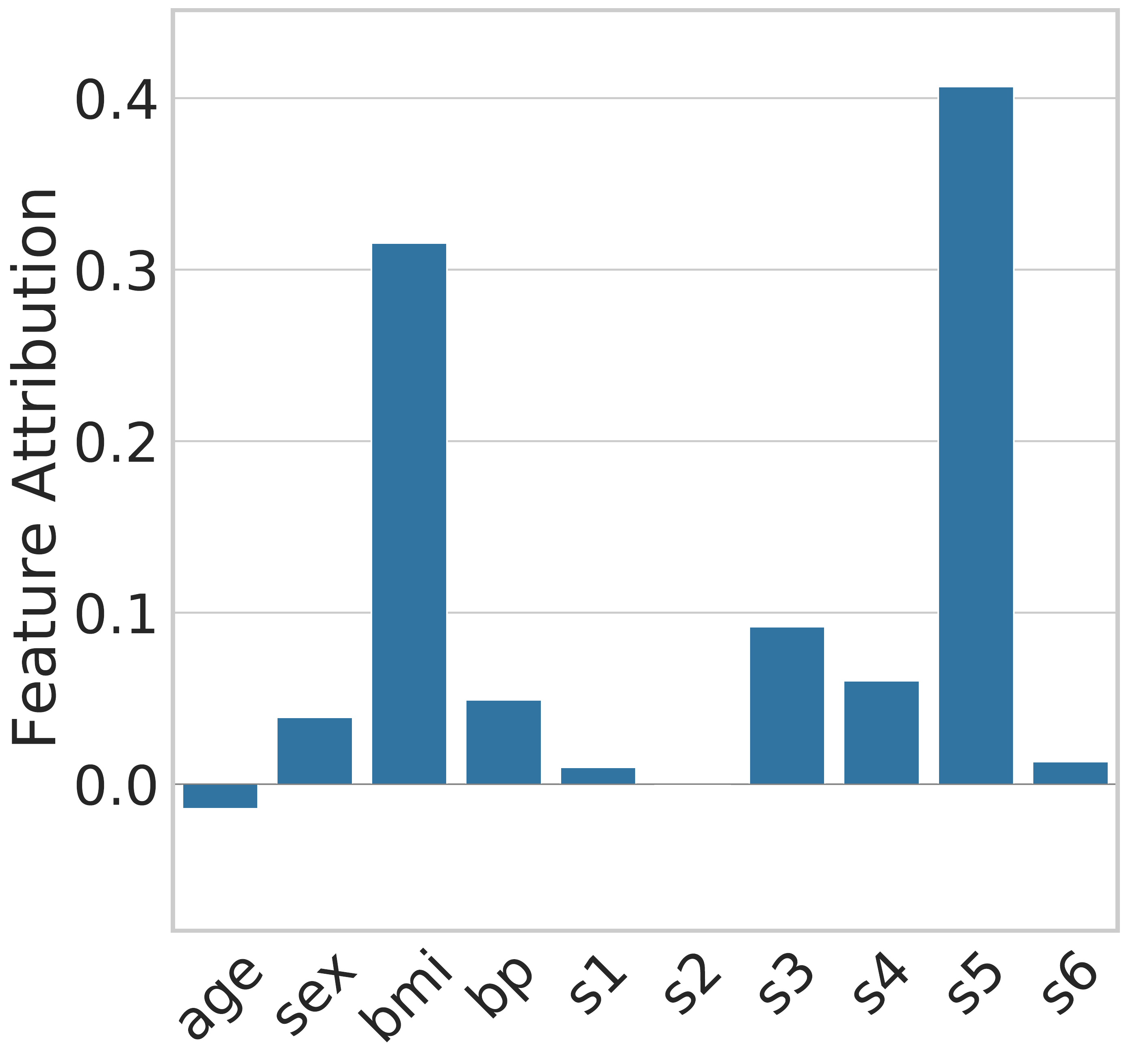}  \\

(a) Diabetes, Linear Regression & (b) Diabetes, Random Forest \\
\end{tabular}

\caption{Explanations depend on the exact shape of the decision boundary (compare Figure \ref{fig:decision-boundary} in the main paper). Every row depicts the explanations of the two different explanation algorithms for another individual. The Figure depicts the first 6 observations from the test set.}
\end{figure*}

\newpage

\begin{center} 
\huge \bf Additional Figures Related to Figure \ref{fig:decision-boundary} (c), (d) in the Main Paper
\end{center} 
\vspace{0.4cm}

\begin{figure*}[h]

\begin{tabular}{cc}

\includegraphics[width=0.24\textwidth]{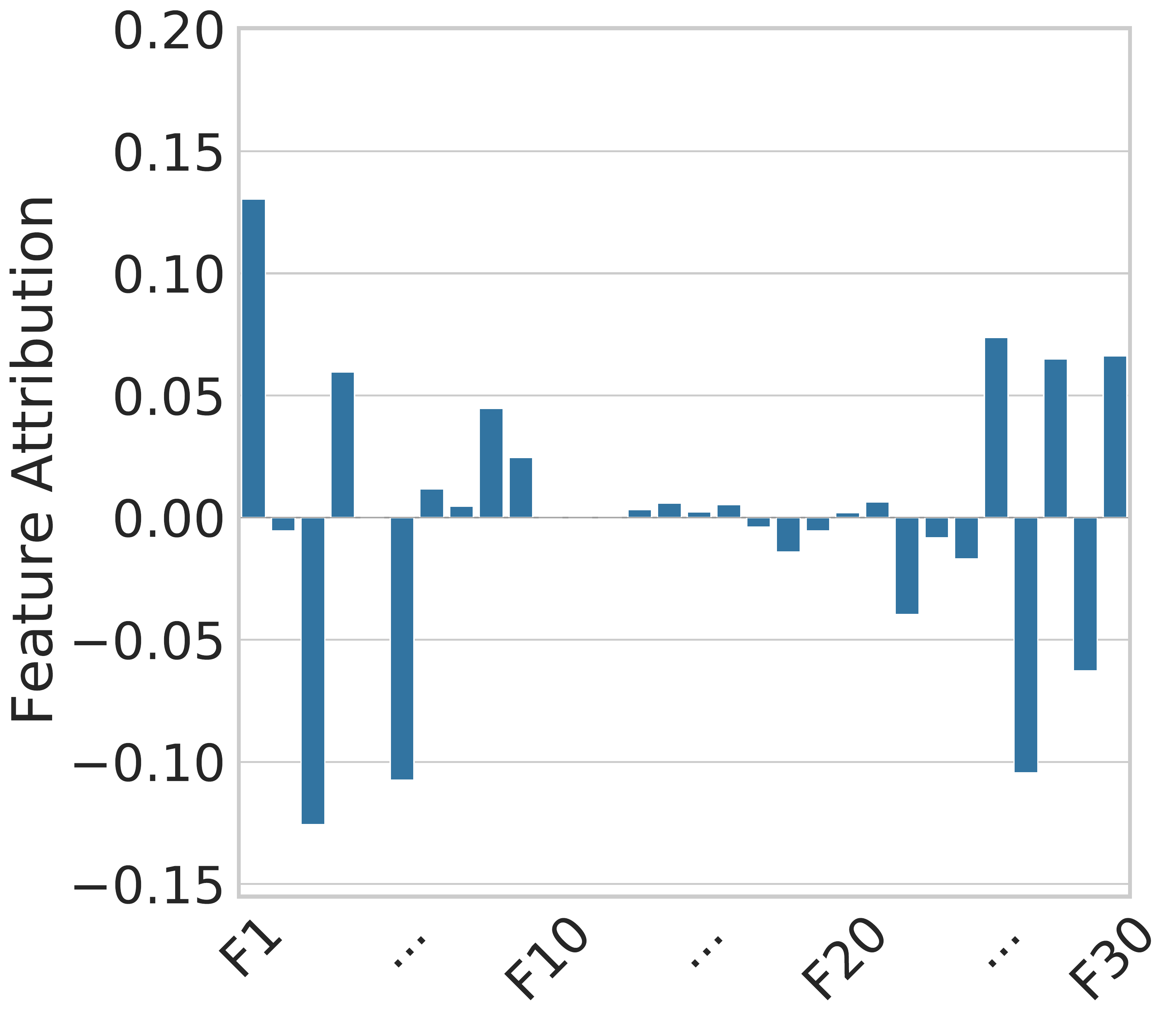} &
\includegraphics[width=0.24\textwidth]{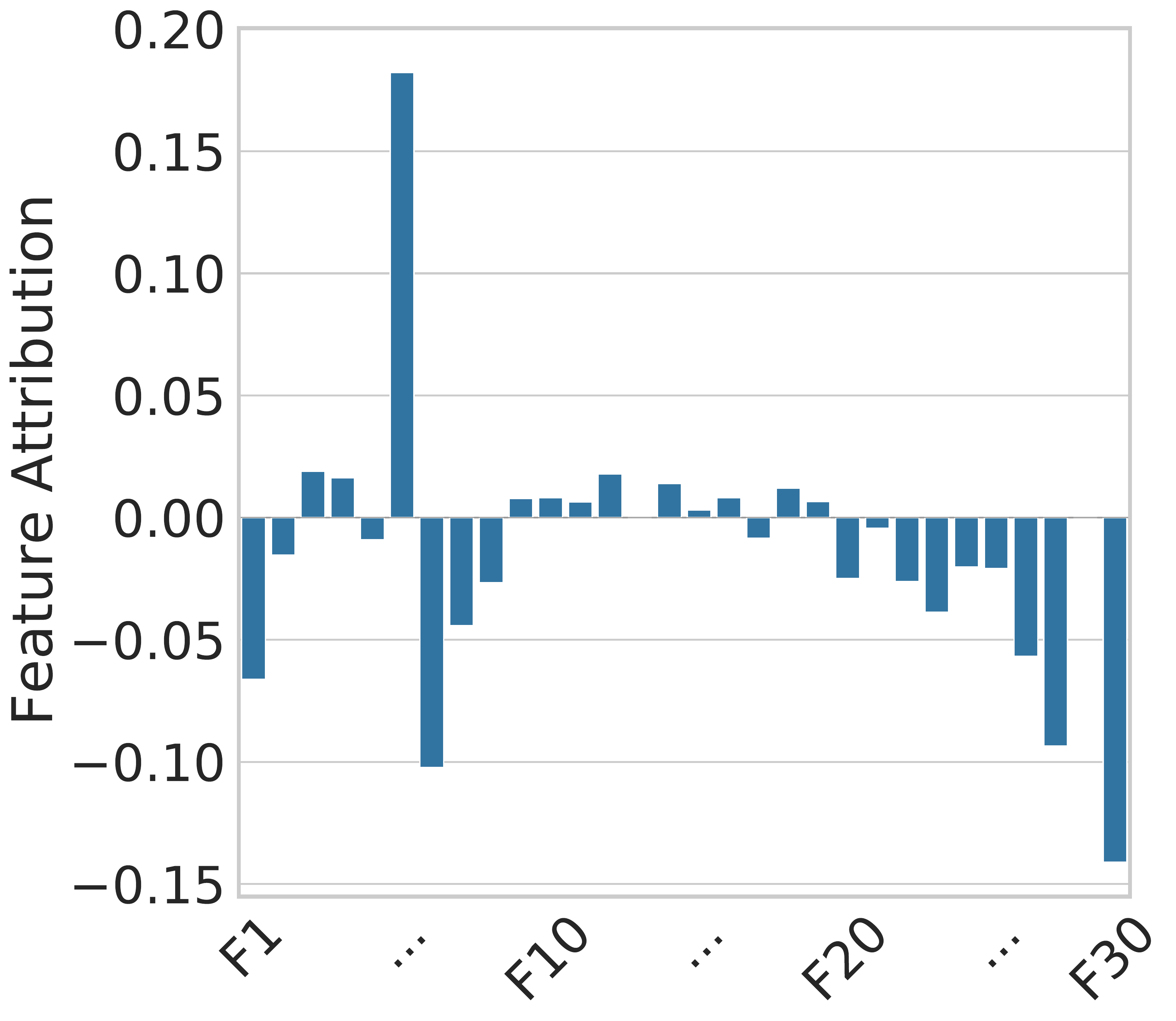}  \\
\includegraphics[width=0.24\textwidth]{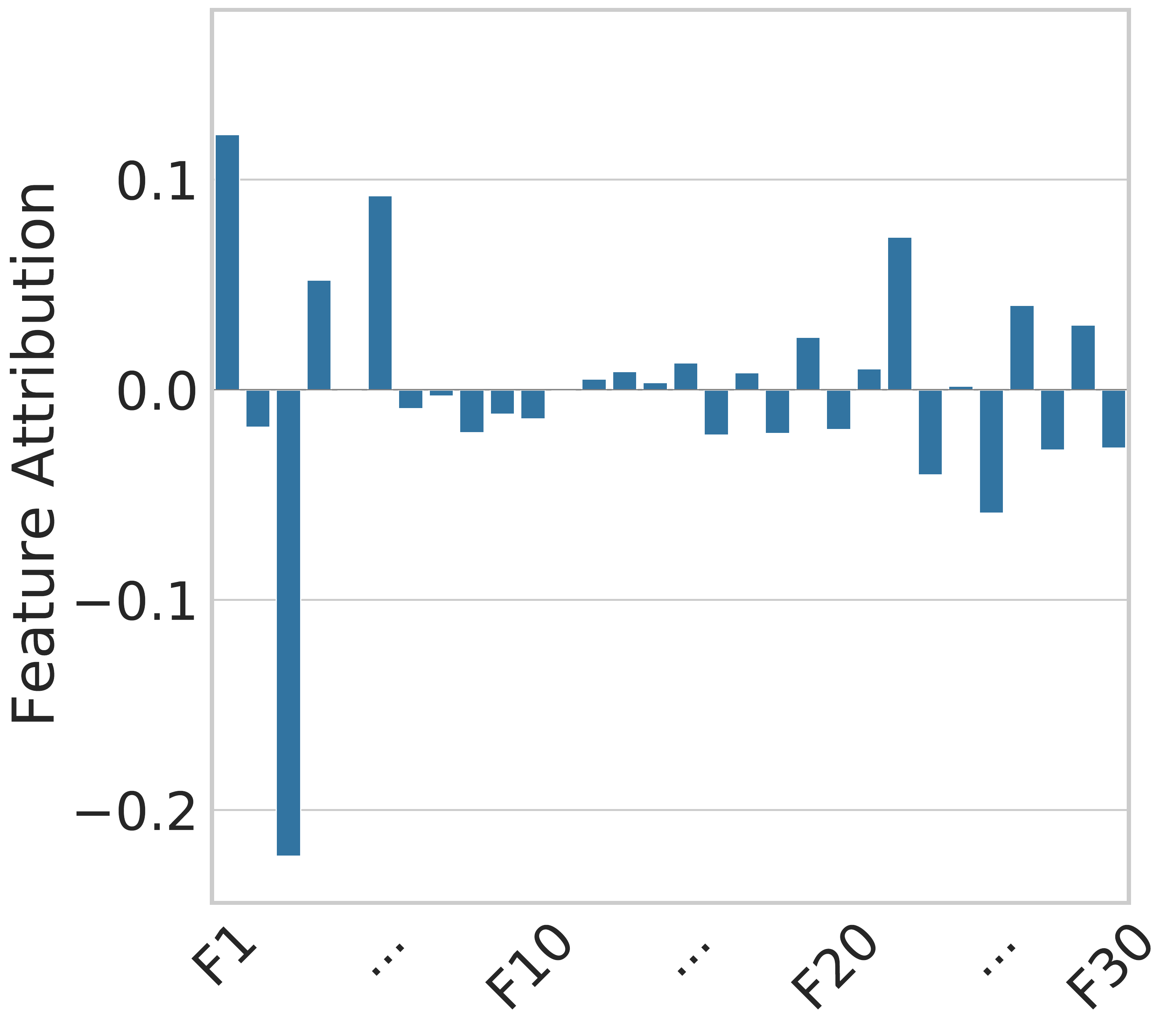} &
\includegraphics[width=0.24\textwidth]{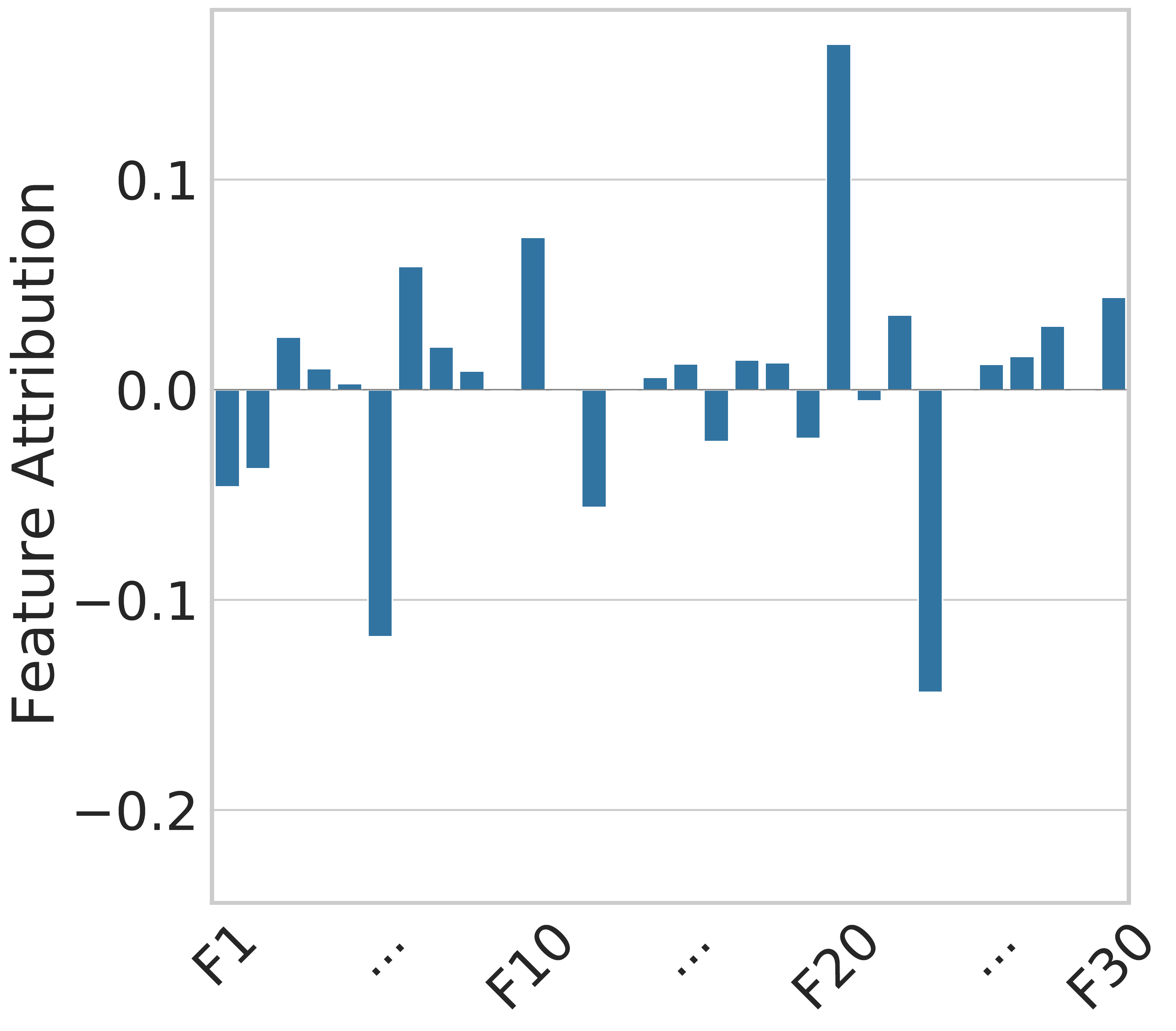}  \\
\includegraphics[width=0.24\textwidth]{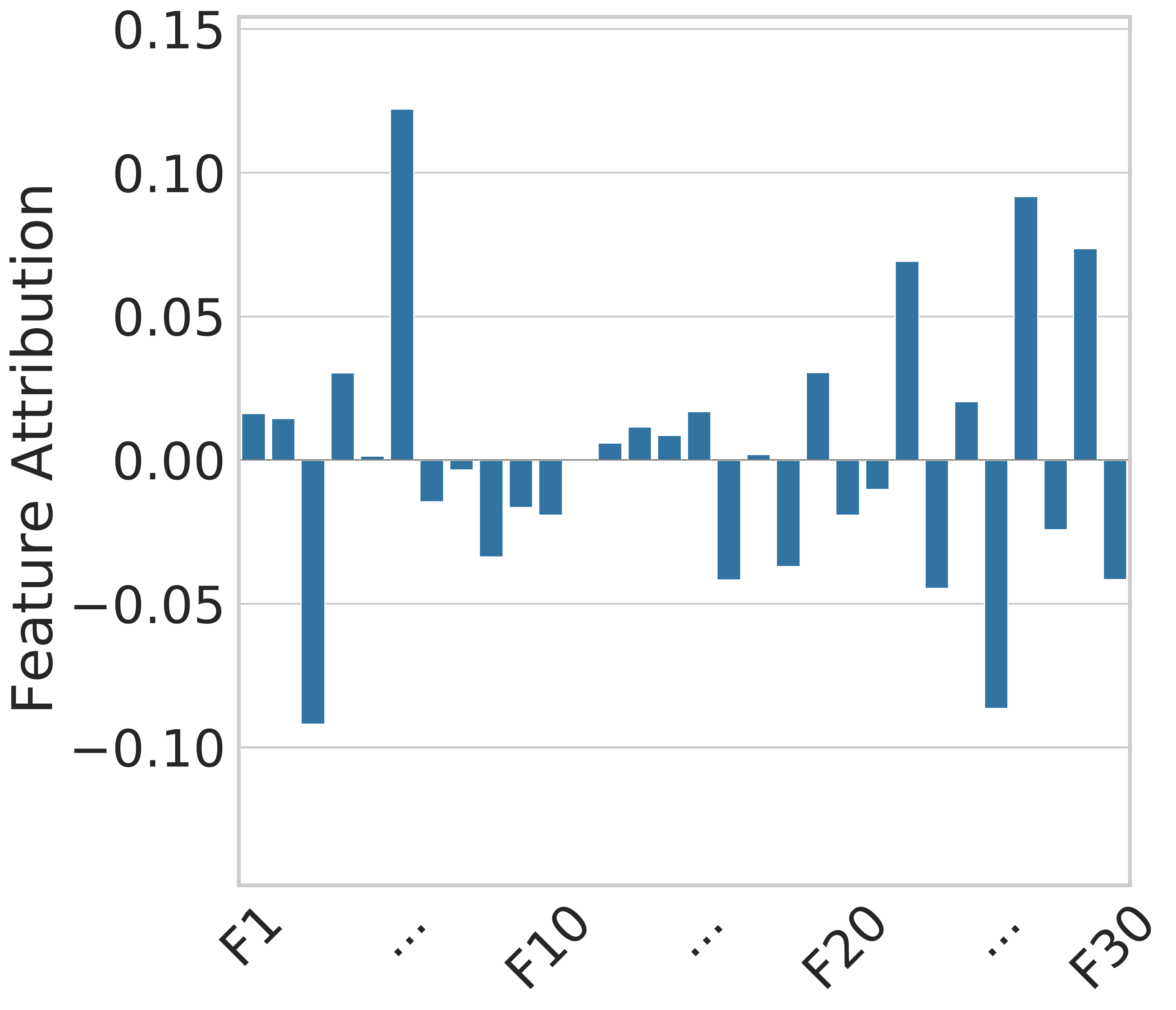} &
\includegraphics[width=0.24\textwidth]{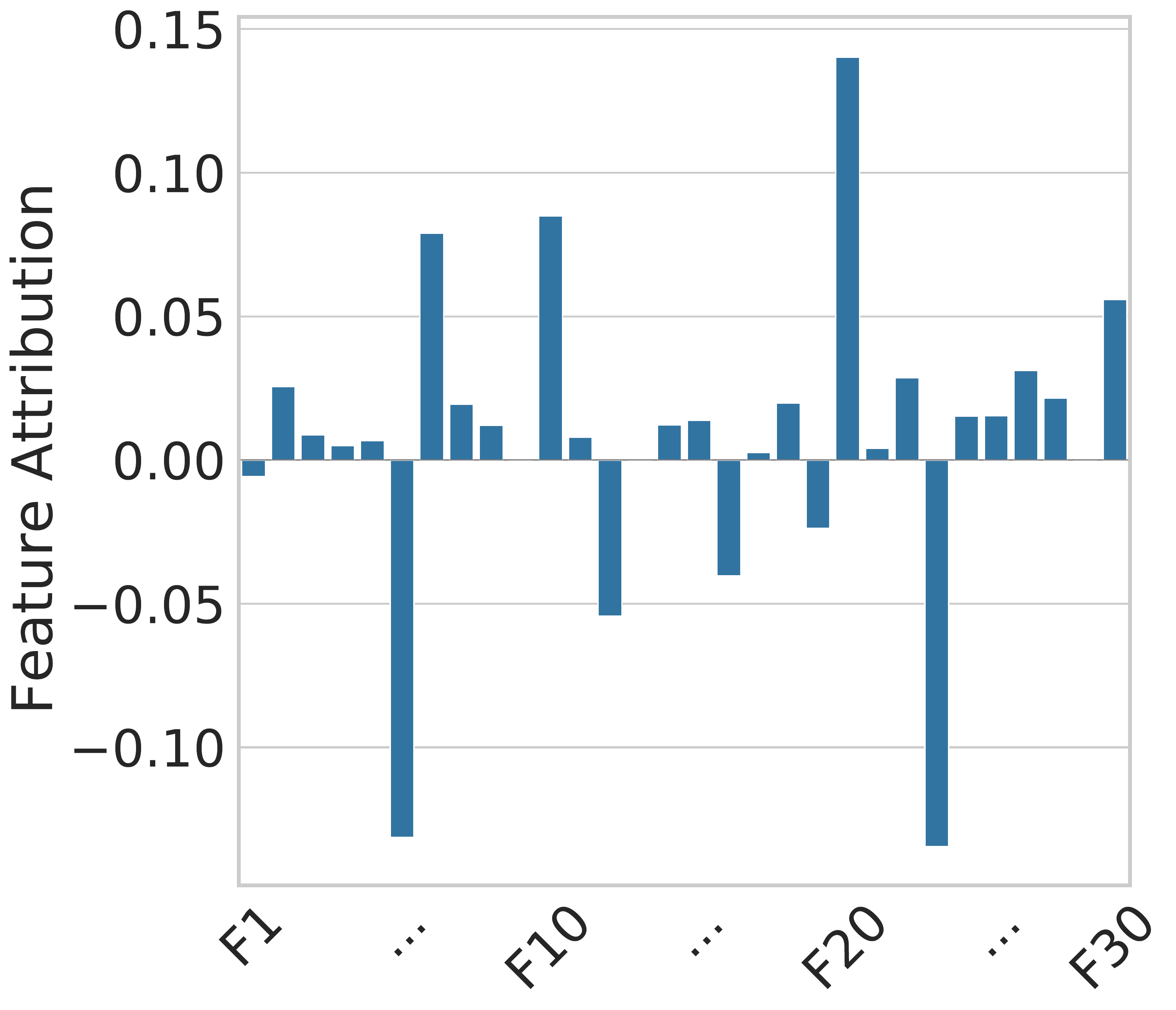}  \\
\includegraphics[width=0.24\textwidth]{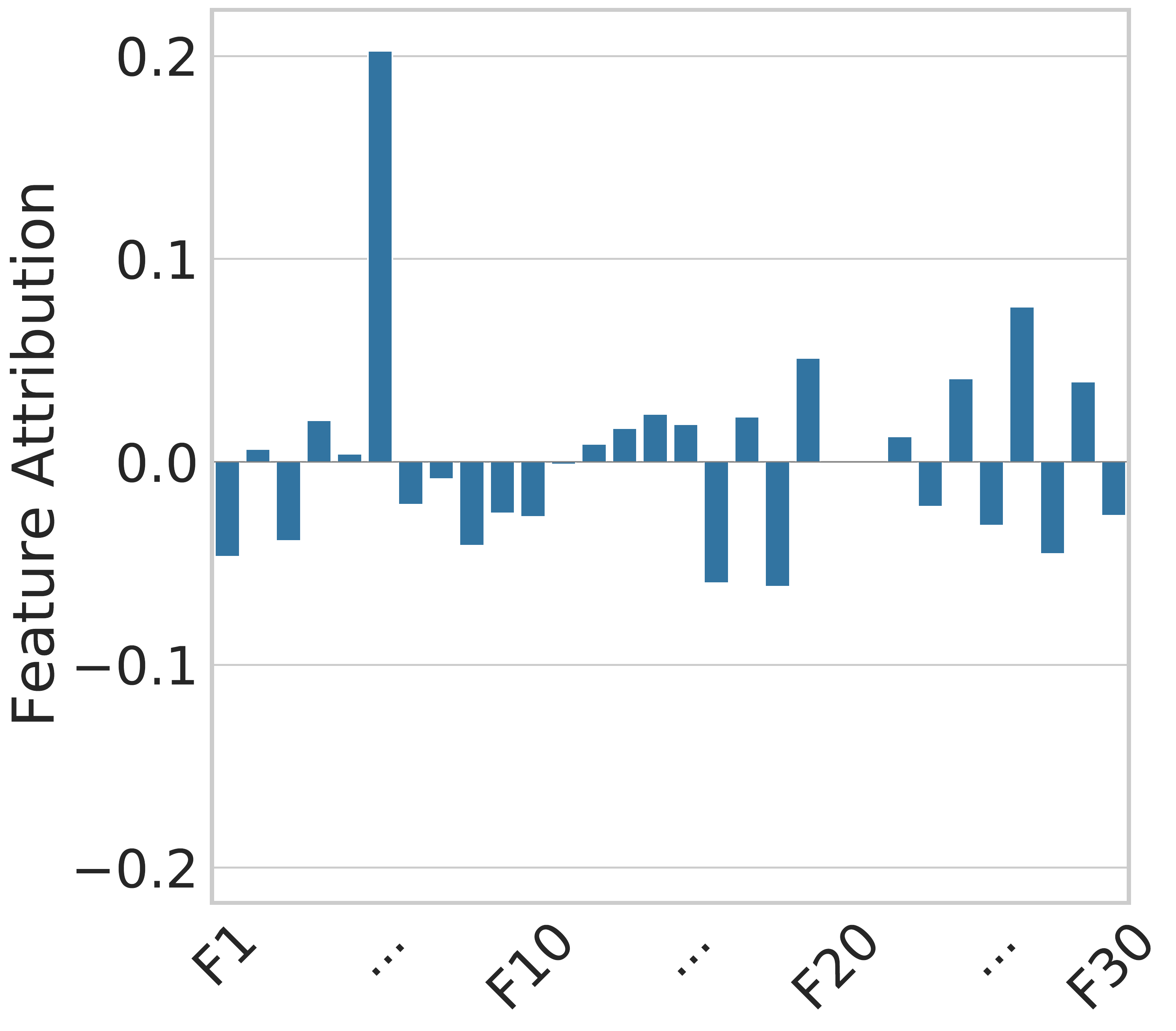} &
\includegraphics[width=0.24\textwidth]{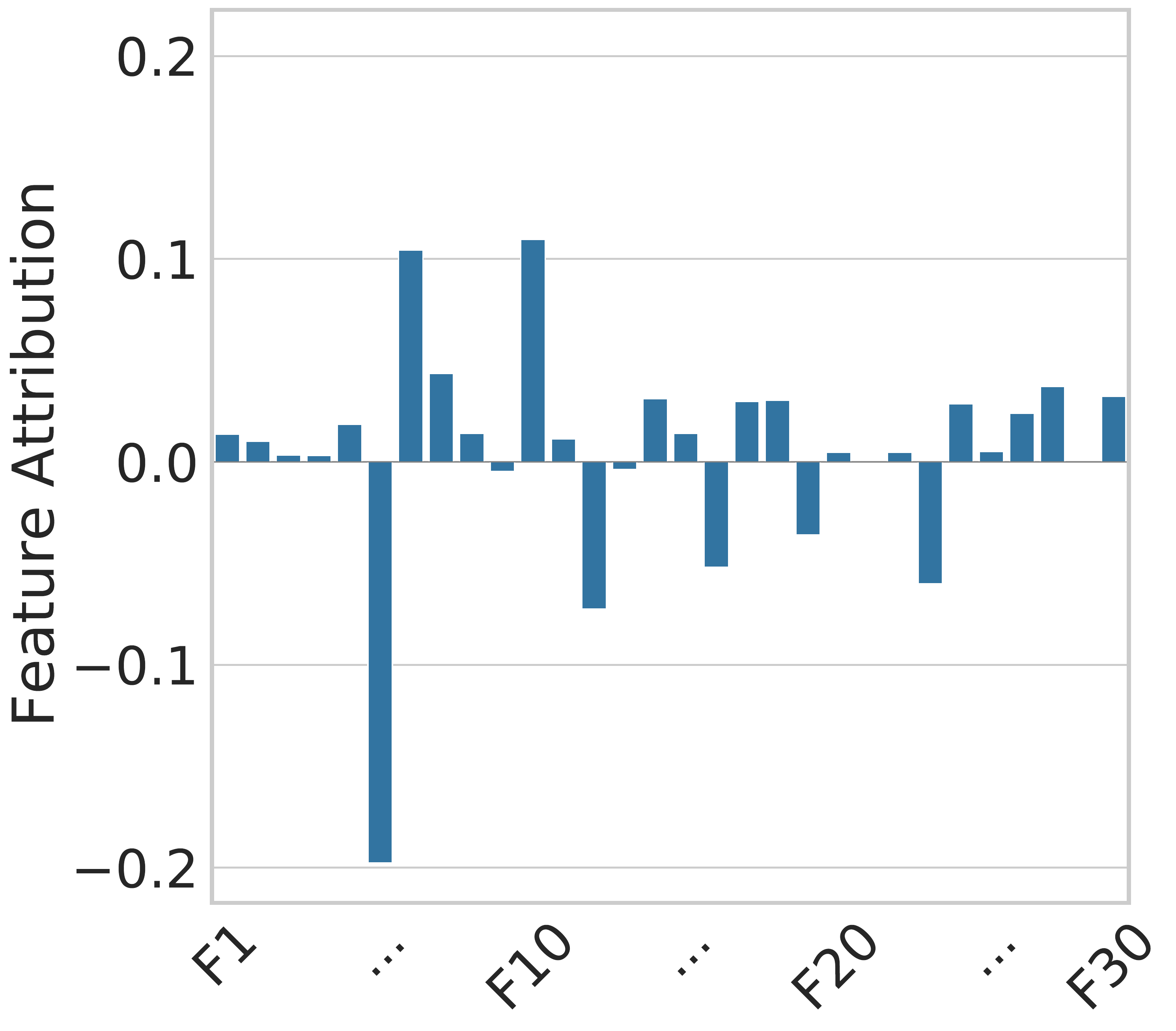}  \\
\includegraphics[width=0.24\textwidth]{supplement_figures/cancer/4_lr_random.pdf} &
\includegraphics[width=0.24\textwidth]{supplement_figures/cancer/4_lr_trained.pdf}  \\
\includegraphics[width=0.24\textwidth]{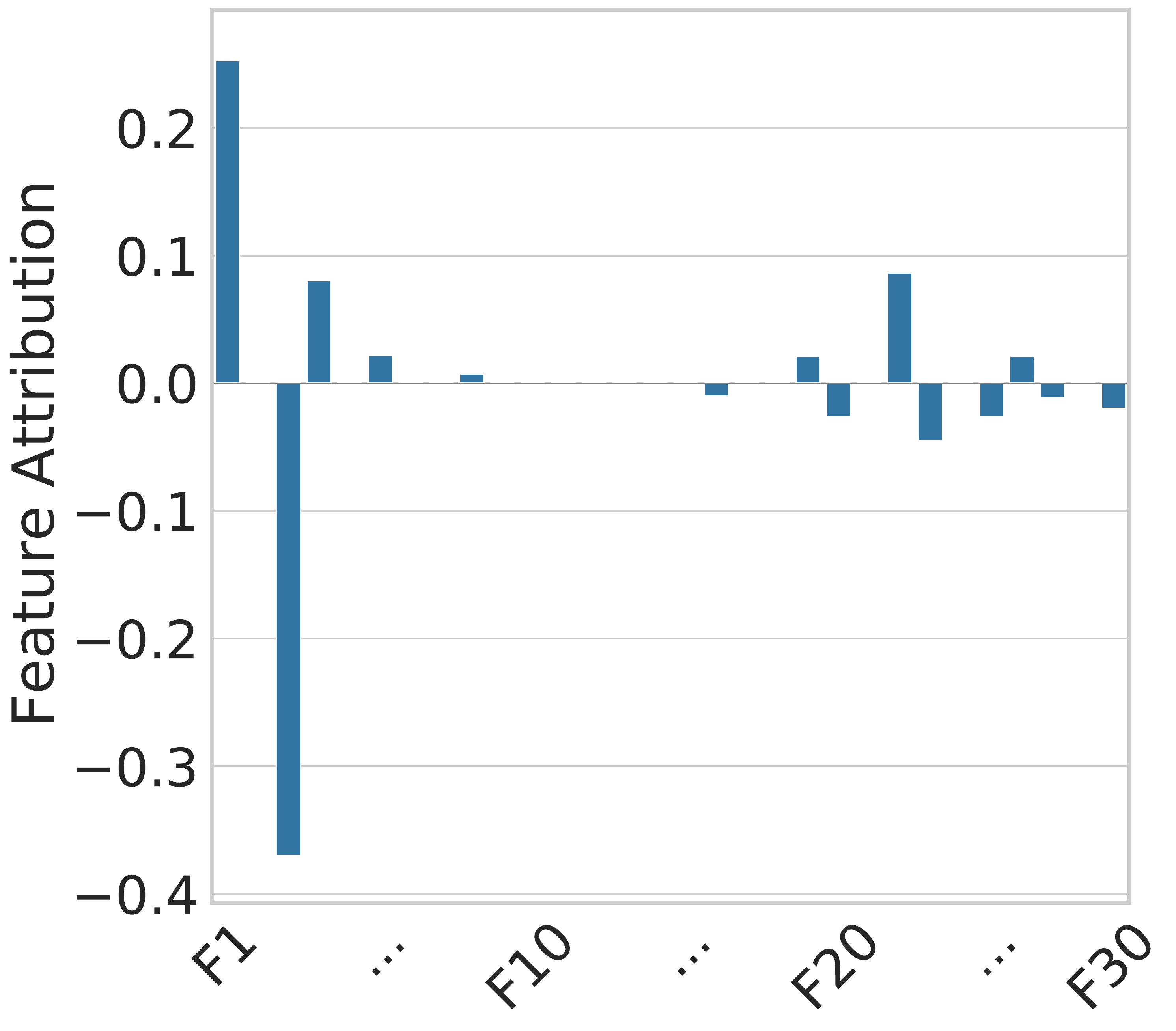} &
\includegraphics[width=0.24\textwidth]{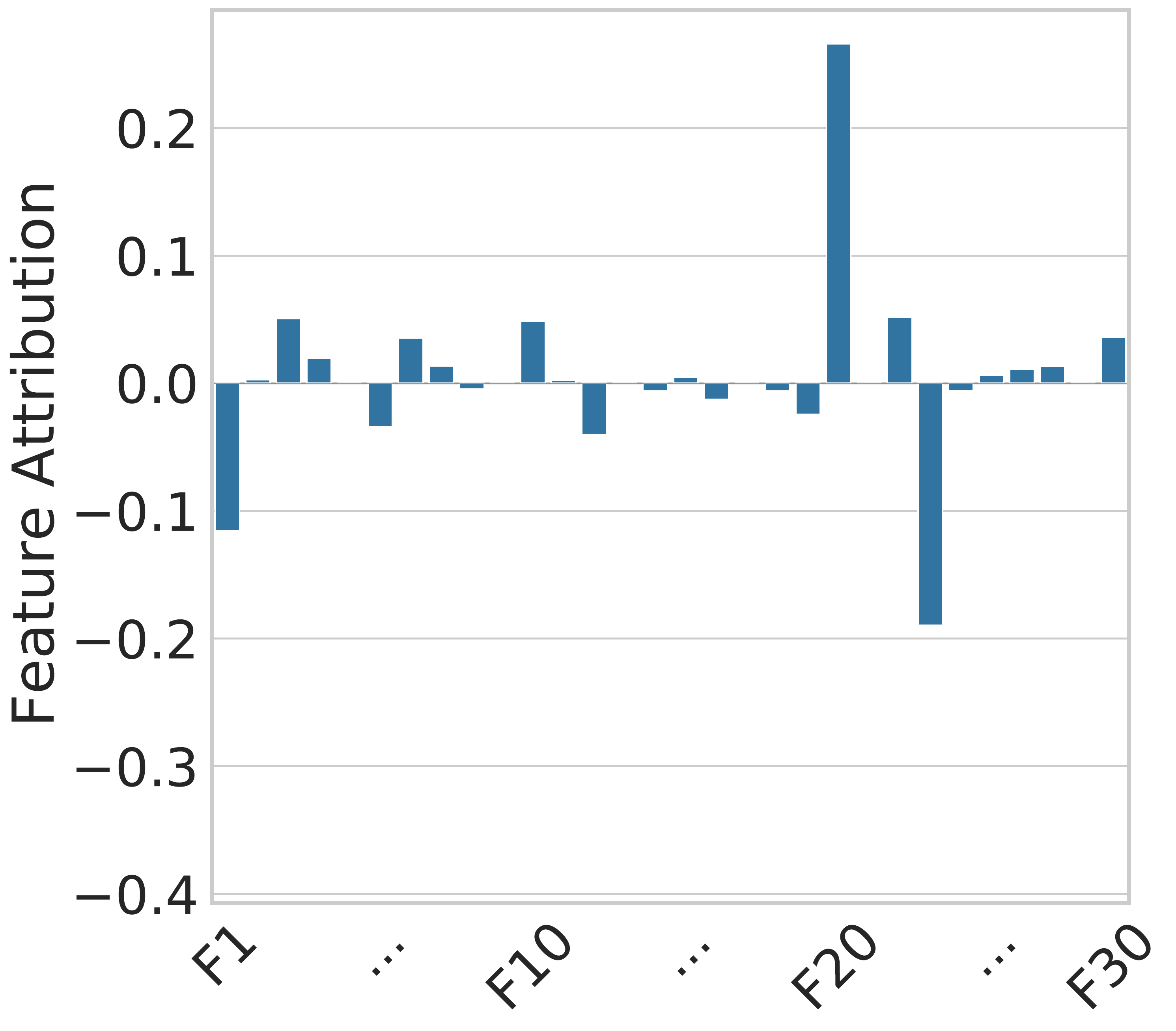}  \\

(a) Breast Cancer, 36\% Accuracy & (b) Breast Cancer, 96\% Accuracy \\
\end{tabular}

\caption{Explanations depend on the exact shape of the decision boundary (compare Figure \ref{fig:decision-boundary} in the main paper). Every row depicts the explanations of the two different explanation algorithms for another individual. The Figure depicts the first 6 observations from the test set.}
\end{figure*}